\documentclass{article}
\usepackage{log_2026}						
\usepackage{etoc}

\usepackage{booktabs}						
\usepackage{multirow}						
\usepackage{amsfonts}						
\usepackage{graphicx}						
\usepackage{duckuments}						

\usepackage{amsmath,amssymb,amsthm}

\usepackage{booktabs}
\usepackage{tabularx}
\usepackage{array}
\usepackage{bbm}

\newcolumntype{Y}{>{\centering\arraybackslash}X}

\newtheorem{proposition}{Proposition}

\usepackage[numbers,compress,sort]{natbib}	

\definecolor{darkgreen}{RGB}{0,120,0}
\definecolor{darkblue}{RGB}{0,70,170}
\definecolor{darkorange}{RGB}{220,120,0}

\newcommand{\first}[1]{\textcolor{darkgreen}{#1}}
\newcommand{\second}[1]{\textcolor{darkblue}{#1}}
\newcommand{\third}[1]{\textcolor{darkorange}{#1}}
\newcommand{\fourth}[1]{\textcolor{red}{#1}}

\title[]{SiGMA: Learning under Graph-Level Heterogeneity with\\ Gated Message-Passing and Attention}

\author[Pellegrin et al.]{%
Raphael Pellegrin \thanks{Equal contribution.}\\
Independent Researcher \\
\email{raphaelpellegrin@alumni.harvard.edu}
\And
Lukas Fesser \footnotemark[1]\\
Harvard University\\
\email{lukas\_fesser@g.harvard.edu}
\And
Melanie Weber\\
Harvard University\\
\email{mweber@seas.harvard.edu}
}

\definecolor{lightgreen}{rgb}{0.56,0.93,0.56}
\definecolor{lightblue}{rgb}{0.68, 0.85, 0.90}
\usepackage{makecell}

\begin{document}

\etocdepthtag.toc{main}

\maketitle

\begin{abstract}
In graph-level learning, graph neural networks are typically applied with a fixed computational template: the same depth, local message-passing operator, and global attention mechanism are used for every input graph. However, different graphs within the same dataset can vary substantially in prediction difficulty and may require different mixtures of local and non-local computation. We study this phenomenon as graph-level heterogeneity in both classification and regression tasks. To address it, we propose SiGMA (\underline{Si}gmoid \underline{G}ated \underline{M}essage-Passing and \underline{A}ttention), a gated hybrid graph neural network architecture that combines message-passing and attention heads within each layer. Inspired by the growing use of gated attention and gated computation in large language models and vision transformers, each head in SiGMA is modulated by a learned node-wise sigmoid gate, allowing the model to adaptively regulate which local or global mechanisms are active for different nodes, layers, and input graphs. The resulting architecture can emulate expressive message-passing backbones, admits near-identity computation paths that are not forced to over-smooth, and enables direct non-local communication beyond message-passing neighborhoods to mitigate over-squashing. Empirically, SiGMA achieves strong performance across graph-level benchmarks, consistently matching or improving over competitive graph transformer and hybrid baselines. Our results suggest that adapting computation to graph-level heterogeneity is a promising direction for building more flexible and effective graph learning architectures.\footnote{We make our code available at: \url{https://github.com/Weber-GeoML/SiGMA}.} 


\end{abstract}
\section{Introduction}

Graph neural networks (GNNs) have become a standard methodology for learning from relational data, with applications ranging from molecular prediction and protein modeling to social and biological networks. Most successful architectures are built around one of two computational paradigms. Message-passing neural networks aggregate information along graph edges and are effective at capturing local structure~\cite{gori2005new, hamilton2017inductive, Kipf:2017tc, xu2018powerful}. Graph transformers, in contrast, use attention to enable long-range interactions between nodes and have led to strong performance on graph benchmarks requiring non-local information~\cite{ying2021transformers,rampavsek2022recipe,muller2023attending}. Hybrid architectures such as GraphGPS combine these two mechanisms, reflecting the now common view that strong graph models should incorporate both local message passing and global attention~\cite{rampavsek2022recipe}. However, these models typically apply the same computational template to every input graph: the same depth, the same local--global balance, and the same set of graph operators are used regardless of the structure or difficulty of the instance.

\begin{figure}[t]
    \centering
    \includegraphics[width=\textwidth]{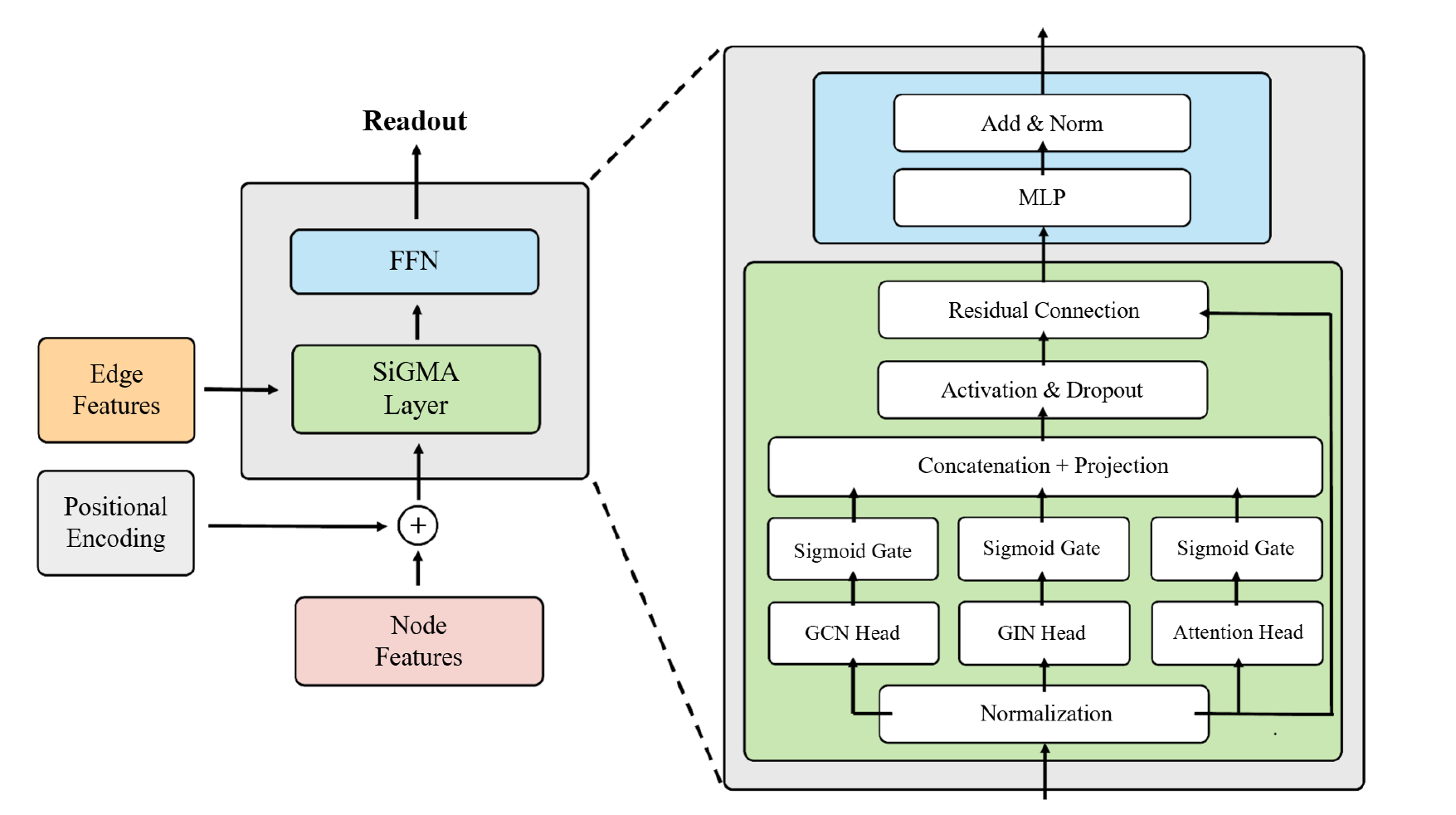}
    \caption{\textbf{The SiGMA architecture.} Node features, positional encodings, and edge features are processed by a stack of SiGMA layers followed by feed-forward blocks. Each SiGMA layer normalizes node representations, applies one or more parallel message-passing (here GCN and GIN) and attention heads, modulates each head with a learned sigmoid gate, concatenates and projects the gated heads, and updates the representation through a residual connection (optionally with dropout). The gating allows the model to ignore individual heads for a given input graph, or skip entire layers.}
    \label{fig:sigma_architecture}
    \vspace{-5mm}
\end{figure}

This fixed-computation perspective is increasingly at odds with evidence that graph learning problems are heterogeneous. Prior work has shown that, even in node-level tasks, different nodes in the same graph can vary substantially in prediction difficulty, and that this variation is related to structural properties such as degree and local graph context~\cite{li2023a}. In this work, we argue that an analogous phenomenon occurs at the graph level: in both graph classification and graph regression tasks, some input graphs are consistently easier or harder than others, and different graphs may benefit from different architectural choices. For example, graphs whose labels depend primarily on local motifs may be well served by message passing, whereas graphs requiring long-range interactions may require attention; similarly, some instances may benefit from greater effective depth while others may be harmed by unnecessary propagation. We refer to this phenomenon as \emph{graph-level heterogeneity}.

Existing architectures only partially address this issue. Pure message-passing GNNs (MPGNNs) have a strong local inductive bias, but they can suffer from over-squashing when information from many distant nodes must pass through narrow graph bottlenecks~\cite{alon2021on,topping2022understanding,black2023understanding}. Increasing the number of message-passing layers expands the receptive field, but it can also lead to over-smoothing, in which node representations become increasingly indistinguishable~\cite{li2018deeper,rong2019dropedge,nguyen2023revisiting}. Graph transformers alleviate some long-range communication issues through attention, but full attention can weaken useful locality biases and may not be necessary for every graph. Hybrid models improve this tradeoff by combining local and global computation, but they usually do so with a fixed allocation of message-passing and attention components. Thus, while prior architectures provide useful computational primitives, they do not explicitly adapt the choice, strength, or effective depth of these primitives to the input.

To address this limitation, we propose SiGMA (\underline{Si}gmoid \underline{G}ated \underline{M}essage-Passing and \underline{A}ttention), a gated hybrid graph architecture inspired by recent advances in NLP \cite{qiu2026gated} and vision \cite{bondarenko2023quantizable, lygizou2025differential, fesser2026unifying}, where gating mechanisms have been used to stabilize training, regulate computation, and improve the flexibility of attention-based models. Each SiGMA layer contains a collection of attention heads and message-passing heads. Attention heads provide non-local communication, while message-passing heads instantiate local graph operators such as GCN \cite{Kipf:2017tc}, GIN \cite{xu2018powerful}, GraphSAGE \cite{hamilton2017inductive}, or GAT \cite{velivckovic2017graph}. Crucially, each head is equipped with a learned sigmoid gate that multiplicatively modulates its output before the heads are concatenated, projected, and added through a residual connection. This allows the model to regulate which heads are active at each layer. As a result, the architecture can emphasize local message passing, global attention, or mixtures of both depending on the input graph, while suppressed heads and layers contribute only weakly to the residual update.

Empirically, we show that these properties translate into strong performance across graph-level classification and regression benchmarks. Our contributions are as follows:
\begin{itemize}
    \item We identify and study \emph{graph-level heterogeneity} in both graph classification and regression tasks, showing that prediction difficulty and preferred computational mechanisms vary across graphs.
    \item We propose \emph{SiGMA (\underline{Si}gmoid \underline{G}ated \underline{M}essage-Passing and \underline{A}ttention)}, a novel gated hybrid architecture inspired by advances in NLP and vision, designed to account for heterogeneity by adaptively regulating attention and message-passing heads.
    \item We characterize several structural properties of SiGMA, showing that, under suitable conditions, it inherits the expressivity of selected MPGNN heads, admits near-identity paths that are not forced to over-smooth, and enables direct non-local communication beyond message-passing neighborhoods.
    \item We demonstrate excellent empirical performance across graph-level benchmarks, including both classification and regression tasks, and obtain competitive results on several datasets.
\end{itemize}
\section{Background and Related Works}

\paragraph{Notation.} Following standard convention, we denote GNN input graphs as $G=(V,E,\mathbf{X})$  with node attributes $\mathbf{X} \in \mathbb{R}^{\vert V \vert \times m}$ and edges $E \subseteq V \times V$, where $V$ is the set of vertices of $G$. Node representations are denoted  $\mathbf{H}^{\ell}$ and are initialized by the node attributes in the input graph: $\mathbf{H}^{0}=\mathbf{X}$.

\paragraph{Message-passing graph neural networks.} Message-Passing (MP)~\citep{gori2005new, hamilton2017inductive} has become one of the most popular learning paradigms in graph learning. Many state-of-the-art GNN architectures, such as \emph{GCN}~\citep{Kipf:2017tc}, \emph{GIN}~\citep{xu2018powerful} and \emph{GAT}~\citep{velivckovic2017graph}, implement MP by iteratively updating their nodes' representations based on the representations of their neighbors. Formally, let $\mathbf{h}_v^l$ denote the representation of node $v$ at layer $l$.  Then the updated representation in layer $l+1$ (i.e., after one MP iteration) is given by 
\begin{equation}
\mathbf{h}_v^{l+1} = \phi_l \Big( \mathbf{h}_v^l,  \bigoplus_{p \in \mathcal{N}_v} \psi_l \left ( \mathbf{h}_p^l, \mathbf{h}_v^l\right)\Big)\text{.}
\end{equation}

Here, $\psi_l$ denotes a message function, and $\phi_l$ an update function (e.g., an MLP) that computes the updated node representation. We refer to the number of MP iterations as the \emph{depth} of the GNN. 

\paragraph{Graph transformers.} Several graph transformer GNN  (GT) architectures have been proposed as an alternative to MPGNNs~\citep{muller2023attending, min2022transformer}. They consist of stacked blocks of multi-head attention layers followed by 
fully-connected feed-forward networks. Formally, a single attention head in layer $l$ computes node features as
$\text{Attn}(\mathbf{H}^l) := \text{softmax}\Big( \frac{\mathbf{QK}^T}{\sqrt{d_k}} \Big) \mathbf{V}$.
Here, the matrices $\mathbf{Q} := \mathbf{H}^{l} \mathbf{W_Q}$, $\mathbf{K} := \mathbf{H}^{l} \mathbf{W_K}$, $\mathbf{V} := \mathbf{H}^{l} \mathbf{W_V}$ are linear projections
of the node features; the softmax is applied row-wise.
Multi-head attention concatenates several such single-attention heads and projects their output into the feature space of $\mathbf{H}^{l}$. By combining the above with additional residual connections and normalization, the transformer layer updates features $\mathbf{H}^{(l)}$ via
\begin{equation}
\mathbf{H}^{(l+1)} := \text{FFN}(\text{MultiHead}(\mathbf{H}^{(l)}) + \mathbf{H}^{(l)}).\end{equation}
Notable instances of GTs include Graphormer~\citep{ying2021transformers} and GraphGPS~\citep{rampavsek2022recipe}. 

\paragraph{Node-level heterogeneity.} Prior work has shown that heterogeneity is already present at the node level in graph learning tasks: even within a single graph, different nodes can vary substantially in classification difficulty. In particular, node-level performance heterogeneity has been linked to structural properties of the underlying graph, with topological quantities such as node degree providing explanatory signal for why some nodes are easier to classify than others~\cite{li2023a}. This suggests that graph neural networks do not behave uniformly across all local neighborhoods, but instead exhibit performance patterns shaped by the structural context in which each prediction is made, consistent with broader evidence that message-passing architectures are sensitive to local graph structure, bottlenecks, and neighborhood growth~\cite{Kipf:2017tc,hamilton2017inductive,alon2021on,weber2020neighborhood}. 


\paragraph{Adaptive operators and gating for GNNs.}
Dasoulas et al.~\cite{dasoulas2021learning} learn a parameterized graph shift operator that adapts a shared local propagation rule to the dataset. SiGMA instead learns node-wise combinations of multiple local message-passing and global attention heads.
To the best of our knowledge, there are only two other works that apply gating mechanisms to GNN heads: Gated-GPS applies gating to GPS \cite{gao2025gated}, but unlike our architecture, gating is applied after the full GPS layer instead of to individual message-passing and attention heads. Concurrent with us, SigGate-GT applies gating within GPS \cite{liu2026capacity}, but only to the attention head. Both these works, and ours, differ from prior applications of gating in GNNs, which focus on using gating in the message-passing mechanism itself (most prominently in \cite{bresson2017residual}).
\section{Graph-Level Heterogeneity}
\label{sec:graph-level-heterogeneity}

We study \emph{graph-level heterogeneity}: variation across graphs within the same dataset in prediction difficulty for a given model architecture. Our central observation is not merely that some graphs
are universally easier than others. Rather, graph difficulty emerges from the interaction between the input and the model: graphs that are difficult for one message-passing mechanism can be comparatively easy for another, and different graphs can favor different depths or combinations of local and non-local computation. This motivates architectures whose computation can adapt across graph instances.

\subsection{Heterogeneity Profiles}

\begin{figure}[h]
    \centering
    \includegraphics[width=0.24\linewidth]{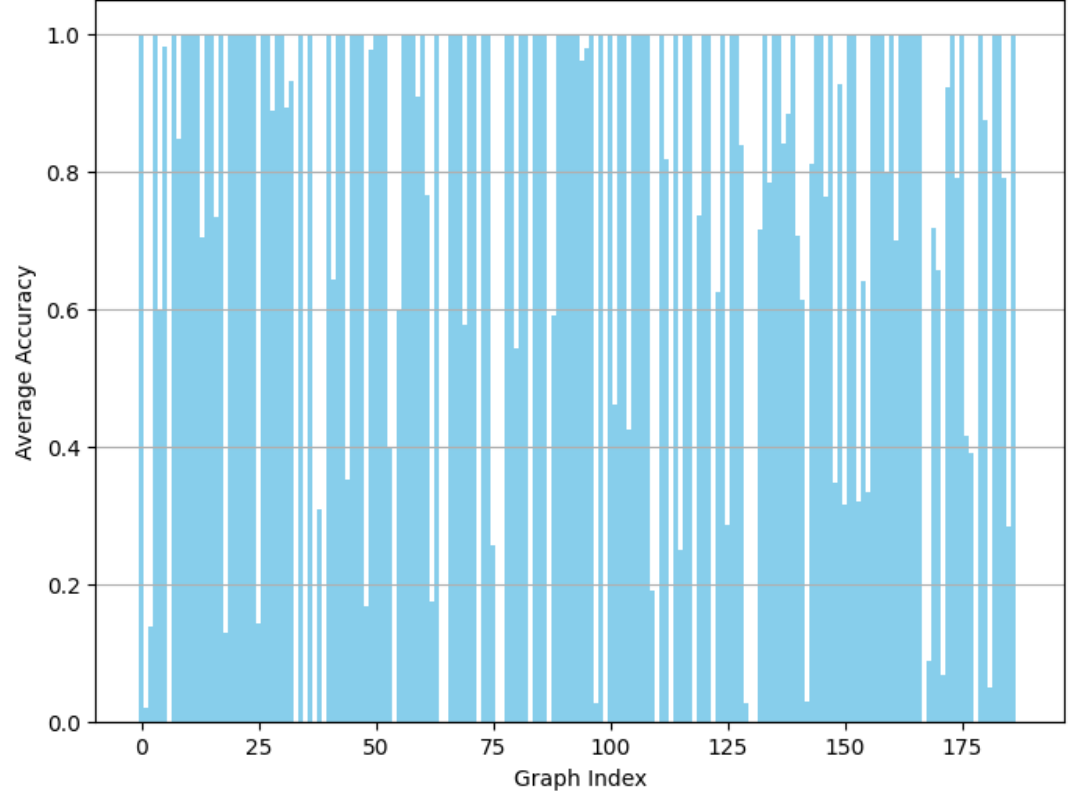}
    \hfill
    \includegraphics[width=0.24\linewidth]{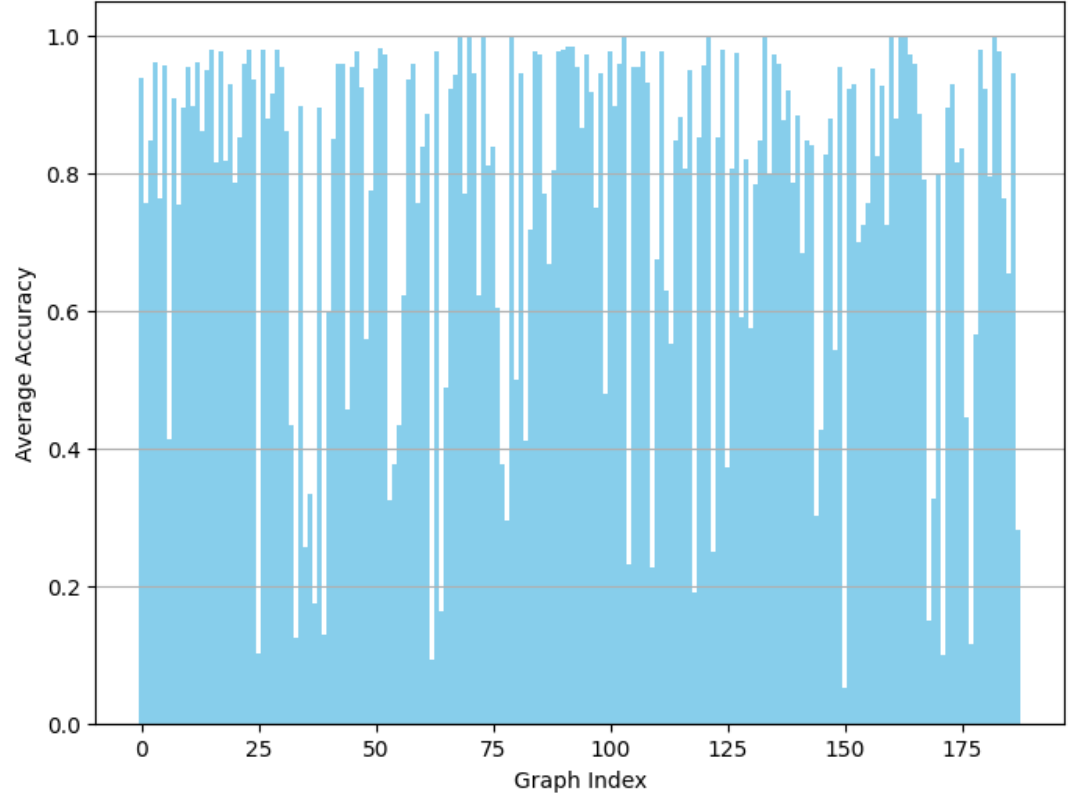}
    \hfill
    \includegraphics[width=0.24\linewidth]{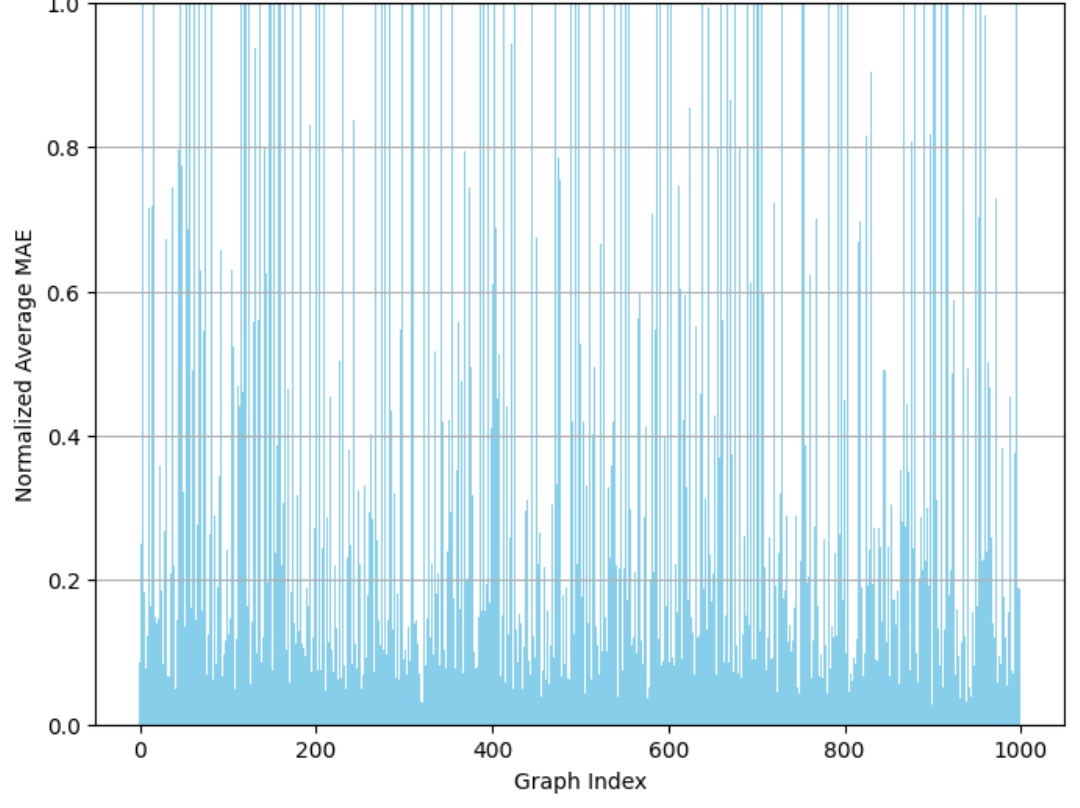}
    \hfill
    \includegraphics[width=0.24\linewidth]{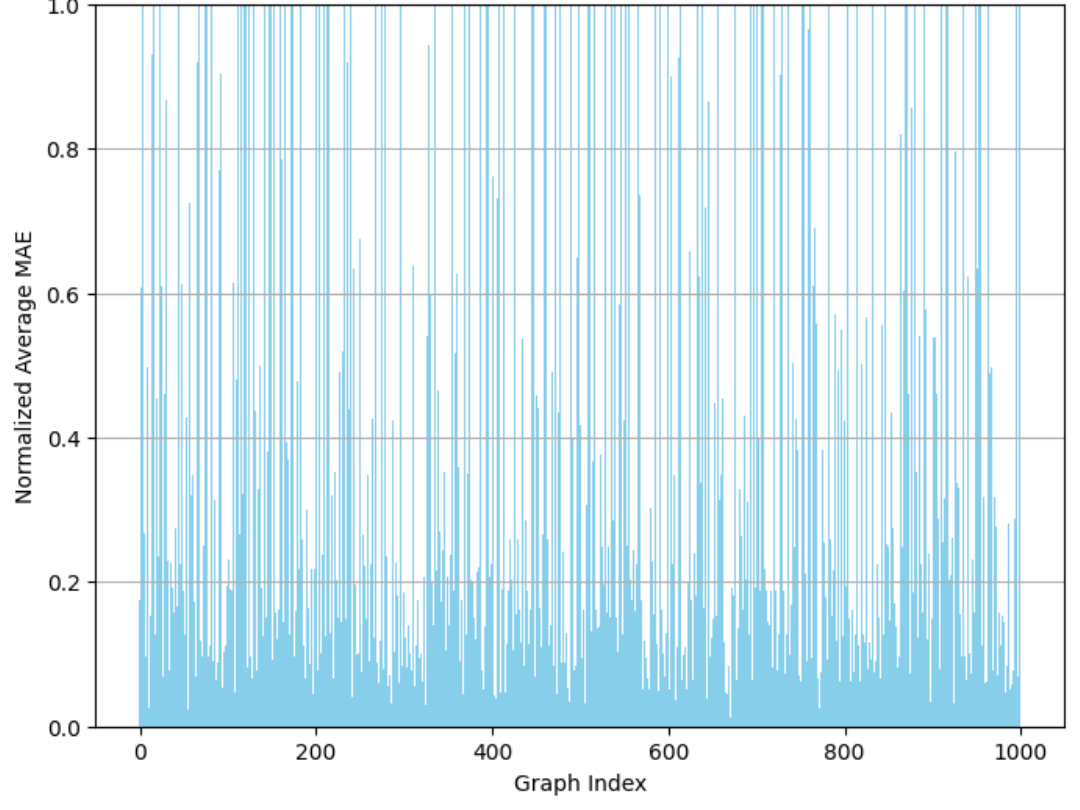}
    \caption{\textbf{Examples of heterogeneity profiles.} From left to right: GCN on MUTAG, GIN on MUTAG, GCN on ZINC, and GIN on ZINC. Profiles like these allow us to study optimal message-passing mechanisms and model depths at the level of the individual graph.}
    \label{fig:hetero_profiles_main}
\end{figure}

To measure graph-level difficulty, we construct heterogeneity profiles from
repeated held-out predictions. For each dataset, we repeatedly sample
\(50/25/25\) train/validation/test splits, train the model for a given number of epochs, select the checkpoint with the best validation performance, and record its prediction on every graph in the test set. We continue until every graph has
appeared in the test set at least \(100\) times. For classification, the score of a graph is its average probability of being classified correctly; for regression, it is its average absolute prediction error. The complete construction is given in App.~\ref{app:heterogeneity-profile-construction}. Fig.~\ref{fig:hetero_profiles_main} plots example profiles for GCN and GIN on the MUTAG and ZINC datasets. 

The resulting profiles span a broad range of per-graph performance. Some graphs are predicted correctly in nearly every trial, whereas others remain difficult
across independently sampled training sets and model initializations. This
phenomenon appears across graph classification and regression datasets, as
well as across message-passing and hybrid architectures. The profiles also remain qualitatively non-uniform under architectural and hyperparameter variations, including changes in model family, width, depth, and learning rate; see App.~\ref{app:heterogeneity-robustness}.

Heterogeneity is already visible during optimization. Models first learn a
subset of graphs, causing instance-level variation to increase, and only later
resolve additional examples. However, the order in which graphs are learned is
not universal. Some graphs that are difficult under GCN message passing are learned readily by GIN, and conversely, for example. Thus, there is both a shared
component of difficulty and an architecture-dependent component. Training
trajectories illustrating this interaction are provided in App.~\ref{app:training-dynamics}.

Graph topology alone does not explain most of the observed variation.
Sparse regressions using common topological statistics, including graph size and connectivity descriptors, attain very low explanatory
power on most datasets (App.~\ref{app:graph-statistics}). A richer measure that combines topology and node attributes, the Tree Mover's Distance \cite{chuang2022tree}, is more informative:
graphs whose nearest structural and feature-space neighbors belong to competing
classes tend to be harder to predict. We summarize this analysis in
App.~\ref{app:tmd-class-separation}. Together, these results indicate that
heterogeneity cannot generally be reduced to a single coarse graph statistic;
it depends on the relationship between graph structure, attributes, labels,
and model inductive bias.

\subsection{Graph-Dependent Computational Preferences}

Using the heterogeneity profiles introduced above, we can show that graphs within the same dataset prefer different message-passing mechanisms. We train models using GCN \cite{Kipf:2017tc}, GIN \cite{xu2018powerful}, and GatedGCN \cite{bresson2017residual} operators and assign each graph to the mechanism under which it achieves its best repeated held-out performance. All models use 4 layers, hidden dimension 64, ReLU activation, dropout 0.1, global mean pooling, and an MLP classifier head; operator-specific architectural choices are limited to those required by the corresponding message-passing rule.

\begin{table}[h]
\centering
\caption{Percentage of graphs that achieve their best average held-out
performance under each message-passing mechanism. No operator is uniformly
preferred across all graphs in a dataset.}
\label{tab:mp-preferences}
\scriptsize
\setlength{\tabcolsep}{4pt}
\begin{tabular}{lrrrrrr}
\toprule
MP mechanism
& MUTAG & ENZYMES & PROTEINS & IMDB & COLLAB & REDDIT \\
\midrule
GCN
& 29.2 & 30.1 & 51.0 & 25.9 & 31.2 & 30.0 \\
GIN
& 40.3 & 31.2 & 27.0 & 62.6 & 49.0 & 65.0 \\
GatedGCN
& 30.5 & 38.7 & 22.1 & 11.5 & 19.8 & 5.0 \\
\bottomrule
\end{tabular}
\vspace{-2mm}
\end{table}

Tab.~\ref{tab:mp-preferences} shows that the preferred local operator varies
substantially across graphs. GIN is selected most often on several datasets,
but it is far from universally optimal. These differences are consistent with the
distinct inductive biases of the operators. GCN performs smoothing-based
neighborhood averaging, GIN emphasizes discriminative multiset aggregation, and
GatedGCN selectively regulates neighboring information. The results suggest
that choosing a message-passing operator solely at the dataset level imposes
one local inductive bias on graphs whose useful local computations may differ.

We perform an analogous analysis over models with \(2\), \(4\), and
\(8\) message-passing layers. Depth determines both the nominal receptive field
and the number of transformations applied to node features. A shallow model may
therefore be sufficient when the prediction-relevant signal is local, whereas
other graphs may benefit from larger receptive fields. At the same time,
unnecessary propagation can introduce irrelevant information or increase
susceptibility to over-smoothing.

\begin{table}[h]
\centering
\caption{Percentage of graphs that achieve their best average
performance with each number of GCN message-passing layers. Different graphs within the same dataset favor different model depths.}
\label{tab:depth-preferences}
\scriptsize
\setlength{\tabcolsep}{4pt}
\begin{tabular}{lrrrrrr}
\toprule
Number of layers
& MUTAG & ENZYMES & PROTEINS & IMDB & COLLAB & REDDIT \\
\midrule
2
& 37.5 & 23.6 & 42.1 & 44.6 & 37.6 & 16.6 \\
4
& 25.5 & 48.1 & 26.2 & 28.1 & 29.4 & 37.2 \\
8
& 37.0 & 28.2 & 31.7 & 27.3 & 32.9 & 46.1 \\
\bottomrule
\end{tabular}
\end{table}

Tab.~\ref{tab:depth-preferences} confirms that no single depth is preferred by
all graphs. PROTEINS and IMDB contain many graphs that favor shallow models,
whereas on REDDIT the largest group favors eight layers, for example. Importantly, this variation does not imply that every graph has a fixed
intrinsic ``correct depth.'' Rather, the broad within-dataset variation shows that a fixed-depth model applies one computational scale to examples that respond differently to neighborhood expansion.

Overall, the evidence points to two forms of graph-level heterogeneity which we exploit in the next section: variation in the preferred local aggregation mechanism and variation in useful computational depth. This motivates SiGMA, which places heterogeneous message-passing and attention heads within the same layer and learns to regulate their contributions for a given input graph.
\section{Model and Design Space}

\subsection{Model Definition}

Fig.~\ref{fig:sigma_architecture} shows the entire SiGMA architecture. At layer $\ell$, SiGMA takes node representations $\mathbf{h}_i^\ell\in\mathbb{R}^d$ and first applies normalization $\widehat{\mathbf{ h}}_i^\ell=\operatorname{Norm}(\mathbf{h}_i^\ell)$. The layer contains a set of attention heads $\mathcal{H}_{\mathrm{att}}^\ell$ and a set of message-passing heads $\mathcal{H}_{\mathrm{mp}}^\ell$ (both potentially empty). Each head produces a $d_h$-dimensional output and is modulated by a learned node-wise sigmoid gate before the heads are combined.

\paragraph{Gated attention heads.} For an attention head $a\in\mathcal{H}_{\mathrm{att}}^\ell$, the only departure from standard attention is the gate, which is computed jointly
with the query - in this exposition, we use headwise gating \cite{qiu2026gated} while the exposition using elementwise gating is in App.~\ref{app:element-wise-gating}:
\begin{equation}
\mathbf{W}_{qg}^a\widehat{ \mathbf{h}}_i^\ell
=
\left[
\mathbf{q}_i^a \;\Vert\; g_i^{\mathrm{att},a}
\right] \in \mathbb{R}^{d_h+1},
\qquad
\mathbf{k}_j^a=\mathbf{W}_k^a\widehat{ \mathbf{h}}_j^\ell \in \mathbb{R}^{d_h},
\qquad
\mathbf{v}_j^a=\mathbf{W}_v^a\widehat{ \mathbf{h}}_j^\ell \in \mathbb{R}^{d_h}.
\end{equation}
The head output is
\begin{equation}
\widetilde{\mathbf{a}}_i^{\mathrm{att},a}
=
\sigma(g_i^{\mathrm{att},a})
\cdot
\sum_{j\in V}
\alpha_{ij}^a \mathbf{v}_j^a \in \mathbb{R}^{d_h},
\qquad
\alpha_{ij}^a
=
\operatorname{softmax}_{j\in V}
\left(
(\mathbf{q}_i^a)^\top \mathbf{k}_j^a/\sqrt{d_h}+\mathbf{B}_{ij}
\right),
\end{equation}
where $\mathbf{B}_{ij}$ masks disallowed attention entries, including attention between different graphs in a batch.

\paragraph{Gated message-passing heads.} For a message-passing head $p\in\mathcal{H}_{\mathrm{mp}}^\ell$, SiGMA computes a head-specific representation and gate,
\begin{equation}
\mathbf{W}_{hg}^p\widehat{ \mathbf{h}}_i^\ell
=
\left[
\mathbf{u}_i^p \;\Vert\; g_i^{\mathrm{mp},p}
\right]  \in \mathbb{R}^{d_h +1},
\end{equation}
then applies the chosen graph operator $\Phi_p$ and gates its
output:
\begin{equation}
\widetilde{\mathbf{a}}_i^{\mathrm{mp},p}
=
\sigma(g_i^{\mathrm{mp},p})
\cdot
\Phi_p\!\left(\{\mathbf{u}_j^p\}_{j\in V},E\right)_i  \in \mathbb{R}^{d_h}.
\end{equation}

The gated attention and message-passing heads are concatenated, projected,
and added through a residual update:
\begin{equation}
\label{eq:sigma_update}
\mathbf{\Delta}_i^\ell
=
\mathbf{W}_{\mathrm{out}}^\ell
\left[
\big\Vert_{a\in\mathcal{H}_{\mathrm{att}}^\ell}
\widetilde{\mathbf{a}}_i^{\mathrm{att},a}
\;\Big\Vert\;
\big\Vert_{p\in\mathcal{H}_{\mathrm{mp}}^\ell}
\widetilde{\mathbf{a}}_i^{\mathrm{mp},p}
\right],
\qquad
\mathbf{h}_i^{\ell+1}=\mathbf{h}_i^\ell+\mathbf{\Delta}_i^\ell .
\end{equation}
The gated MPNN and attention block may be followed by the usual feed-forward, normalization, and dropout components. In practice, we use a variety of MPGNN architectures as the MPGNN heads and outline common architectures in App.~\ref{app:mpgnn-heads}. 

\paragraph{Intuition.} The gates in this architecture are node-wise sigmoid modulators attached to individual heads. They allow the model to regulate, for each node and each layer, how strongly a given attention or message-passing mechanism contributes to the update. If a graph contains regions whose prediction-relevant signal is primarily local, the corresponding nodes can emphasize message-passing heads; if long-range interactions are useful, they can retain larger attention-head activations. Moreover, because the gated outputs enter through a residual update, layers whose heads are suppressed
make only a small change to the representation. In this sense, although the nominal number of layers is fixed, the model can learn an input-dependent effective depth: different graphs, and even different nodes within a graph, may use different subsets of heads and different numbers of active layers.

\paragraph{Design space.} This formulation exposes a flexible design space in which the practitioner can choose both the number and type of heads in each hybrid block. The message-passing heads may instantiate different local graph operators, such as GCN, GIN, GINE, GraphSAGE, GAT, or domain-specific variants, while the attention heads may use full-graph attention, graph-masked attention, or structurally biased attention mechanisms. The relative number of attention and message-passing heads controls the model's local--global capacity, ranging from purely local MPGNN-like architectures to fully global graph transformers, with hybrid configurations in between. Since each head is gated separately, heterogeneous collections of heads can be included without requiring all of them to be equally useful for every graph, node, or layer. 

\subsection{Inherited Architectural Properties}

Our contribution is primarily architectural and empirical rather than theoretical. Nevertheless, the gated hybrid architecture has several useful structural properties inherited from its constituent mechanisms. Our analysis considers the stylized update \eqref{eq:sigma_update} and treats components such as activation functions, normalization, and feed-forward blocks as optional. We present the main results here and defer all proofs to App.~\ref{app:proofs}.

\begin{proposition}[Expressivity inherited from message passing]
Consider an $L$-layer SiGMA network whose message-passing heads include,
in each layer $\ell$, a head implementing the $\ell$-th update of some
message-passing GNN $\mathcal{M}$. Then, under suitable conditions, every graph function representable by $\mathcal{M}$
is also representable by SiGMA. Consequently, SiGMA is at least as
expressive as the most expressive message-passing architecture satisfying these conditions.
\end{proposition}

This is a model-class containment property that applies broadly to architectures capable of instantiating the corresponding message-passing operators. For example, if the selectable message-passing heads include a GIN head with injective aggregation and injective graph readout satisfying these conditions, then SiGMA inherits the standard $1$-WL expressivity guarantee of GIN as a special case. Similarly, SiGMA can exceed $1$-WL when given positional and structural encodings that have been shown to
increase the expressivity of MPGNNs and GTs \cite{pellegrin2026higher, fesser2024effective}, although we do not formalize this here.

Oversmoothing can be formalized through the graph Dirichlet energy. For node representations $\mathbf{H}^\ell\in\mathbb{R}^{|V|\times d}$ at layer $\ell$, define
\begin{equation}
\mathcal{E}_G(\mathbf{H}^\ell)
=
\frac{1}{2}
\sum_{(i,j)\in E}
\|\mathbf{h}_i^\ell-\mathbf{h}_j^\ell\|_2^2 .
\end{equation}
Small Dirichlet energy means that adjacent nodes have similar representations, and complete collapse to a constant representation gives $\mathcal{E}_G(\mathbf{H}^\ell)=0$. Thus, oversmoothing corresponds to the (exponentially fast) decay of $\mathcal{E}_G(\mathbf{H}^\ell)$ toward zero as model depth increases \cite{kiani2024unitary}.\\

\begin{proposition}[No forced oversmoothing]
Consider an $L$-layer SiGMA network. Under boundedness assumptions on the ungated head outputs and output projections, and assuming the gate logits can be made arbitrarily negative (e.g., via a gate bias), for every graph $G$, depth $L$, and $\varepsilon>0$, there exist gate values such that $\max_{i\in V}\|\mathbf{h}_i^L-\mathbf{h}_i^0\|_2\leq\varepsilon$.
\end{proposition}

Consequently, $\mathcal{E}_G(\mathbf{H}^L)$ can be made arbitrarily close to $\mathcal{E}_G(\mathbf{H}^0)$, so increasing the nominal depth of SiGMA does not by itself force Dirichlet-energy collapse or oversmoothing. This near-identity construction follows from the gated residual updates and applies more generally to architectures equipped with this mechanism.

Over-squashing can be formalized through the sensitivity of one node representation to another node's input features \cite{fesser2024mitigating}. For node representations $\mathbf{h}_i^\ell$ and input features $\mathbf{x}_j$, define the pairwise influence
\begin{equation}
S_{ij}^\ell
=
\left\|
\frac{\partial \mathbf{h}_i^\ell}{\partial \mathbf{x}_j}
\right\| .
\end{equation}
Small $S_{ij}^\ell$ means that information from node $j$ has little effect on the representation of node $i$. In purely local message-passing networks, $S_{ij}^\ell=0$ whenever $\operatorname{dist}_G(i,j)>\ell$, and even within the $\ell$-hop neighborhood these sensitivities may decay sharply across graph bottlenecks. Over-squashing refers to this compression of long-range information into vanishing influence on node representations. \\

\begin{proposition}[No forced over-squashing]
Consider a SiGMA layer containing at least one attention head whose mask allows the entry $(i, j), i \neq j$. Under some mild assumptions on the weight matrices, $\mathbf{h}_i^{\ell+1}$ can depend directly on $\mathbf{h}_j^\ell$, and hence the influence $S_{ij}^{\ell+1}$ need not vanish even when $\operatorname{dist}_G(i,j)>\ell+1$.
\end{proposition}

Hence, unlike purely local one-hop message passing, SiGMA is not forced to transmit long-range information only through graph paths and can use attention heads to bypass local bottlenecks associated with over-squashing. This property is not specific to SiGMA: direct communication between non-adjacent nodes is shared by graph transformers and other architectures using full-graph attention. SiGMA inherits this property whenever such an attention head is included.

\section{Experimental Evaluation}

We evaluate SiGMA on a diverse collection of graph-classification and -regression benchmarks. Our evaluation is designed to answer three questions. First, how does SiGMA compare with existing methods across diverse graph-level benchmarks? Second, how does adaptive gating affect the performance and consistency of the hybrid architecture? Third, do heterogeneous message-passing heads provide an advantage over homogeneous head configurations?



\begin{table}[h]
\centering
\caption{Test performance on five benchmarks from~\cite{JMLR:Dwivedi}. Shown is the mean $\pm$ s.d. of 5 runs with different random seeds. The baseline results were obtained from their respective original papers. GCN$^{+}$, GIN$^{+}$, Exphormer, GRIT, GraphGPS, EGT, and GatedGCN all use edge features. Accuracies are reported in percent. The top $\first{1^{\mathrm{st}}}$, $\second{2^{\mathrm{nd}}}$, $\third{3^{\mathrm{rd}}}$, and $\fourth{4^{\mathrm{th}}}$ results are highlighted.} 
\label{tab:benchmark-results-1}
\vspace{-0.5em}
\scriptsize
\setlength{\tabcolsep}{3pt}
\renewcommand{\arraystretch}{1.05}
\begin{tabularx}{\textwidth}{@{}lYYYYY@{}}
\toprule
& ZINC & MNIST & CIFAR10 & PATTERN & CLUSTER \\
Metric & MAE $\downarrow$ & Acc. $\uparrow$ & Acc. $\uparrow$ & Acc. $\uparrow$ & Acc. $\uparrow$ \\
\midrule



GT~\cite{dwivedi2022graph}
& $0.226{\scriptstyle\pm0.014}$
& $90.831{\scriptstyle\pm0.161}$
& $59.753{\scriptstyle\pm0.293}$
& $84.808{\scriptstyle\pm0.068}$
& $73.169{\scriptstyle\pm0.622}$ \\




EGT~\cite{hussain2022global}
& $0.108{\scriptstyle\pm0.009}$
& $98.173{\scriptstyle\pm0.087}$
& $68.702{\scriptstyle\pm0.409}$
& $86.821{\scriptstyle\pm0.020}$
& \second{$79.232{\scriptstyle\pm0.348}$} \\

GraphGPS~\cite{rampavsek2022recipe}
& $0.070{\scriptstyle\pm0.004}$
& $98.051{\scriptstyle\pm0.126}$
& $72.298{\scriptstyle\pm0.356}$
& $86.685{\scriptstyle\pm0.059}$
& $78.016{\scriptstyle\pm0.180}$ \\







Graph ViT/MLP-Mixer~\cite{he2023generalization}
& $0.073{\scriptstyle\pm0.001}$
& \fourth{$98.460{\scriptstyle\pm0.090}$}
& $73.960{\scriptstyle\pm0.330}$
& --
& -- \\

Exphormer~\cite{shirzad2023exphormer}
& --
& $98.414{\scriptstyle\pm0.038}$
& $74.754{\scriptstyle\pm0.194}$
& $86.734{\scriptstyle\pm0.008}$
& -- \\

GRIT~\cite{ma2023graph}
& \second{$0.059{\scriptstyle\pm0.002}$}
& $98.108{\scriptstyle\pm0.111}$
& $76.468{\scriptstyle\pm0.881}$
& \second{$87.196{\scriptstyle\pm0.076}$}
& \first{$80.026{\scriptstyle\pm0.277}$} \\

GRED~\cite{ding2023recurrent}
& $0.077{\scriptstyle\pm0.002}$
& $98.383{\scriptstyle\pm0.012}$
& \third{$76.855{\scriptstyle\pm0.185}$}
& $86.759{\scriptstyle\pm0.020}$
& $78.495{\scriptstyle\pm0.103}$ \\

GEAET~\cite{liang2024graph}
& --
& \third{$98.513{\scriptstyle\pm0.086}$}
& \fourth{$76.634{\scriptstyle\pm0.427}$}
& $86.993{\scriptstyle\pm0.026}$
& -- \\

TIGT~\cite{choi2024topology}
& \first{$0.057{\scriptstyle\pm0.002}$}
& $98.231{\scriptstyle\pm0.132}$
& $73.963{\scriptstyle\pm0.361}$
& $86.681{\scriptstyle\pm0.062}$
& $78.025{\scriptstyle\pm0.223}$ \\


GMN~\cite{behrouz2024graph}
& --
& $98.391{\scriptstyle\pm0.182}$
& $74.560{\scriptstyle\pm0.381}$
& \third{$87.090{\scriptstyle\pm1.260}$}
& -- \\

Graph-Mamba~\cite{wang2024graph}
& --
& $98.420{\scriptstyle\pm0.080}$
& $73.700{\scriptstyle\pm0.340}$
& $86.710{\scriptstyle\pm0.050}$
& $76.800{\scriptstyle\pm0.360}$ \\
\midrule
GCN$^{+}$ \cite{luo2025can}
& $0.076{\scriptstyle\pm0.009}$
& $98.382{\scriptstyle\pm0.095}$
& $69.824{\scriptstyle\pm0.413}$
& $87.021{\scriptstyle\pm0.095}$
& $77.109{\scriptstyle\pm0.872}$ \\

GIN$^{+}$ \cite{luo2025can}
& \fourth{$0.065{\scriptstyle\pm0.004}$}
& $98.285{\scriptstyle\pm0.103}$
& $69.592{\scriptstyle\pm0.287}$
& $86.842{\scriptstyle\pm0.048}$
& $74.794{\scriptstyle\pm0.213}$ \\

GatedGCN$^{+}$ \cite{luo2025can}
& $0.077{\scriptstyle\pm0.005}$
& \first{$98.712{\scriptstyle\pm0.137}$}
& \second{$77.218{\scriptstyle\pm0.381}$}
& \fourth{$87.029{\scriptstyle\pm0.037}$}
& \third{$79.128{\scriptstyle\pm0.235}$}\\
\midrule
SiGMA (Ours) & \third{$0.060{\scriptstyle\pm 0.003}$} & \second{$98.628{\scriptstyle\pm 0.105}$} & \first{$79.528{\scriptstyle\pm 0.180}$} & \first{$87.395{\scriptstyle\pm0.194}$} & 
\fourth{$79.087{\scriptstyle\pm0.158}$}\\
\bottomrule
\end{tabularx} 
\vspace{-0.5em}
\end{table}

\subsection{Experimental Setup}

We consider two groups of graph-level benchmarks. Following ~\cite{luo2025can}, we use “graph-level benchmarks” to denote datasets whose fundamental examples, splits, and training units are individual graphs. The first comprises five widely used datasets from the benchmark suite of Dwivedi et al.~\cite{JMLR:Dwivedi}: ZINC, MNIST, CIFAR10, PATTERN, and CLUSTER. These datasets span molecular regression, image-derived graph classification, and synthetic pattern recognition. The second group comprises Peptides-func, Peptides-struct, PascalVOC-SP, COCO-SP, and MalNet-Tiny~\cite{dwivedi2022LRGB}, covering long-range molecular, superpixel, and program-analysis tasks. For each dataset, we use the official train, validation, and test splits and select the checkpoint with the best validation performance. Unless stated otherwise, we report the mean and standard deviation over multiple random seeds, following the protocol of the corresponding dataset. We compare SiGMA with message-passing, transformer-based, and hybrid architectures. We tune the number of layers, hidden dimension, attention heads, number and type of message-passing heads, dropout, learning rate, and weight decay, and report the resulting hyperparameters in App.~\ref{app:experimental-details}. Dataset statistics are provided in App.~\ref{app:datasets}, and additional results on the TU datasets can be found in App.~\ref{app:additional-results-TU}.

\subsection{Main Results}

Table~\ref{tab:benchmark-results-1} reports results on ZINC, MNIST, CIFAR10, PATTERN, and CLUSTER. On MNIST, SiGMA obtains an accuracy of $98.628 \pm 0.105$, outperforming all reported baselines except GatedGCN$^{+}$. SiGMA achieves accuracies of $79.528 \pm 0.180$ on CIFAR10 and $87.395 \pm 0.194$ on PATTERN, surpassing the graph transformer and hybrid baselines reported in the table, including GRIT, GRED, GEAET, and Graph-Mamba. It also remains competitive on CLUSTER, with an accuracy of $79.087 \pm 0.158$, and on ZINC, where it attains an MAE of $0.060 \pm 0.003$. These results demonstrate that SiGMA performs consistently well across tasks with different graph structures and objectives.

Table~\ref{tab:lrgb-results} reports results on Peptides-func, Peptides-struct, PascalVOC-SP, COCO-SP, and MalNet-Tiny. SiGMA achieves an average precision of $0.7080 \pm 0.0063$ on Peptides-func and an MAE of $0.2441 \pm 0.0017$ on Peptides-struct, demonstrating competitive performance on both long-range molecular tasks. Its strongest relative results occur on the superpixel benchmarks, where it reaches F1 scores of $0.4687 \pm 0.0070$ on PascalVOC-SP and $0.4215 \pm 0.0055$ on COCO-SP, outperforming the strongest graph transformer considered here. On MalNet-Tiny, SiGMA obtains an accuracy of $0.9400 \pm 0.0038$, remaining competitive with existing methods while trailing GatedGCN$^{+}$. Taken together, the results show that SiGMA is competitive across all five benchmarks and particularly effective on tasks that require local graph structure to be integrated with long-range information.

\begin{table}[h] 
\centering 
\caption{Test performance on five datasets from Long-Range Graph Benchmarks (LRGB)~\cite{dwivedi2022LRGB} and MalNet-Tiny~\cite{freitas2020large}. The baseline results were obtained from their respective original papers.} 
\label{tab:lrgb-results} 
\vspace{-0.5em} 
\scriptsize 
\setlength{\tabcolsep}{3pt} 
\renewcommand{\arraystretch}{1.05} 
\begin{tabularx}{\textwidth}{@{}lYYYYY@{}} 
\toprule & Peptides-func & Peptides-struct & PascalVOC-SP & COCO-SP & MalNet-Tiny \\ 
Metric & Avg. Prec. $\uparrow$ & MAE $\downarrow$ & F1 score $\uparrow$ & F1 score $\uparrow$ & Acc. $\uparrow$ \\ 
\midrule GT~\cite{dwivedi2022graph} & $0.6326{\scriptstyle\pm0.0126}$ & $0.2529{\scriptstyle\pm0.0016}$ & $0.2694{\scriptstyle\pm0.0098}$ & $0.2618{\scriptstyle\pm0.0031}$ & -- \\ 
SAN~\cite{kreuzer2021rethinking} & $0.6439{\scriptstyle\pm0.0075}$ & $0.2545{\scriptstyle\pm0.0012}$ & $0.3230{\scriptstyle\pm0.0039}$ & $0.2592{\scriptstyle\pm0.0158}$ & -- \\ 
GraphGPS~\cite{rampavsek2022recipe} & $0.6535{\scriptstyle\pm0.0041}$ & $0.2500{\scriptstyle\pm0.0005}$ & $0.3748{\scriptstyle\pm0.0109}$ & $0.3412{\scriptstyle\pm0.0044}$ & $0.9350{\scriptstyle\pm0.0041}$ \\ GraphGPS~\cite{tonshoff2023where} & $0.6534{\scriptstyle\pm0.0091}$ & $0.2509{\scriptstyle\pm0.0014}$ & \third{$0.4440{\scriptstyle\pm0.0065}$} & \third{$0.3884{\scriptstyle\pm0.0055}$} & -- \\ 
DRew~\cite{gutteridge2023drew} & \second{$0.7150{\scriptstyle\pm0.0044}$} & $0.2536{\scriptstyle\pm0.0015}$ & $0.3314{\scriptstyle\pm0.0024}$ & -- & -- \\ 
Exphormer~\cite{shirzad2023exphormer} & $0.6258{\scriptstyle\pm0.0092}$ & $0.2512{\scriptstyle\pm0.0025}$ & $0.3446{\scriptstyle\pm0.0064}$ & $0.3430{\scriptstyle\pm0.0108}$ & \third{$0.9402{\scriptstyle\pm0.0021}$} \\ 
GEAET~\cite{liang2024graph} & $0.6485{\scriptstyle\pm0.0035}$ & $0.2547{\scriptstyle\pm0.0009}$ & $0.3933{\scriptstyle\pm0.0027}$ & $0.3219{\scriptstyle\pm0.0052}$ & -- \\ 
GECO~\cite{sancak2025scalable} & $0.6975{\scriptstyle\pm0.0025}$ & $0.2464{\scriptstyle\pm0.0009}$ & $0.4210{\scriptstyle\pm0.0080}$ & $0.3320{\scriptstyle\pm0.0032}$ & -- \\ 
Graph-Mamba~\cite{wang2024graph} & $0.6739{\scriptstyle\pm0.0087}$ & $0.2478{\scriptstyle\pm0.0016}$ & $0.4191{\scriptstyle\pm0.0126}$ & \second{$0.3960{\scriptstyle\pm0.0175}$} & $0.9340{\scriptstyle\pm0.0027}$ \\ GSSC~\cite{huang2024can} & \third{$0.7081{\scriptstyle\pm0.0062}$} & $0.2459{\scriptstyle\pm0.0020}$ & \second{$0.4561{\scriptstyle\pm0.0039}$} & -- & \second{$0.9406{\scriptstyle\pm0.0064}$} \\ 
\midrule
GCN$^{+}$ \cite{luo2025can} & \first{$0.7261{\scriptstyle\pm0.0067}$} & \first{$0.2421{\scriptstyle\pm0.0016}$} & $0.3357{\scriptstyle\pm0.0087}$ & $0.2733{\scriptstyle\pm0.0041}$ & $0.9354{\scriptstyle\pm0.0045}$ \\
GIN$^{+}$ \cite{luo2025can} & $0.7059{\scriptstyle\pm0.0089}$ & \second{$0.2429{\scriptstyle\pm0.0019}$} & $0.3189{\scriptstyle\pm0.0105}$ & $0.2483{\scriptstyle\pm0.0046}$ & $0.9325{\scriptstyle\pm0.0040}$ \\
GatedGCN$^{+}$ \cite{luo2025can} & $0.7006{\scriptstyle\pm0.0033}$ & \third{$0.2431{\scriptstyle\pm0.0020}$} & \fourth{$0.4263{\scriptstyle\pm0.0057}$} & \fourth{$0.3802{\scriptstyle\pm0.0015}$} & \first{$0.9460{\scriptstyle\pm0.0057}$} \\
\midrule
SiGMA (Ours) & \fourth{$0.7080{\scriptstyle\pm0.0063}$} &
\fourth{$0.2441{\scriptstyle\pm 0.0017}$} & \first{$0.4687{\scriptstyle\pm 0.0070}$} & \first{$0.4215{\scriptstyle\pm 0.0055}$} & 
\fourth{$0.9400{\scriptstyle\pm  0.0038}$} \\
\bottomrule 
\end{tabularx} 
\vspace{-0.5em} 
\end{table}

\subsection{Ablation Results}

Having established that SiGMA is competitive across a diverse collection of graph-level benchmarks, we next examine the architectural choices underlying its performance. For each dataset, we fix the hyperparameter configuration of the best SiGMA model (including depth, hidden dimension, learning rate, gate type, and head counts) and retrain six variants over five random seeds. MP-only replaces all attention heads with message-passing heads; Attention-only makes the converse replacement. The remaining variants retain SiGMA's hybrid head configuration but disable gating for the message-passing heads, the attention heads, or all heads (which we call Mixture of Message-Passing and Attention (MMA), see App.~\ref{app:MMA}), or place gating on the whole layer (SiGMA-lite, see App.~\ref{app:sigma-lite}). All other training settings match the corresponding SiGMA configuration.

\begin{table}[h]
\vspace{-2mm}
\centering
\caption{Architectural ablations across seven graph-level benchmarks. We compare SiGMA with variants that remove message passing or attention, or disable gating for one or both head types. The final column reports the average rank across datasets. Results are mean $\pm$ s.d. over five seeds. We do not run SiGMA with ungated attention for CLUSTER, as our best model has no attention head.}
\label{tab:architecture-ablations}
\resizebox{\textwidth}{!}{
\begin{tabular}{lcccccccc}
\toprule
Model
& Peptides-func $\uparrow$
& Peptides-struct $\downarrow$
& PascalVOC-SP $\uparrow$
& MNIST $\uparrow$
& CIFAR10 $\uparrow$
& PATTERN $\uparrow$
& CLUSTER $\uparrow$
& Avg.\ rank $\downarrow$ \\
\midrule
MP-only
& $0.6965{\scriptstyle\pm0.0040}$
& $0.2476{\scriptstyle\pm0.0012}$
& $0.4203{\scriptstyle\pm0.0113}$
& $98.600{\scriptstyle\pm0.049}$
& $75.678{\scriptstyle\pm0.372}$
& $87.137{\scriptstyle\pm0.056}$
& $\mathbf{79.087{\scriptstyle\pm0.158}}$
& $4.79$ \\

Attention-only
& $0.6998{\scriptstyle\pm0.0127}$
& $0.2673{\scriptstyle\pm0.0087}$
& $0.3908{\scriptstyle\pm0.0025}$
& $97.650{\scriptstyle\pm0.187}$
& $61.086{\scriptstyle\pm0.971}$
& $87.293{\scriptstyle\pm0.168}$
& $31.552{\scriptstyle\pm0.051}$
& $6.29$ \\

SiGMA, ung. MP
& $\mathbf{0.7087{\scriptstyle\pm0.0068}}$
& $0.2494{\scriptstyle\pm0.0023}$
& $0.4579{\scriptstyle\pm0.0067}$
& $\mathbf{98.642{\scriptstyle\pm0.108}}$
& $79.520{\scriptstyle\pm0.348}$
& $87.404{\scriptstyle\pm0.150}$
& $79.014{\scriptstyle\pm0.110}$
& $3.21$ \\

SiGMA, ung. att.
& $0.6898{\scriptstyle\pm0.0084}$
& $\mathbf{0.2438{\scriptstyle\pm0.0021}}$
& $0.4590{\scriptstyle\pm0.0104}$
& $98.576{\scriptstyle\pm0.076}$
& $79.754{\scriptstyle\pm0.339}$
& $87.330{\scriptstyle\pm0.210}$
& --
& $3.83$ \\
SiGMA-lite 
& $0.7019{\scriptstyle\pm0.0043}$
& $0.2449{\scriptstyle\pm0.0019}$
& $\mathbf{0.4692{\scriptstyle\pm0.0073}}$
& $98.566{\scriptstyle\pm0.122}$
& $79.278{\scriptstyle\pm0.601}$
& $87.332{\scriptstyle\pm0.405}$
& $78.465{\scriptstyle\pm0.089}$
& $3.86$ \\
MMA 
& $0.6853{\scriptstyle\pm0.0090}$
& $0.2478{\scriptstyle\pm0.0056}$
& $0.4691{\scriptstyle\pm0.0095}$
& $98.616{\scriptstyle\pm0.063}$
& $\mathbf{80.138{\scriptstyle\pm0.287}}$
& $\mathbf{87.421{\scriptstyle\pm0.125}}$
& $79.014{\scriptstyle\pm0.110}$
& $3.21$ \\
SiGMA
& $0.7080{\scriptstyle\pm0.0063}$
& $0.2441{\scriptstyle\pm0.0017}$
& $0.4687{\scriptstyle\pm0.0070}$
& $98.628{\scriptstyle\pm0.105}$
& $79.528{\scriptstyle\pm0.180}$
& $87.395{\scriptstyle\pm0.194}$
& $\mathbf{79.087{\scriptstyle\pm0.158}}$
& $\mathbf{2.36}$ \\
\bottomrule
\end{tabular}}
\end{table}

Table~\ref{tab:architecture-ablations} reports the results together with each model's average rank across datasets. SiGMA achieves the best average rank of $2.36$ and ranks among the three strongest variants on every dataset, demonstrating the most consistent performance across tasks. No single ablation dominates uniformly: the fully ungated hybrid (MMA) performs best on CIFAR10 and PATTERN, disabling message-passing gates gives the strongest Peptides-func and MNIST results, and disabling attention gates gives the lowest Peptides-struct error. Nevertheless, each ungated variant has a worse average rank than SiGMA \footnote{In most cases, the differences between SiGMA and MMA are not statistically significant at $\alpha=0.05$ in Tab.~\ref{tab:architecture-ablations}.}. These results suggest that gating is particularly valuable as a mechanism for maintaining robust performance across heterogeneous tasks.

The MP-only and Attention-only variants serve as controls for whether SiGMA's performance can be attributed to either computational mechanism in isolation. Both are less consistent than the gated hybrid, with attention-only performing poorly on CIFAR10 and CLUSTER. Further experiments on the TU datasets, reported in Appendices~\ref{app:additional-results-TU} and~\ref{app:gate-analysis}, examine the distinct contribution of combining different message-passing operators within the same model. They show that heterogeneous head configurations can improve upon homogeneous alternatives on several datasets, providing additional evidence for the flexibility afforded by making multiple local inductive biases available within a single layer.

\section{Discussion}



In this paper, we introduced graph-level heterogeneity as the observation that graphs within the same dataset can differ substantially in the computational mechanisms they require. Our analysis shows that prediction difficulty emerges from the interaction between a graph and a model’s inductive biases: changing the message-passing operator or network depth changes not only aggregate performance, but also which individual graphs are easy or difficult. These findings challenge the standard practice of selecting a single computational template at the dataset level and motivate architectures that can adapt their computation across inputs.

To address this heterogeneity, we proposed SiGMA, which combines gated message-passing and attention heads within each layer. This design makes local aggregation, non-local communication, and effective depth adaptable across nodes, layers, and graphs. Our theoretical results establish that SiGMA retains the expressivity of its message-passing backbone, admits near-identity paths that avoid forcing over-smoothing, and enables direct communication between distant nodes to alleviate over-squashing. Its strong performance across graph-level benchmarks, together with the ablation results, demonstrates the promise of adaptively combining multiple aggregation mechanisms. A controlled synthetic benchmark in Appendix~\ref{app:routing analysis} further provides direct evidence of graph-specific routing: SiGMA learns to favor the operator appropriate for each graph type, while selectively masking that operator disproportionately harms performance on the corresponding subset. 

\paragraph{Limitations and future work.} Several directions could further develop this perspective. Our heterogeneity profiles characterize how graphs interact with separately trained architectures. Although the controlled benchmark isolates SiGMA's capacity for graph-specific operator routing, extending this mechanistic analysis to real-world datasets could reveal how its learned gates relate to graph structure, prediction difficulty, and operator preference. Likewise, although the gates can suppress individual heads, the current implementation does not convert this selectivity into reduced computation. Future work could therefore investigate sparse or conditional execution, more diverse collections of message-passing operators, and scalable alternatives to full attention. Extending the framework to directed graphs, hypergraphs, and larger graph-level tasks would also help establish how broadly adaptive hybrid computation can address heterogeneity.

\section{Acknowledgments}
LF is supported by the Kempner Graduate Fellowship at Harvard University. MW was supported in part by NSF award CBET-2112085, an Alfred P. Sloan Research Fellowship in Mathematics, and the AI2050 program at Schmidt Sciences (Grant G-25-69786). The computations in this paper were run on the FASRC Cannon cluster supported by the FAS Division of Science Research Computing Group at Harvard University and on the Kempner Institute AI Cluster at Harvard University.

\subsection*{GenAI Usage Statement}

In preparing this manuscript, the authors used generative AI tools for writing code and to assist with drafting and revising portions of the text and mathematical proofs. All AI-assisted material was reviewed, verified, and corrected by the authors. The authors take full responsibility for the final content of the work, including all claims, proofs, references, and artifacts. 

\bibliographystyle{unsrtnat}
\bibliography{reference}

\clearpage
\appendix
\etocdepthtag.toc{appendix}

\etocsettagdepth{main}{none}
\etocsettagdepth{appendix}{subsection}

\renewcommand{\contentsname}{Appendices}
\tableofcontents
\clearpage

\section{Heterogeneity Analysis}

\subsection{Construction of Graph-Level Heterogeneity Profiles}
\label{app:heterogeneity-profile-construction}

We use \emph{graph-level heterogeneity profiles} to estimate how consistently a
trained model succeeds or fails on each individual graph in a dataset. A single
test-set prediction is generally not sufficient for this purpose: its outcome
depends on the sampled training set, model initialization, stochastic
optimization, and checkpoint selection. We therefore evaluate every graph
repeatedly across independently trained models and random data splits.

Let
\begin{equation}
    \mathcal{D}=\{(G_i,y_i)\}_{i=1}^{n}
\end{equation}
denote a graph-level classification or regression dataset. In each trial
\(t\), we randomly partition \(\mathcal{D}\) into training, validation, and test
sets using a \(50/25/25\) split. We train the selected architecture for
\(300\) epochs on the training set and retain the checkpoint with the best
validation performance. For every graph \(G_i\) assigned to the test set in
trial \(t\), we record the prediction
\(\widehat{y}_{i}^{(t)}\). We repeat this procedure with independently sampled
splits and random initializations until every graph has appeared in the test
set at least \(T=100\) times. Denoting the set of trials in which \(G_i\)
appears in the test set by
\begin{equation}
    \mathcal{T}_i
    =
    \left\{
        t : G_i \in \mathcal{D}_{\mathrm{test}}^{(t)}
    \right\},
\end{equation}
we thus ensure that \(\lvert\mathcal{T}_i\rvert \geq T\) for every graph.

For a classification task, the graph-level performance score is the empirical
probability that \(G_i\) is classified correctly:
\begin{equation}
    s_i^{\mathrm{cls}}
    =
    \frac{1}{\lvert\mathcal{T}_i\rvert}
    \sum_{t\in\mathcal{T}_i}
    \mathbbm{1}
    \left[
        \widehat{y}_{i}^{(t)}=y_i
    \right].
\end{equation}
Hence, \(s_i^{\mathrm{cls}}=1\) means that the graph is classified correctly
whenever it appears in the test set, whereas \(s_i^{\mathrm{cls}}=0\) means
that it is never classified correctly. For a regression task, we instead
compute the average graph-level absolute error,
\begin{equation}
    s_i^{\mathrm{reg}}
    =
    \frac{1}{\lvert\mathcal{T}_i\rvert}
    \sum_{t\in\mathcal{T}_i}
    \left\|
        \widehat{y}_{i}^{(t)}-y_i
    \right\|_1,
\end{equation}
where the \(\ell_1\) error is averaged over output dimensions for
multi-target regression. Larger classification scores indicate easier graphs,
whereas larger regression scores indicate harder graphs.

\begin{figure}[h]
    \centering

    \begin{minipage}{0.32\textwidth}
        \centering
        \includegraphics[width=\linewidth]{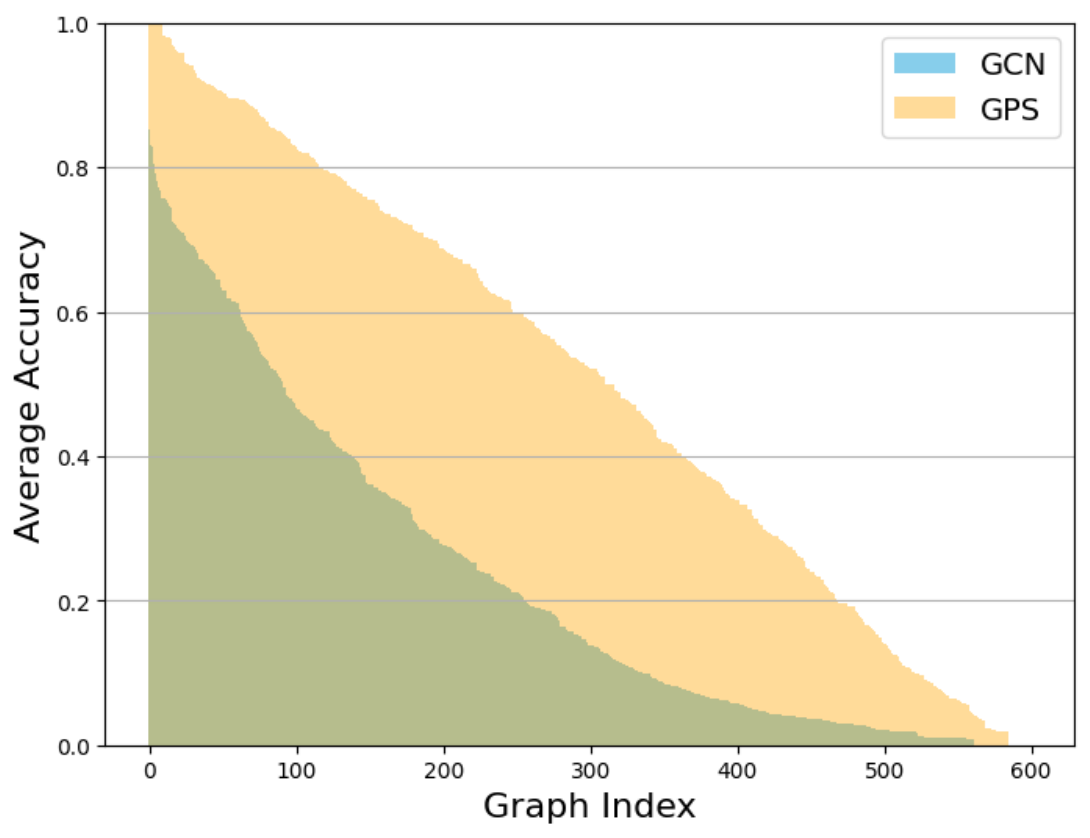}
    \end{minipage}
    \hfill
    \begin{minipage}{0.32\textwidth}
        \centering
        \includegraphics[width=\linewidth]{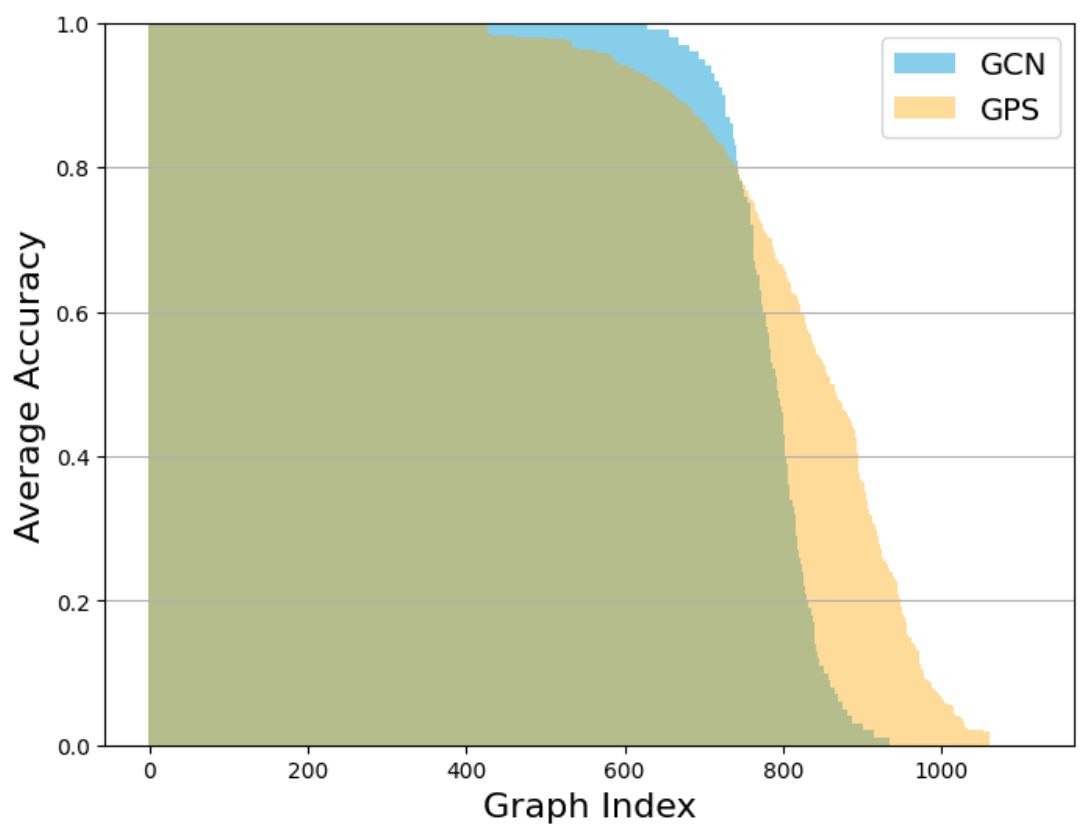}
    \end{minipage}
    \hfill
    \begin{minipage}{0.32\textwidth}
        \centering
        \includegraphics[width=\linewidth]{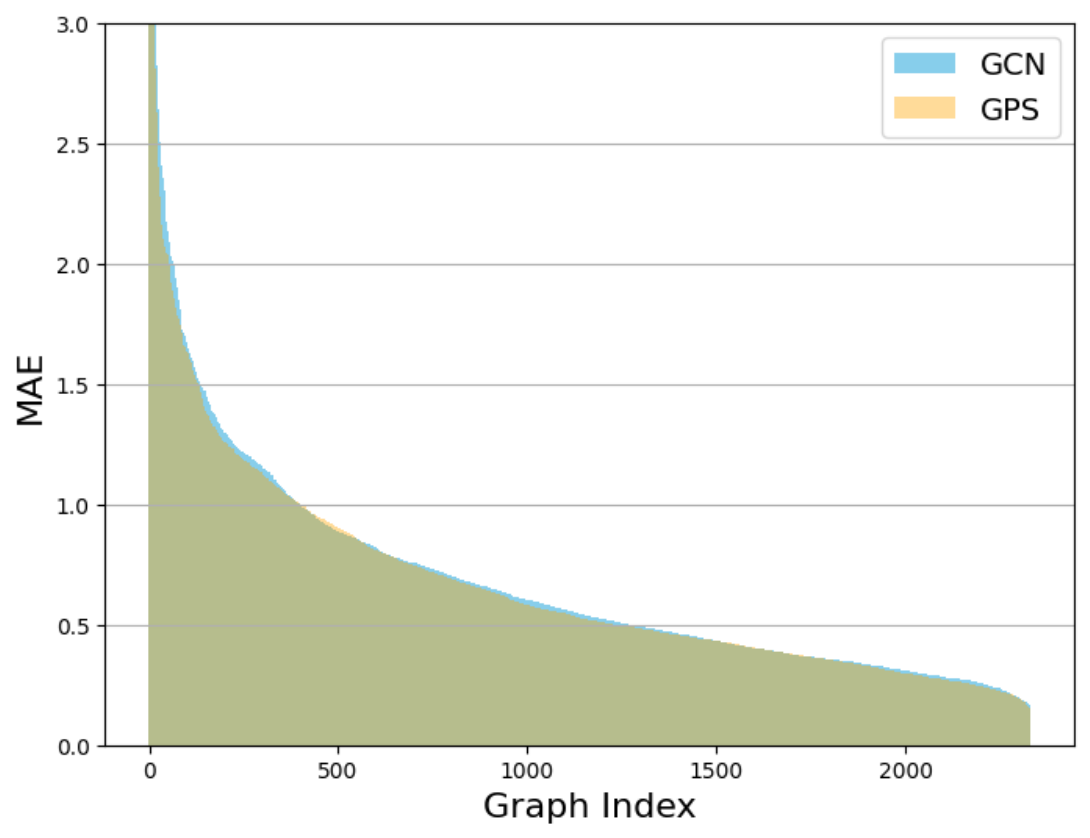}
    \end{minipage}

    \caption{
        Graph-level heterogeneity profiles for GCN and GraphGPS on
        ENZYMES (left), Proteins (center), and Peptides-struct (right).
        Across both message-passing and hybrid architectures, individual
        graphs vary substantially in prediction difficulty.
    }
    \label{fig:gcn-graphgps-profiles}
\end{figure}

\begin{figure}[h]
    \centering

    \begin{minipage}{0.48\textwidth}
        \centering
        \includegraphics[width=\linewidth]{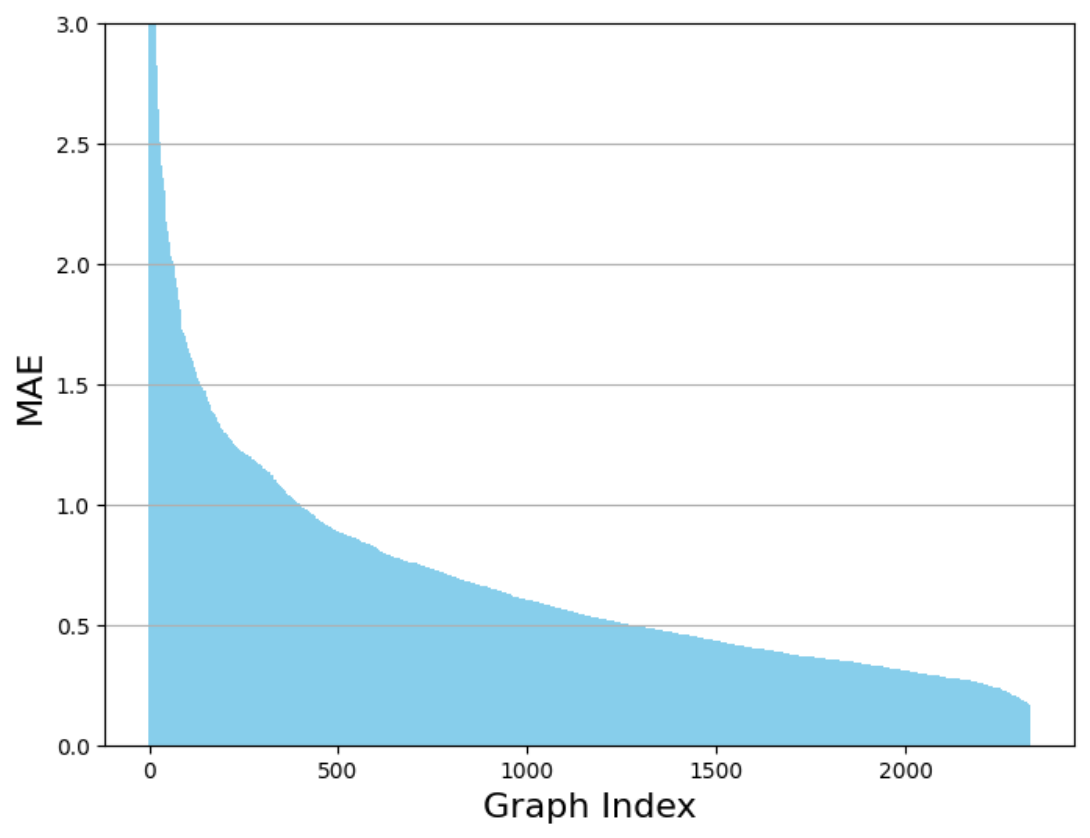}
    \end{minipage}
    \hfill
    \begin{minipage}{0.48\textwidth}
        \centering
        \includegraphics[width=\linewidth]{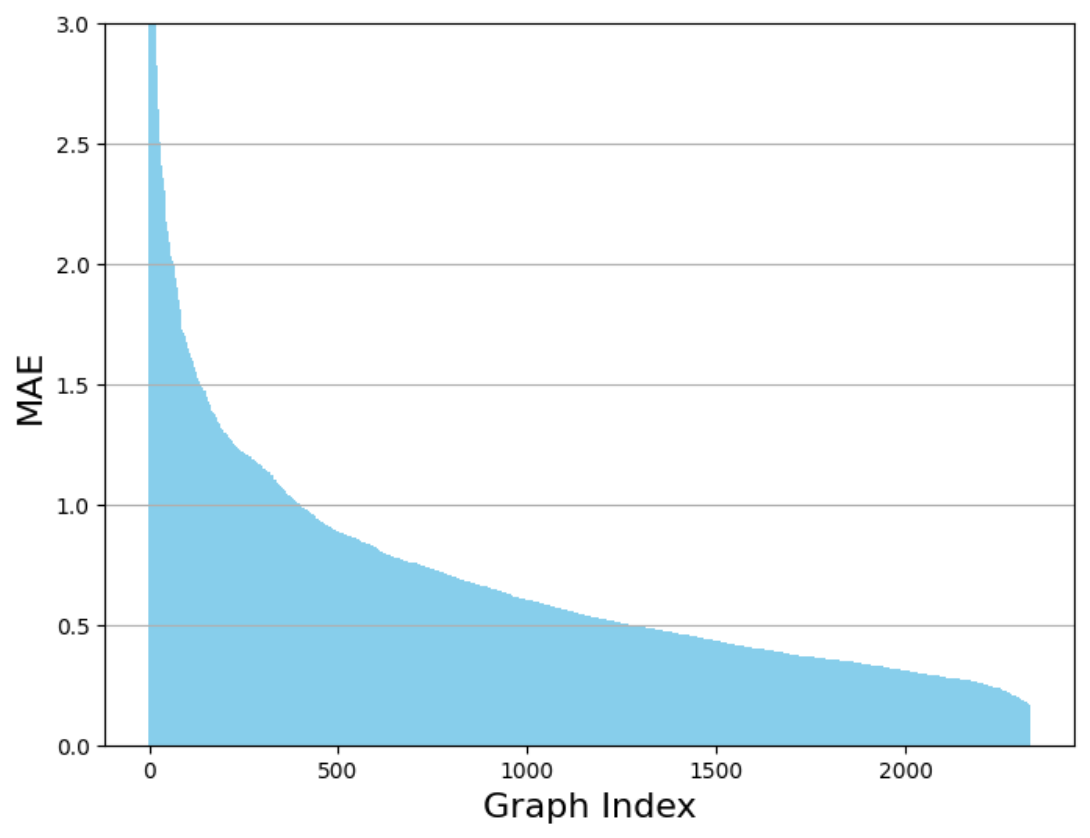}
    \end{minipage}

    \caption{
        Graph-level heterogeneity profiles for a four-layer GCN on Peptides-struct, evaluated on the validation set (left) and test set (right). Both splits exhibit substantial variation in per-graph regression error.
    }
    \label{fig:zinc-validation-test-profiles}
\end{figure}

\begin{figure}[h]
    \centering

    \begin{minipage}{0.24\textwidth}
        \centering
        \includegraphics[width=\linewidth]{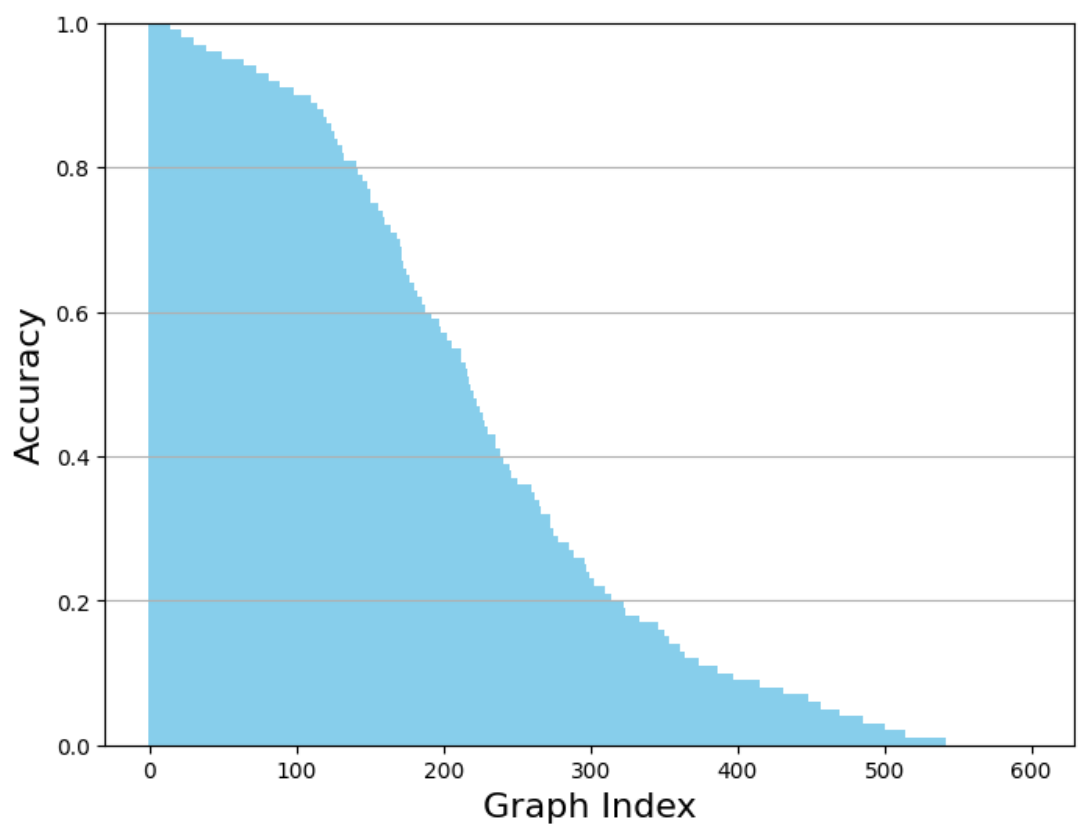}
    \end{minipage}
    \hfill
    \begin{minipage}{0.24\textwidth}
        \centering
        \includegraphics[width=\linewidth]{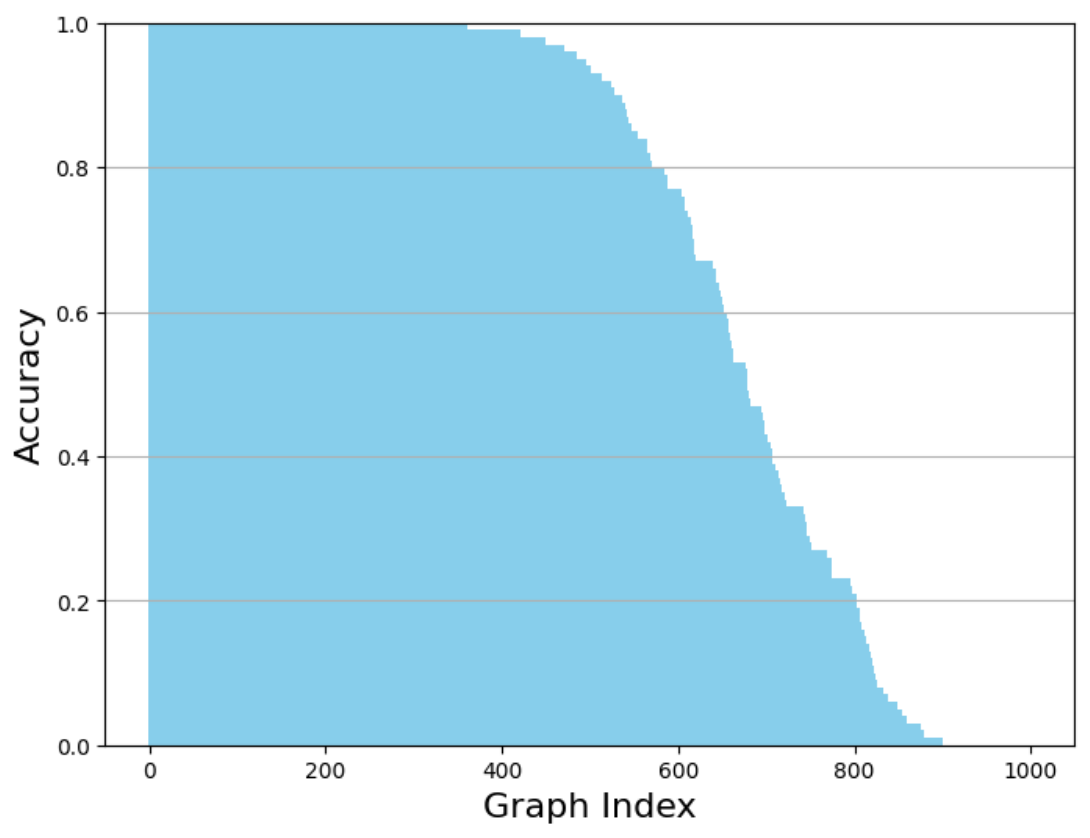}
    \end{minipage}
    \hfill
    \begin{minipage}{0.24\textwidth}
        \centering
        \includegraphics[width=\linewidth]{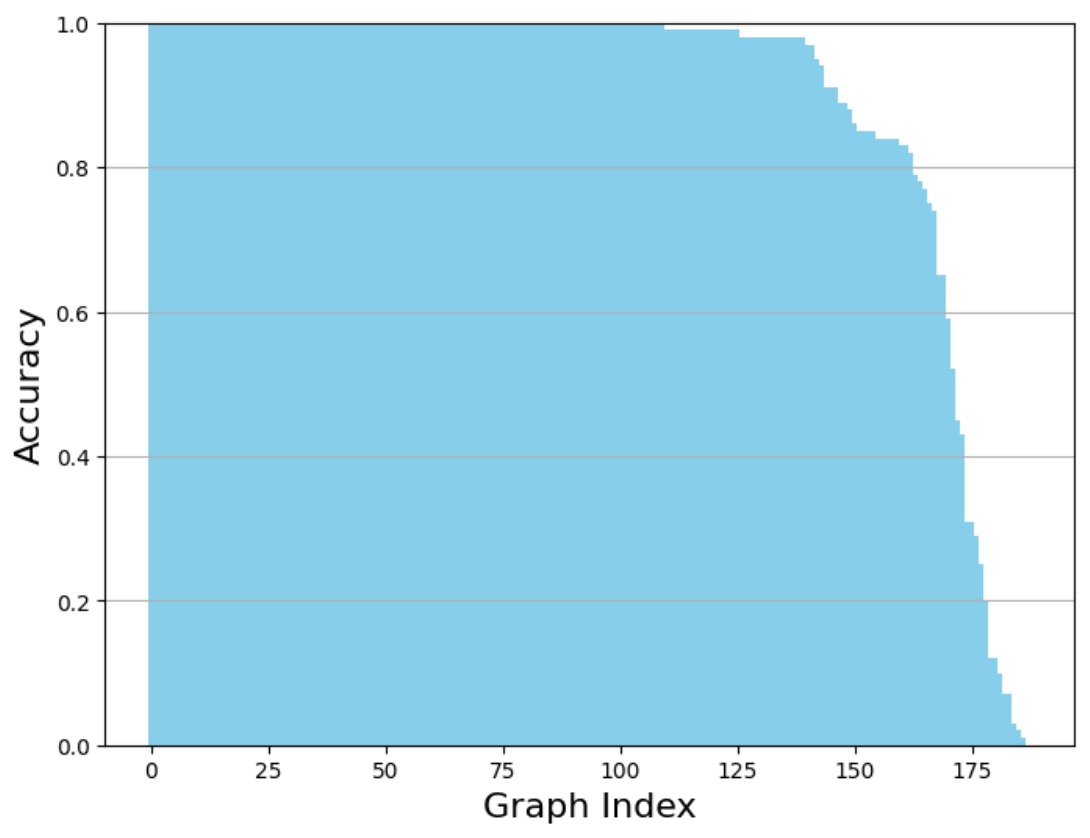}
    \end{minipage}
    \hfill
    \begin{minipage}{0.24\textwidth}
        \centering
        \includegraphics[width=\linewidth]{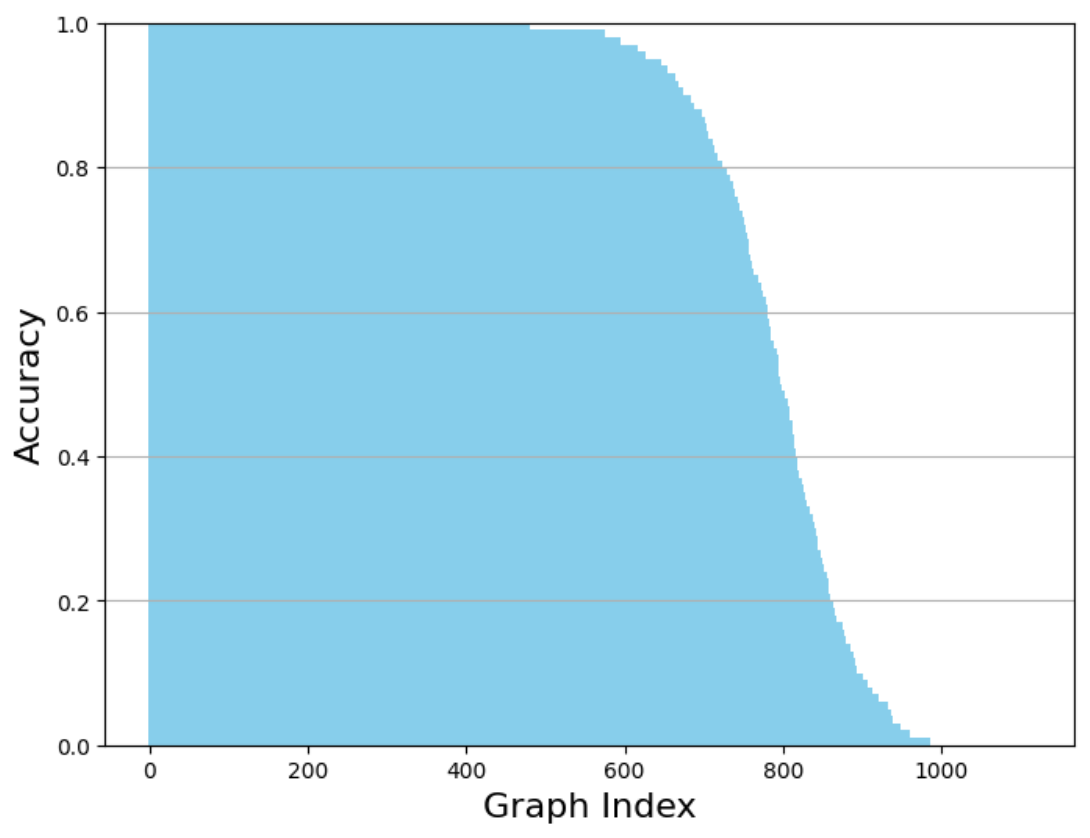}
    \end{minipage}

    \caption{
        Graph-level heterogeneity profiles for GIN on ENZYMES, IMDB-BINARY,
        MUTAG, and PROTEINS, from left to right. Substantial variation in
        prediction difficulty persists across all four datasets.
    }
    \label{fig:gin-heterogeneity-profiles}
\end{figure}

\begin{figure}[h!]
    \centering

    \begin{minipage}{0.24\textwidth}
        \centering
        \includegraphics[width=\linewidth]{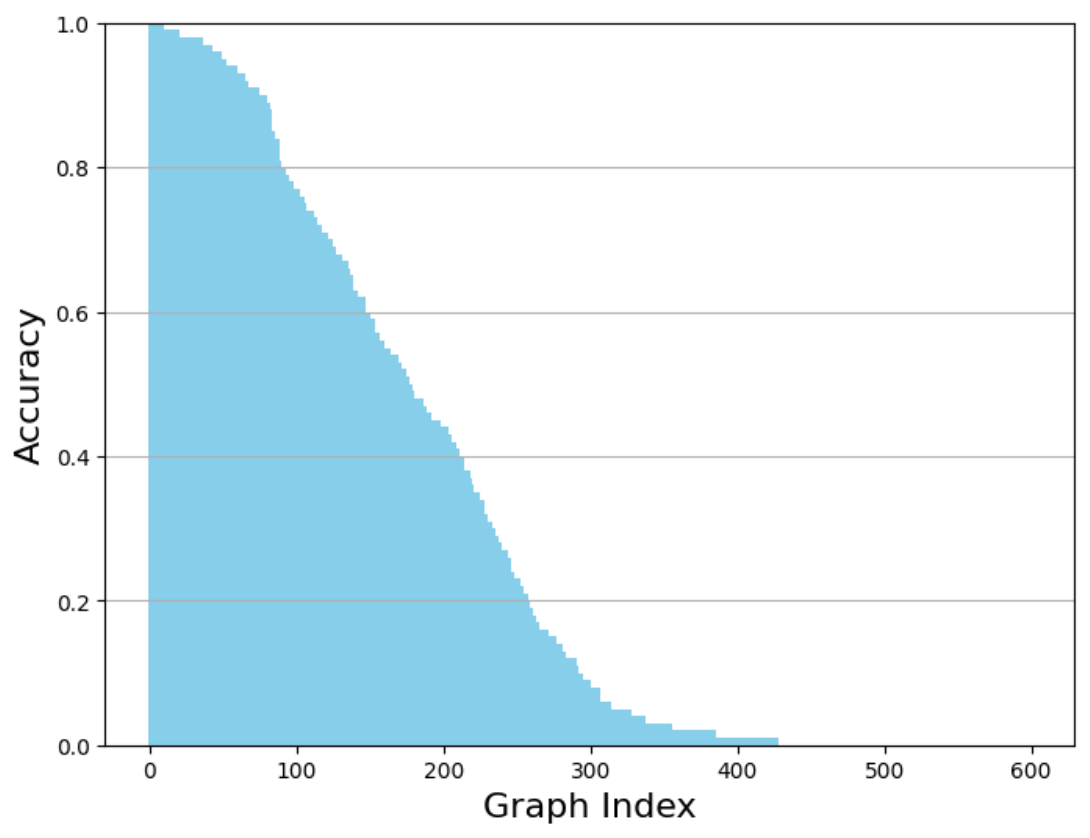}
    \end{minipage}
    \hfill
    \begin{minipage}{0.24\textwidth}
        \centering
        \includegraphics[width=\linewidth]{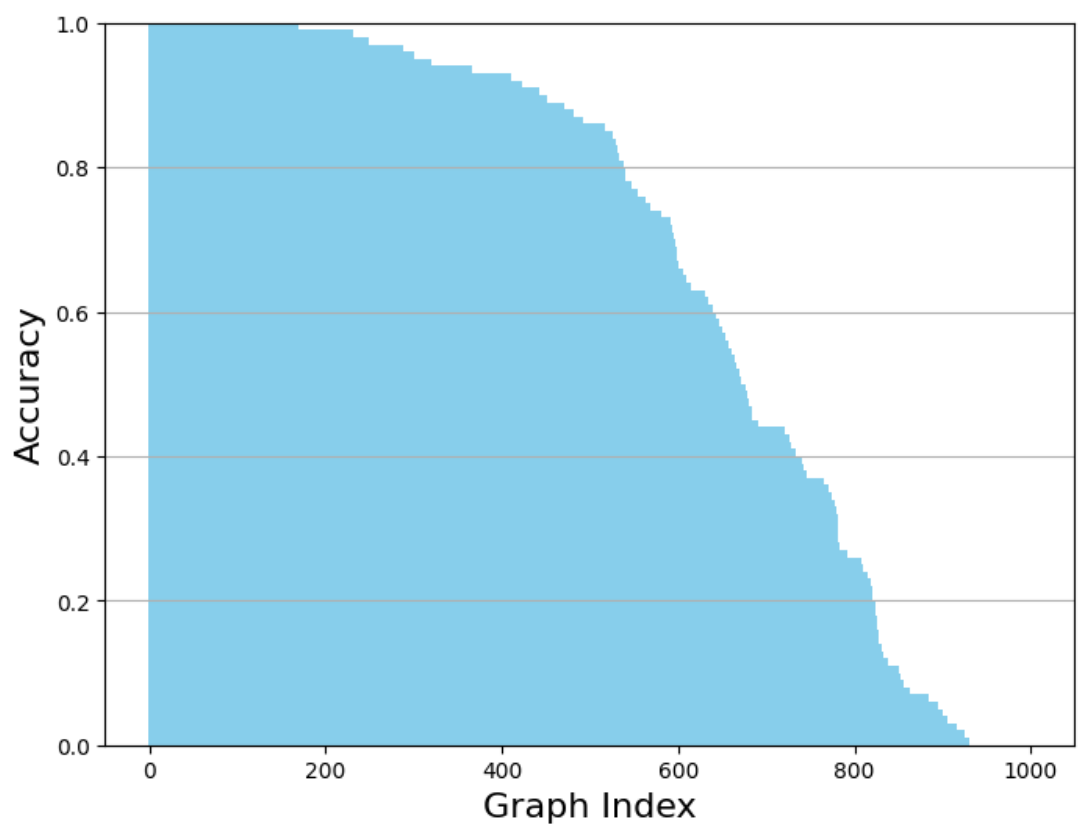}
    \end{minipage}
    \hfill
    \begin{minipage}{0.24\textwidth}
        \centering
        \includegraphics[width=\linewidth]{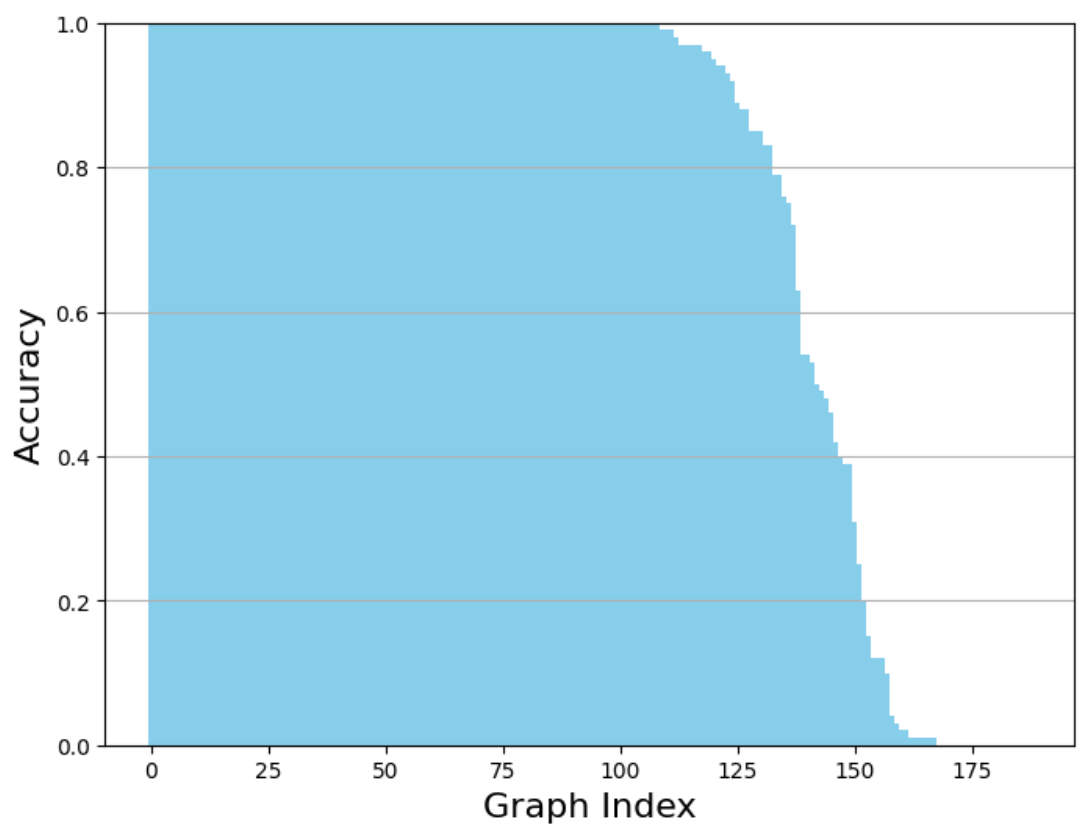}
    \end{minipage}
    \hfill
    \begin{minipage}{0.24\textwidth}
        \centering
        \includegraphics[width=\linewidth]{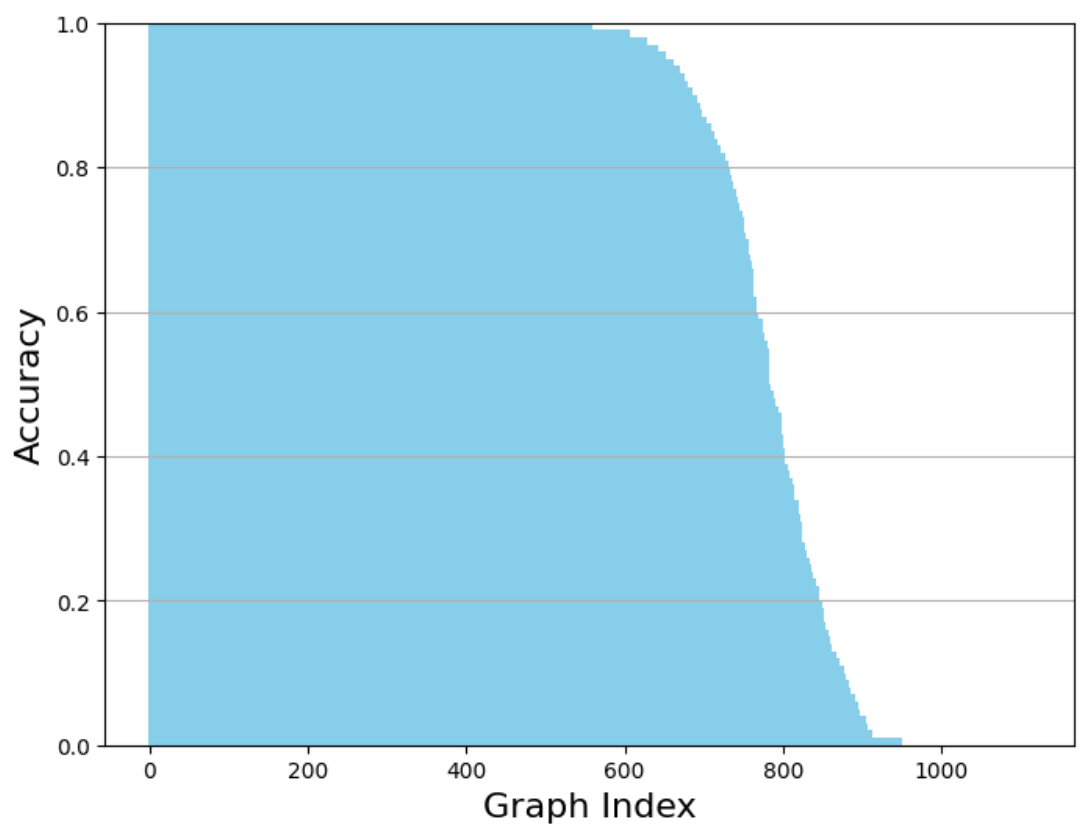}
    \end{minipage}

    \caption{
        Graph-level heterogeneity profiles for GAT on ENZYMES, IMDB-BINARY,
        MUTAG, and PROTEINS, from left to right. Substantial variation in
        prediction difficulty persists across all four datasets.
    }
    \label{fig:gat-heterogeneity-profiles}
\end{figure}

The heterogeneity profile of a model on a dataset is the collection
\begin{equation}
    \mathcal{H}
    =
    \{s_i\}_{i=1}^{n}.
\end{equation}
For visualization, we can sort graphs by their estimated score and plot the
resulting ordered sequence as in Fig.~\ref{fig:gcn-graphgps-profiles},   ~\ref{fig:zinc-validation-test-profiles}, ~\ref{fig:gin-heterogeneity-profiles}, ~\ref{fig:gat-heterogeneity-profiles},  ~\ref{fig:gcn-mutag-by-class},~\ref{fig:gcn-mutag-by-class-augmented},     ~\ref{fig:gcn-enzymes-by-class-dr},     ~\ref{fig:gcn-sigma-enzymes-by-class}. Sometimes, we prefer to order the graphs by their graph index as in     Fig.~\ref{fig:hetero_profiles_main}. A nearly flat profile indicates that the model
performs similarly across graph instances. In contrast, a profile spanning a
wide range of values indicates substantial graph-level heterogeneity: some
graphs are consistently predicted correctly or with low error, while others
remain difficult across many independently trained models. Sorting is used
only for visualization; graph identities are retained when comparing profiles
across architectures, hyperparameter settings, and training epochs.

We use the same repeated-evaluation procedure to determine graph-dependent
architectural preferences. Let \(\mathcal{M}\) be a collection of
message-passing mechanisms, and let \(s_{i,m}\) denote the estimated score of
mechanism \(m\in\mathcal{M}\) on graph \(G_i\). For classification, the
preferred mechanism is
\begin{equation}
    m_i^\star
    =
    \arg\max_{m\in\mathcal{M}} s_{i,m},
\end{equation}
whereas for regression, it is
\begin{equation}
    m_i^\star
    =
    \arg\min_{m\in\mathcal{M}} s_{i,m}.
\end{equation}
The percentages reported in Tab.~\ref{tab:mp-preferences} are the fractions
of graphs assigned to each mechanism. Analogously, for a collection of
candidate depths \(\mathcal{L}\), we define
\begin{equation}
    L_i^\star
    =
    \arg\max_{L\in\mathcal{L}} s_{i,L}
\end{equation}
for classification and replace the maximum by a minimum for regression. These
assignments produce the preference percentages in section~\ref{sec:graph-level-heterogeneity}. Ties between $m$ different models are resolved by adding $1/m$ to the number of graphs that attain their best performance with that particular model (e.g. message passing operator or number of layers). The results in Tab.~\ref{tab:mp-preferences} use GCN, GIN, and GatedGCN models with 4 layers, hidden dimension of 64, ReLU activation, dropout of 0.1, global mean pooling, and an MLP classifier head. Tab.~\ref{tab:depth-preferences} uses GCN with the exact same hyperparameters and varies only the model depth. Because every score is averaged over repeated held-out
predictions, these comparisons measure persistent graph-level differences
rather than the outcome of a single split or training run.

\subsection{Additional Heterogeneity Profiles Across Datasets and Models}
\label{app:additional-heterogeneity-profiles}


\subsection{Robustness to Model Architecture and Hyperparameters}
\label{app:heterogeneity-robustness}

Graph-level heterogeneity is not specific to a single GNN
architecture. Section~\ref{sec:graph-level-heterogeneity} and
Appendix~\ref{app:additional-heterogeneity-profiles} visualize qualitatively similar
profiles for GCN, GIN, GAT, and GraphGPS. This includes both
message-passing and transformer-based architectures, as well as graph
classification and regression tasks. Across these settings, we consistently
observe substantial variation in per-graph performance: some graphs are
predicted correctly in nearly every trial, whereas others remain difficult
across independently trained models. The persistence of this pattern across
architecturally different models suggests that graph-level heterogeneity is not
an artifact of a particular aggregation rule or of message passing alone.

The same qualitative behavior is also observed across the model widths, depths,
and learning rates considered in the earlier experiments (not plotted here). In particular, the heterogeneity profiles remain non-uniform under changes to these
hyperparameters: varying model capacity or optimization settings changes the
overall level of performance, but does not eliminate the large differences
between easy and difficult graph instances. Taken together, these results show that graph-level heterogeneity persists
across multiple model families and reasonable hyperparameter configurations.

\subsection{Consistency of Graph Difficulty Across Training}
\label{app:training-dynamics}

To examine when graph-level heterogeneity emerges, we track predictions on
individual test graphs throughout training rather than only evaluating the
checkpoint selected at the end. After every epoch, we record the prediction
made for each graph in the test set and compute the corresponding graph-level
accuracy across repeated runs. This allows us to distinguish graphs that are
learned early and consistently from those that are learned only after prolonged
training or remain misclassified throughout.

\begin{figure}[h]
    \centering
    \begin{minipage}{0.48\textwidth}
        \centering
        \includegraphics[width=\linewidth]{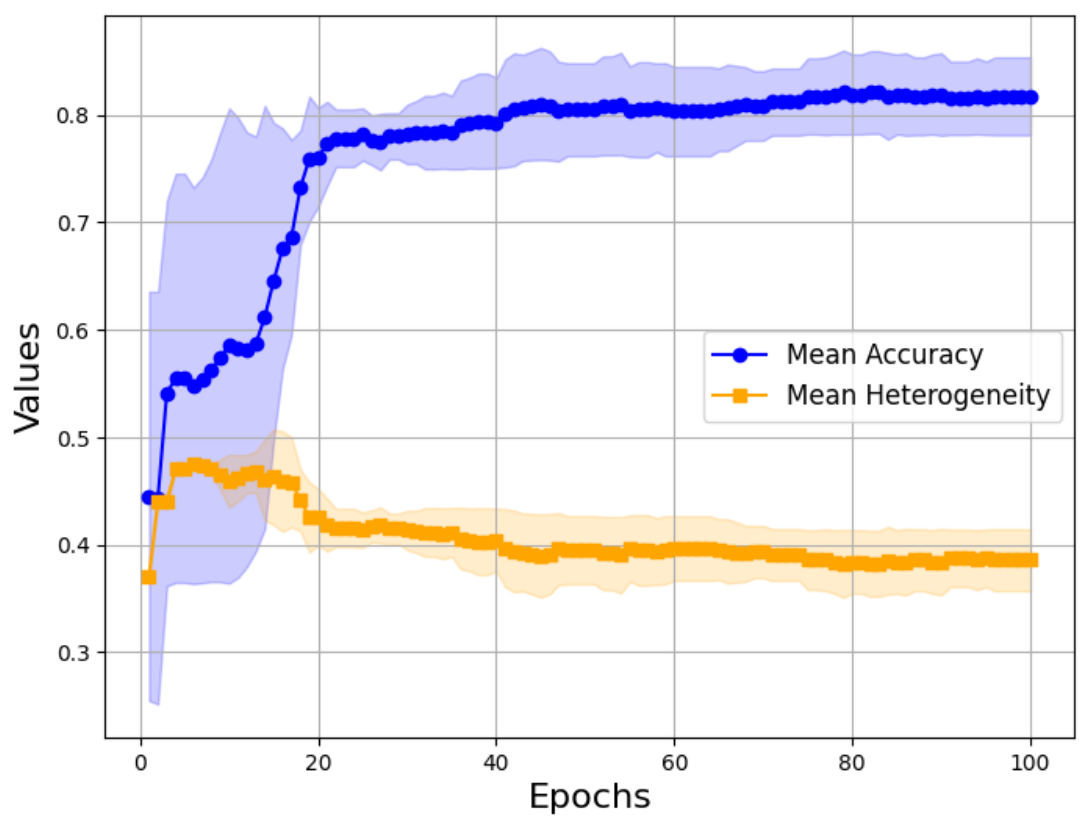}
    \end{minipage}
    \hfill
    \begin{minipage}{0.48\textwidth}
        \centering
        \includegraphics[width=\linewidth]{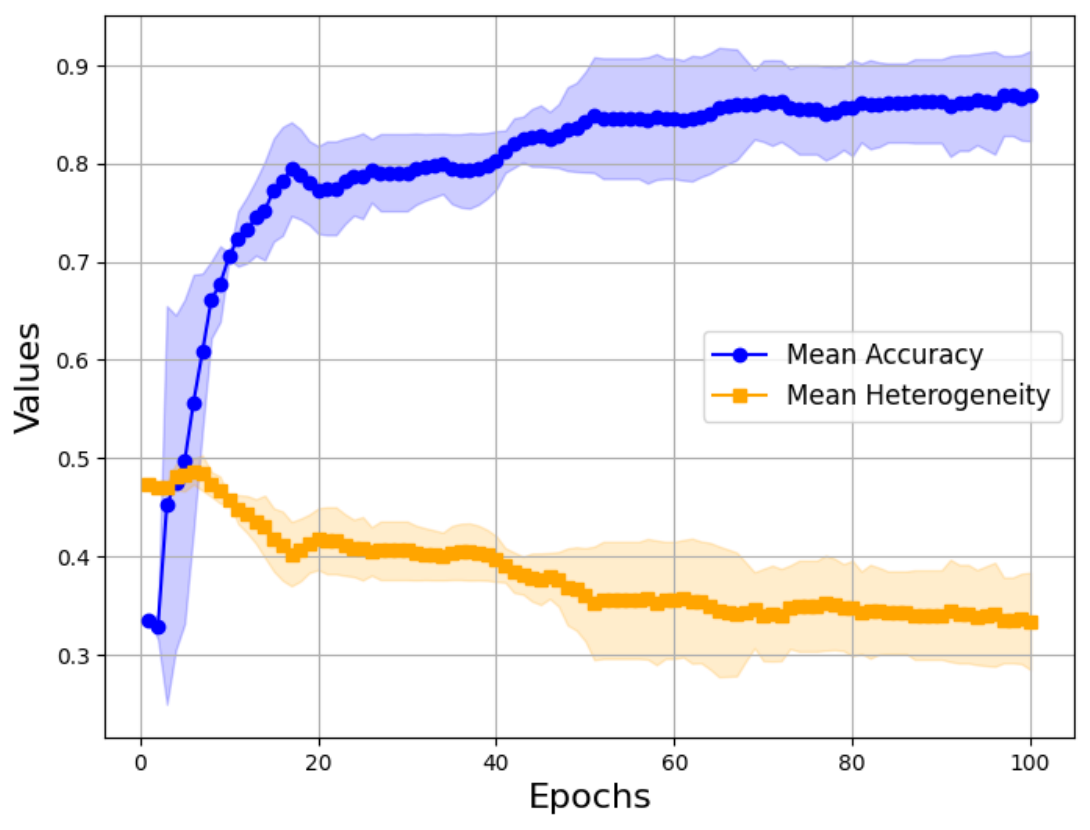}
    \end{minipage}
    \caption{
        Training dynamics on MUTAG with GIN (left) and GraphGPS with GIN backbone. Graph-level performance is recorded throughout training and averaged across repeated runs. Heterogeneity first increases as the models learn some graphs before others and then decreases as additional graphs are learned.
    }
    \label{fig:mutag-training-dynamics}
\end{figure}

Fig.~\ref{fig:mutag-training-dynamics} shows this analysis on MUTAG. Across
the considered architectures, heterogeneity initially increases: the models
quickly learn a subset of graphs, while their predictions on the remaining
graphs are still close to chance. As training continues, average accuracy
increases and the dispersion in graph-level performance gradually decreases
because additional graphs are learned. This produces a characteristic
``learning in stages'' pattern in which different subsets of graphs are learned
at different points during optimization.

Importantly, these results do not imply that difficulty is an intrinsic
property of the data alone. Rather, graph difficulty emerges from the
interaction between an input graph and the model used to process it. Some
graphs are consistently easy or difficult across several architectures,
indicating a partially shared component of instance difficulty. However, the
ordering of graphs is only partially preserved across models. In particular,
graphs that are difficult under GCN message passing, or under GraphGPS with a
GCN local head, can be easy under GIN message passing or GraphGPS with a GIN
local head, and conversely. Changing the local aggregation rule can therefore
reverse the relative difficulty of individual graph instances.

\begin{figure}[h]
    \centering
    \begin{minipage}{0.48\textwidth}
        \centering
        \includegraphics[width=\linewidth]{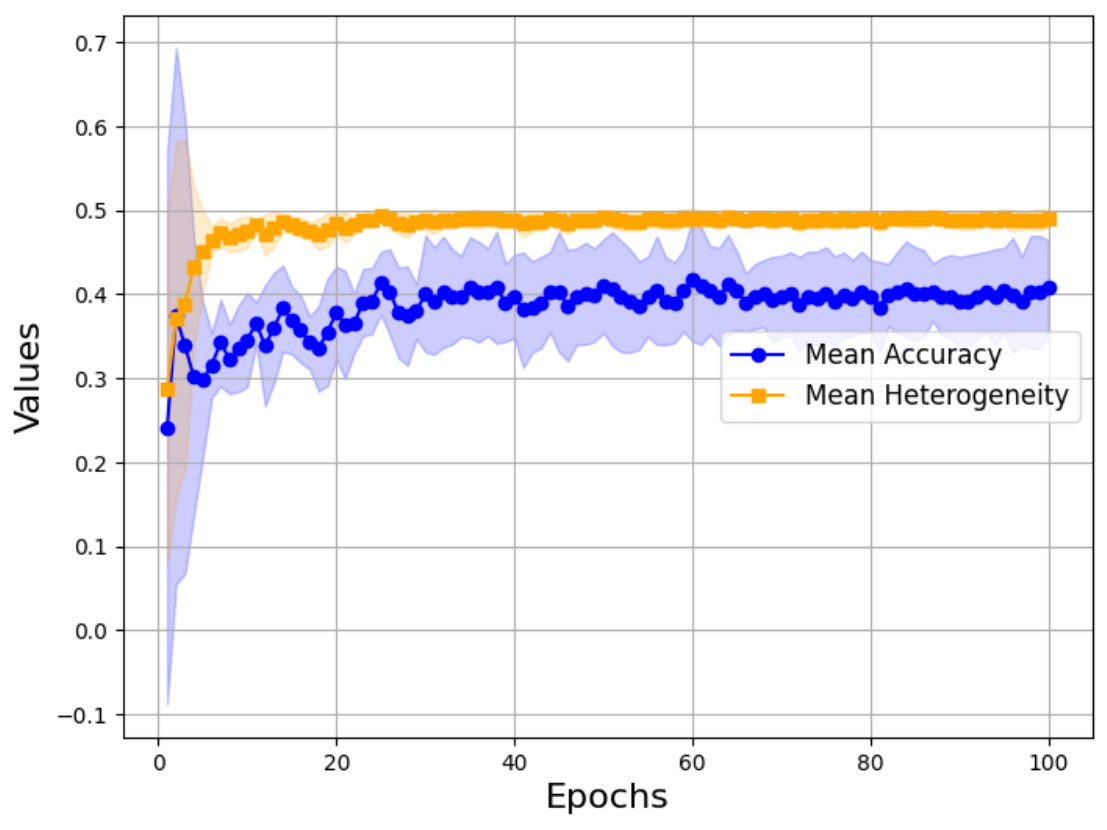}
    \end{minipage}
    \hfill
    \begin{minipage}{0.48\textwidth}
        \centering
        \includegraphics[width=\linewidth]{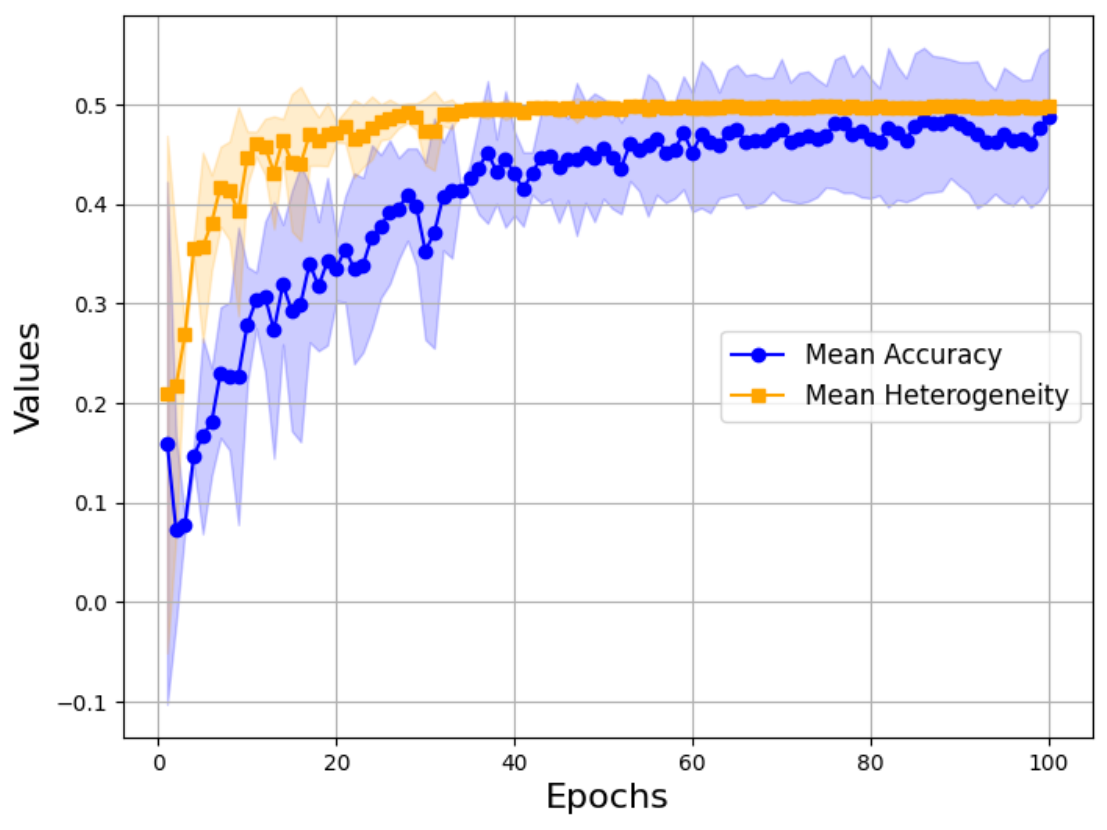}
    \end{minipage}
    \caption{
        Training dynamics on ENZYMES with GIN (left) and GraphGPS with GIN backbone. Graph-level performance is recorded throughout training and averaged across repeated runs.
    }
    \label{fig:enzymes-training-dynamics}
\end{figure}

We observe a different qualitative training behavior on ENZYMES. As shown in
Fig.~\ref{fig:enzymes-training-dynamics}, graph-level heterogeneity rises
during the early stages of training and subsequently remains high, presumably because ENZYMES is a much harder dataset than MUTAG which neither GIN nor GraphGPS can solve to the extent that they can solve MUTAG. At the same time, comparisons between GIN and GraphGPS show both overlap and disagreement in which graphs remain difficult (not plotted here). 

These results complement the final heterogeneity profiles in
Appendix~\ref{app:additional-heterogeneity-profiles}. The profiles do not merely
capture transient errors from incompletely trained models: differences between
graphs arise early and can persist for a subset of examples even after extended
training. At the same time, neither the final profiles nor the training
trajectories define a universal ordering from easy to difficult. They instead
contain both a shared component, reflecting graphs that tend to be easy or
difficult across several models, and an architecture-dependent component,
reflecting graphs whose difficulty changes substantially with the chosen
message-passing operator or hybrid architecture. 

\subsection{Graph Statistics and Prediction Difficulty}
\label{app:graph-statistics}

A natural hypothesis is that graph-level prediction difficulty is a consequence of topological properties of the input graph. This would parallel
findings in node-level learning, where quantities such as node degree can be
strongly associated with classification performance. We therefore investigate
whether standard graph statistics explain the graph-level heterogeneity
observed in Section~\ref{sec:graph-level-heterogeneity}.

For each graph, we compute a collection of classical topological descriptors:
edge density, average degree, degree assortativity, pseudo-diameter, average
clustering coefficient, transitivity, algebraic connectivity, curvature gap,
and the relative size of the largest clique \cite{costa2007characterization, hagberg2008exploring, newman2003structure, fiedler1973algebraic, fesser2024augmentations}. Edge density and average degree
measure graph sparsity and local connectivity; assortativity captures whether
nodes tend to connect to nodes of similar degree; the pseudo-diameter measures
the largest observed shortest-path distance; and the clustering coefficient
and transitivity quantify the prevalence of triangles. Algebraic connectivity
is given by the second-smallest eigenvalue of the combinatorial graph
Laplacian and measures global connectivity \cite{fiedler1973algebraic}. The curvature gap summarizes the
separation between intra- and inter-community edges using Ollivier--Ricci
curvature \cite{fesser2024augmentations}, while the relative clique size records the size of the largest
clique normalized by the number of nodes.

We use these statistics to predict the normalized graph-level GCN accuracy
estimated from the repeated evaluation procedure in
Appendix~\ref{app:heterogeneity-profile-construction}. Following the analysis
used previously for node-level heterogeneity, we fit a sparse multivariate
linear regression with an \(\ell_1\) penalty:
\begin{equation}
    \widehat{\boldsymbol{\beta}}
    =
    \arg\min_{\boldsymbol{\beta},\beta_0}
    \left\{
        \frac{1}{n}
        \sum_{i=1}^{n}
        \left(
            s_i
            -
            \beta_0
            -
            \boldsymbol{x}_i^\top\boldsymbol{\beta}
        \right)^2
        +
        \lambda\lVert\boldsymbol{\beta}\rVert_1
    \right\},
\end{equation}
where \(s_i\) is the normalized average accuracy on graph \(G_i\) and
\(\boldsymbol{x}_i\) contains its standardized topological statistics. The
sparsity penalty selects a small subset of potentially explanatory graph
properties and reduces sensitivity to correlations among the descriptors.

\begin{table}[t]
    \centering
    \caption{
        Coefficients of the sparse multivariate regressions predicting
        normalized graph-level GCN accuracy from topological graph statistics.
        Results are shown for ENZYMES, IMDB-BINARY, MUTAG, and PROTEINS.
    }
    \label{tab:topological-regression}
\resizebox{\textwidth}{!}{%
\begin{tabular}{lrrrr}
\toprule
Features & ENZYMES & Imdb & MUTAG & Proteins \\
\midrule
Edge Density & $4.151617\mathrm{e}{-02}$ & $2.898623\mathrm{e}{-01}$ & $-3.969576\mathrm{e}{+00}$ & $-2.560137\mathrm{e}{-01}$ \\
Average Degree & $6.882024\mathrm{e}{-02}$ & $2.217939\mathrm{e}{-03}$ & $1.999772\mathrm{e}{+00}$ & $3.252736\mathrm{e}{-02}$ \\
Degree Assortativity & $-3.918472\mathrm{e}{-02}$ & $-3.185174\mathrm{e}{-01}$ & $-8.107750\mathrm{e}{-01}$ & $4.863875\mathrm{e}{-03}$ \\
Diameter & $2.235096\mathrm{e}{-03}$ & $-6.465376\mathrm{e}{-01}$ & $9.388648\mathrm{e}{-05}$ & $3.698471\mathrm{e}{-04}$ \\
Average Clustering Coefficient & $-2.483046\mathrm{e}{-01}$ & $-4.472245\mathrm{e}{-03}$ & $4.440892\mathrm{e}{-16}$ & $-2.026657\mathrm{e}{-01}$ \\
Transitivity & $2.299293\mathrm{e}{-01}$ & $-7.803414\mathrm{e}{-01}$ & $0$ & $2.238747\mathrm{e}{-01}$ \\
Algebraic Connectivity & $-2.117908\mathrm{e}{-02}$ & $-2.505663\mathrm{e}{-02}$ & $4.693080\mathrm{e}{+00}$ & $1.093306\mathrm{e}{-01}$ \\
Curvature Gap & $-3.988714\mathrm{e}{-07}$ & $-5.182994\mathrm{e}{-07}$ & $-5.617996\mathrm{e}{-04}$ & $2.957293\mathrm{e}{-08}$ \\
Rel.\ Size largest Clique & $2.773373\mathrm{e}{-02}$ & $4.727640\mathrm{e}{-03}$ & $0$ & $9.840849\mathrm{e}{-03}$ \\
\midrule
$R^{2}$ & 0.01670 & 0.04559 & 0.39944 & 0.01626 \\
\bottomrule
\end{tabular}%
}
\end{table}

With the exception of MUTAG, the resulting models explain very little of the
observed variation in graph-level accuracy, with coefficients of determination
below \(0.05\). Moreover, the selected coefficients are not consistent across
datasets. For example, higher algebraic connectivity is associated with higher
accuracy on PROTEINS, whereas the opposite tendency appears on ENZYMES. Similar
inconsistencies occur for other descriptors, indicating that no single
topological property provides a robust explanation of graph difficulty across
datasets.

We additionally examine individual Pearson correlations between graph-level
accuracy and selected topological quantities on MUTAG and ENZYMES. These
results are reported in Tab.~\ref{tab:topological-correlations}. Most
correlations are small in magnitude. We compute
unadjusted two-sided \(p\)-values. None of the correlations on MUTAG is
statistically significant at the \(0.05\) level, and only pseudo-diameter is
nominally significant on ENZYMES. For instance, the largest reported
correlation is between average degree and accuracy on MUTAG, but its magnitude
is only approximately \(0.12\). On ENZYMES, the largest
absolute correlation is between pseudo-diameter and accuracy
(\(r=0.084\), \(p=0.040\)); all remaining correlations are nonsignificant and
have absolute magnitudes below \(0.07\).


\begin{table}[t]
    \centering
    \caption{
        Pearson correlations between topological graph properties and
        graph-level GCN accuracy. Parentheses contain unadjusted two-sided
        \(p\)-values.
    }
    \label{tab:topological-correlations}
    \begin{tabular}{lcc}
        \toprule
        Property & MUTAG & ENZYMES \\
        \midrule
        Average degree
            & \(0.12423\; (0.08940)\)
            & \(0.02403\; (0.55689)\) \\
        Degree assortativity
            & \(-0.01695\; (0.81741)\)
            & \(0.04393\; (0.28267)\) \\
        Pseudo-diameter
            & \(0.01769\; (0.80959)\)
            & \(0.08380\; (0.04017)\) \\
        Average clustering coefficient
            & \(0.04105\; (0.57594)\)
            & \(0.06259\; (0.12566)\) \\
        Transitivity
            & \(0.03575\; (0.62621)\)
            & \(-0.06216\; (0.12829)\) \\
        Algebraic connectivity
            & \(-0.03241\; (0.65883)\)
            & \(-0.02461\; (0.54740)\) \\
        \bottomrule
    \end{tabular}
\end{table}

\begin{figure}[t]
    \centering
    \includegraphics[width=0.72\textwidth]{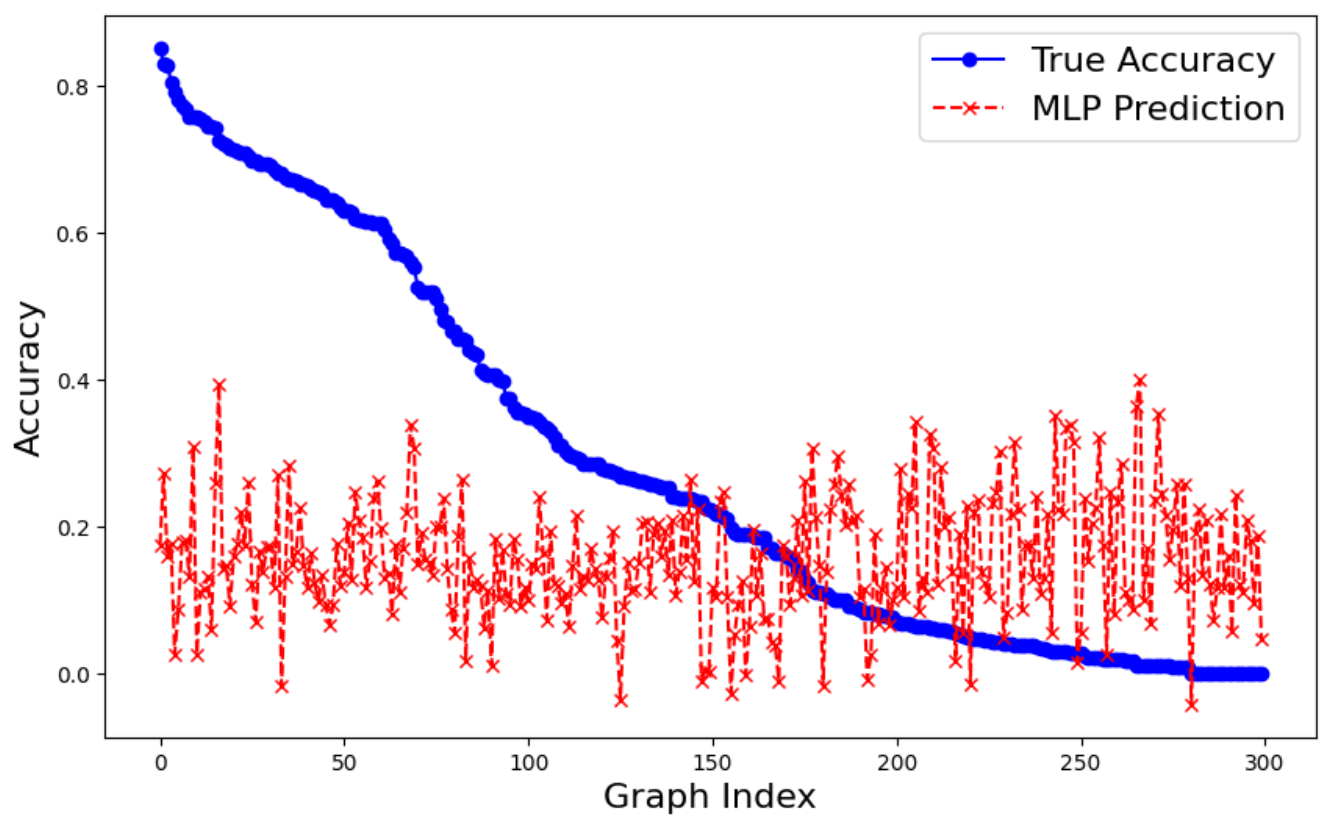}
    \caption{
        Accuracy of an MLP that predicts graph-level GCN performance on
        ENZYMES using only classical topological graph statistics.
    }
    \label{fig:topology-mlp}
\end{figure}

As an additional nonlinear test, we train a multilayer perceptron to predict
whether individual ENZYMES graphs are classified correctly using only the same
topological descriptors. Its limited predictive performance, shown in
Fig.~\ref{fig:topology-mlp}, is consistent with the sparse regression
analysis: allowing a nonlinear predictor does not make ordinary graph
statistics sufficient to recover the observed difficulty profile.

\subsection{Tree Mover's Distance and Class Separation}
\label{app:tmd-class-separation}

The weak explanatory power of standard graph statistics suggests that graph
difficulty depends on a richer interaction between topology and node features.
We therefore use the Tree Mover's Distance (TMD), which compares graphs through
their attributed computation trees and jointly captures structural and feature
similarity \cite{chuang2022tree}. For a labeled graph \(G_i\), we define its \emph{class-distance
ratio} as
\begin{equation}
    \rho(G_i)
    =
    \frac{
        \min\limits_{\substack{G_j \in \mathcal{D}\setminus\{G_i\}\\ y_j=y_i}}
        \operatorname{TMD}(G_i,G_j)
    }{
        \min\limits_{\substack{G_j \in \mathcal{D}\\ y_j\neq y_i}}
        \operatorname{TMD}(G_i,G_j)
    }.
\end{equation}
A ratio below one means that \(G_i\) is closer (using TMD) to a graph of the same class than
to any graph of another class. Conversely, \(\rho(G_i)>1\) indicates that the
nearest graph under TMD has a different label, making the example poorly
separated from competing classes.

\begin{figure}[h]
    \centering
    \includegraphics[width=1.0\textwidth]{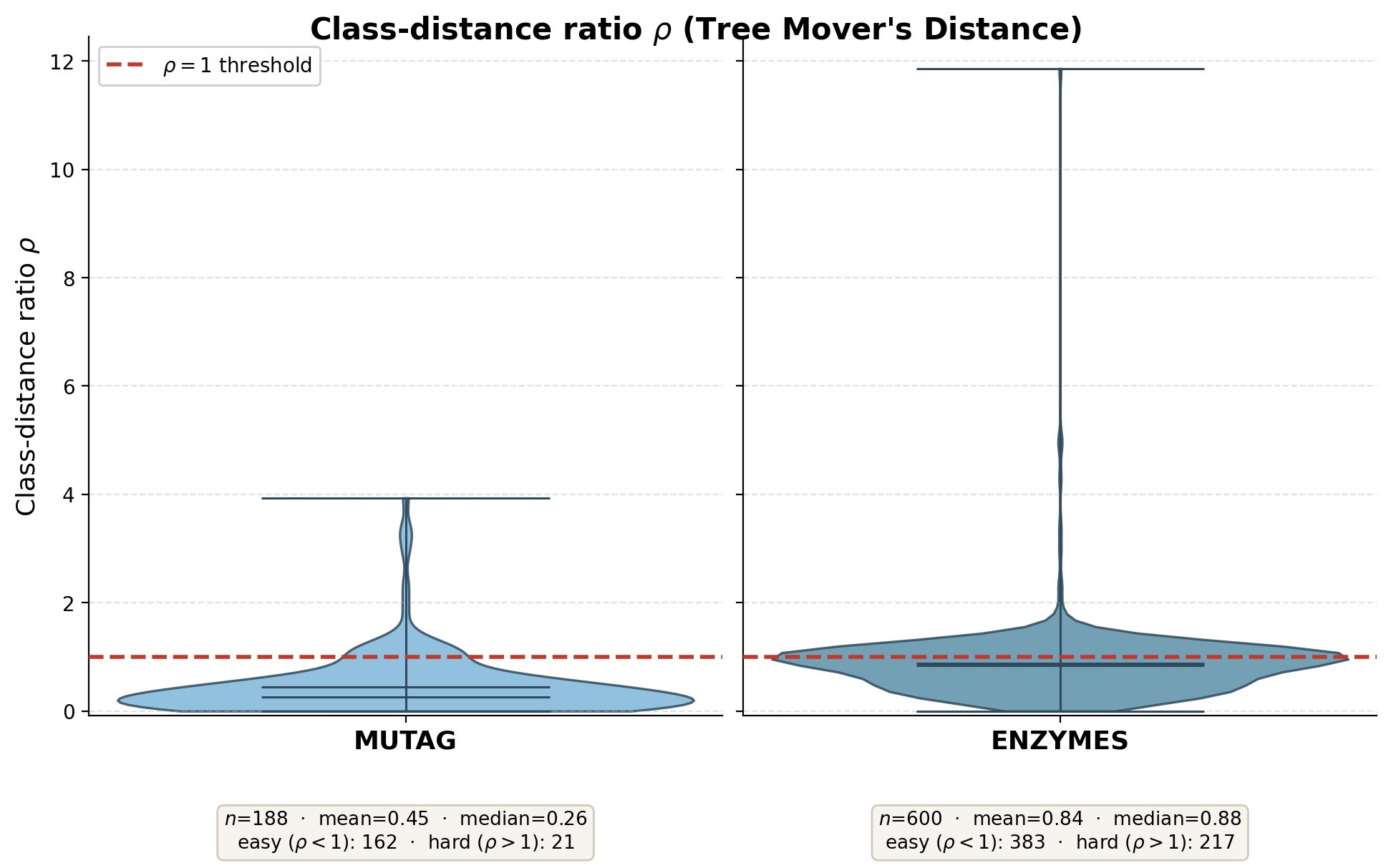}
    \caption{
        Class-distance ratios for MUTAG and ENZYMES computed using the Tree
        Mover's Distance. Several graphs are closer (using TMD) to graphs from another
        class than to graphs sharing their own label.
    }
    \label{fig:class-distance-ratios}
\end{figure}

\begin{figure}[h]
    \centering
    \includegraphics[width=1.0\textwidth]{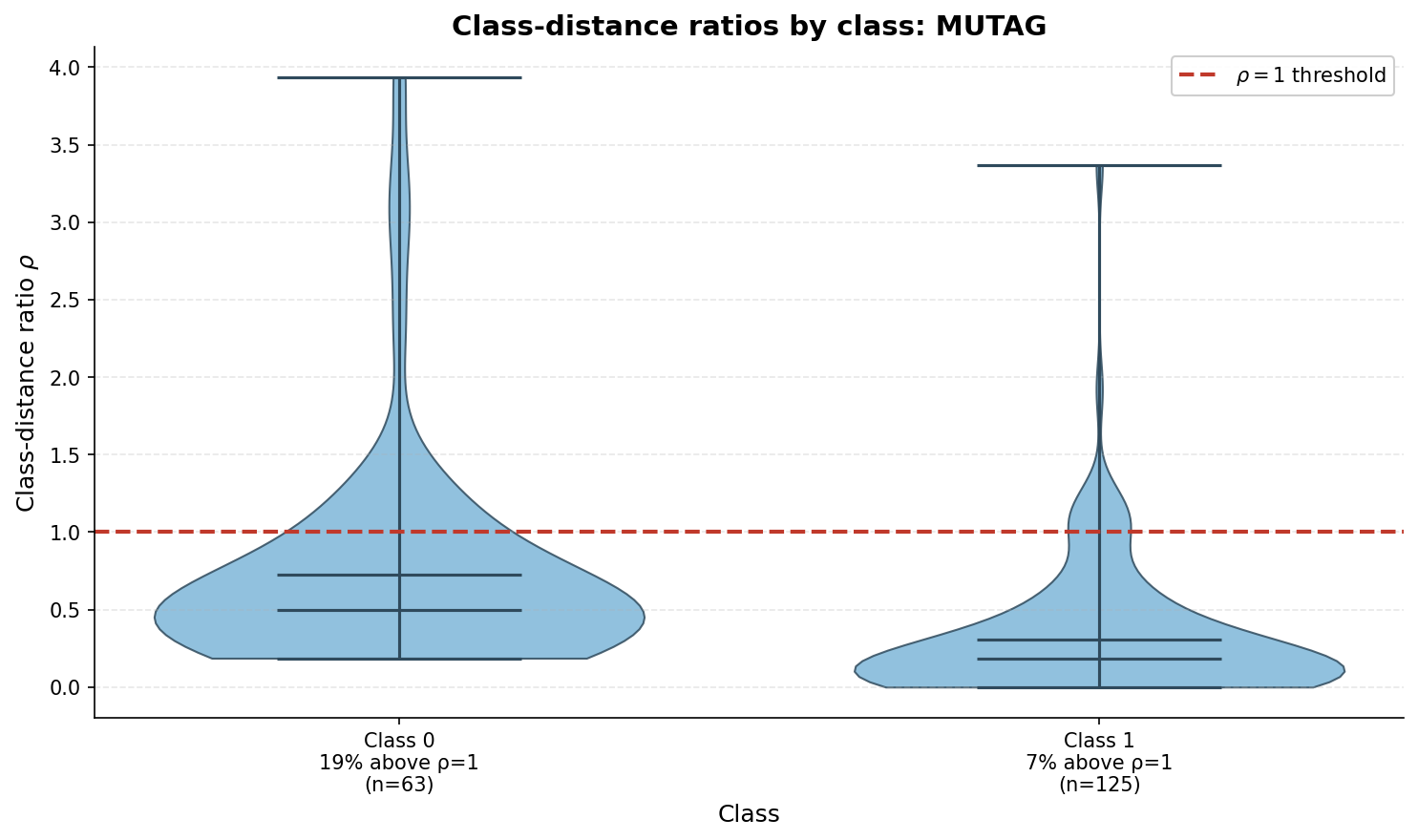}
    \caption{
        Class-distance ratios for MUTAG by label. Graphs in class 0 are harder to classify.
    }
    \label{fig:class-distance-ratios-mutag-by-class}
\end{figure}

\begin{figure}[h]
    \centering
    \includegraphics[width=1.0\textwidth]{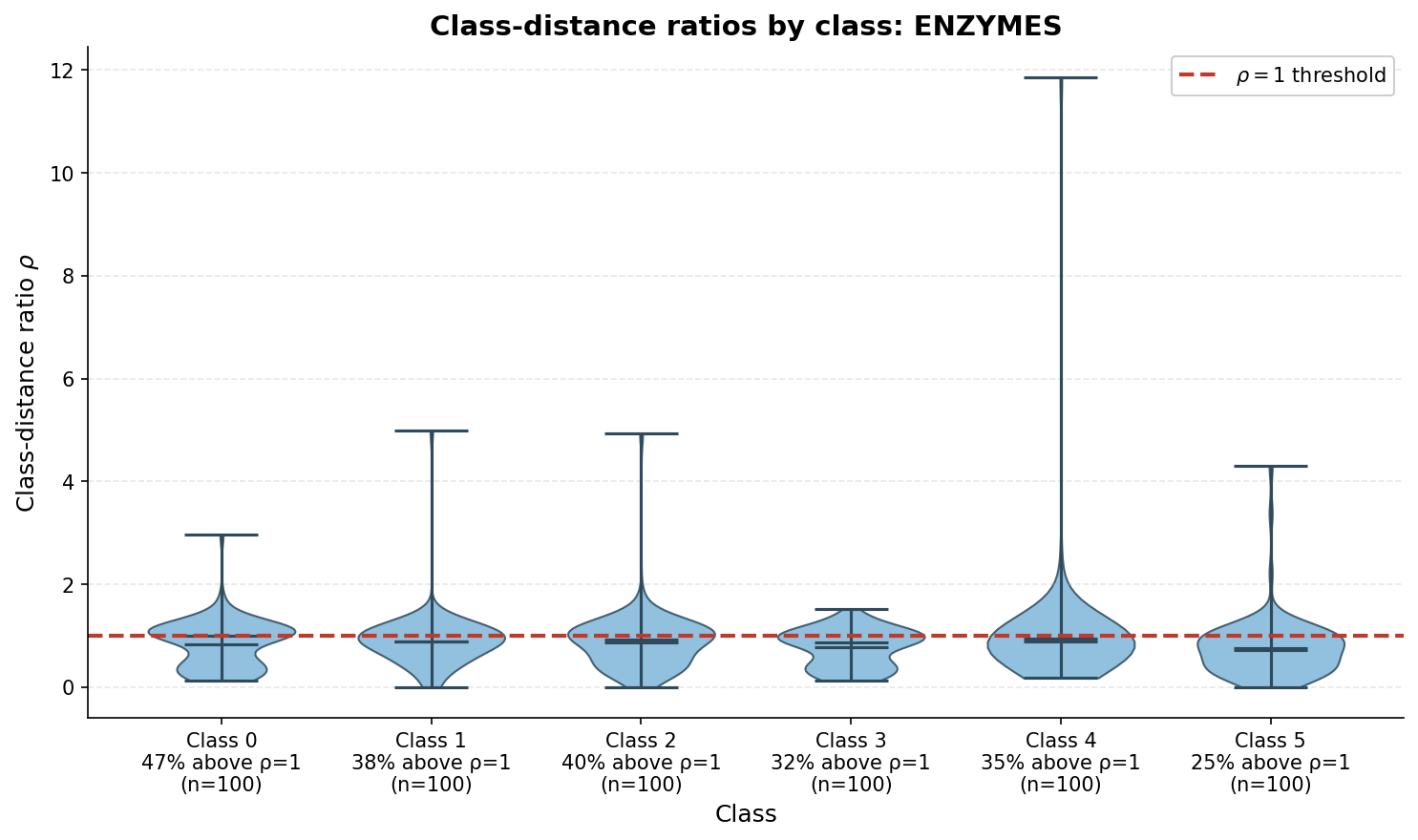}
    \caption{
        Class-distance ratios for ENZYME by label. Graphs in class 0 are harder to classify.
    }
    \label{fig:class-distance-ratios-enzymes-by-class}
\end{figure}

As shown in Fig.~\ref{fig:class-distance-ratios}, both MUTAG and ENZYMES
contain graphs with ratios substantially larger than one. The class-distance
ratio is negatively correlated with average GNN accuracy, with Pearson
correlations of \(-0.672\) on MUTAG and \(-0.0373\) on ENZYMES: graphs that are
relatively close to another class are more difficult to classify. This is
consistent with stability results for message-passing GNNs, which bound the
distance between graph representations in terms of TMD \cite{chuang2022tree}. The analysis therefore
suggests that graph-level heterogeneity is partly driven by class overlap in a
joint structural and feature space, rather than by topology alone.

\begin{figure}[h]
    \centering
    \includegraphics[width=1.0\textwidth]{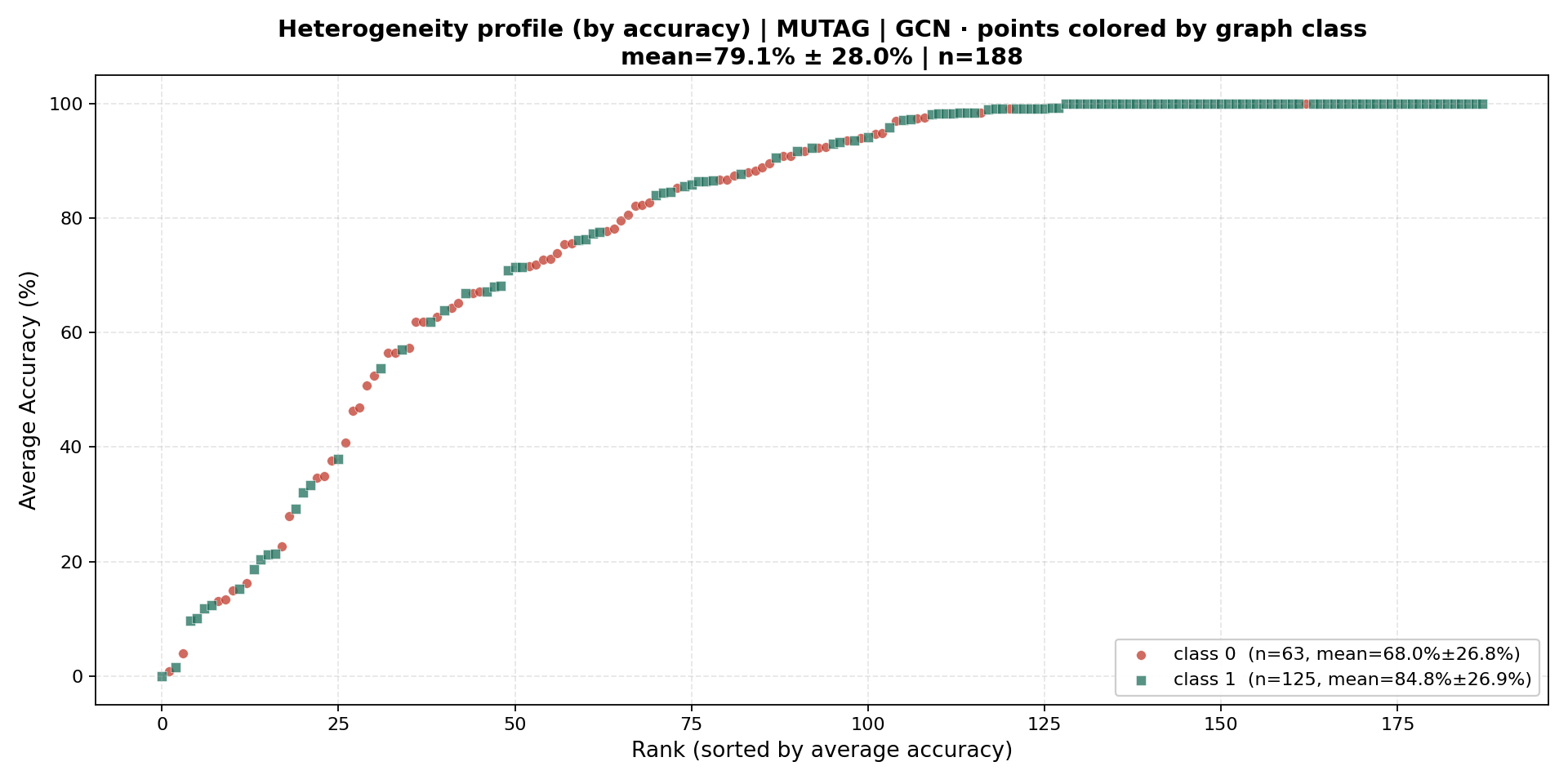}
    \caption{Heterogeneity profile of GCN on MUTAG. We note that class 0 is harder to classify than class 1. 
    }
    \label{fig:gcn-mutag-by-class}
\end{figure}

\begin{figure}[h]
    \centering
    \includegraphics[width=1.0\textwidth]{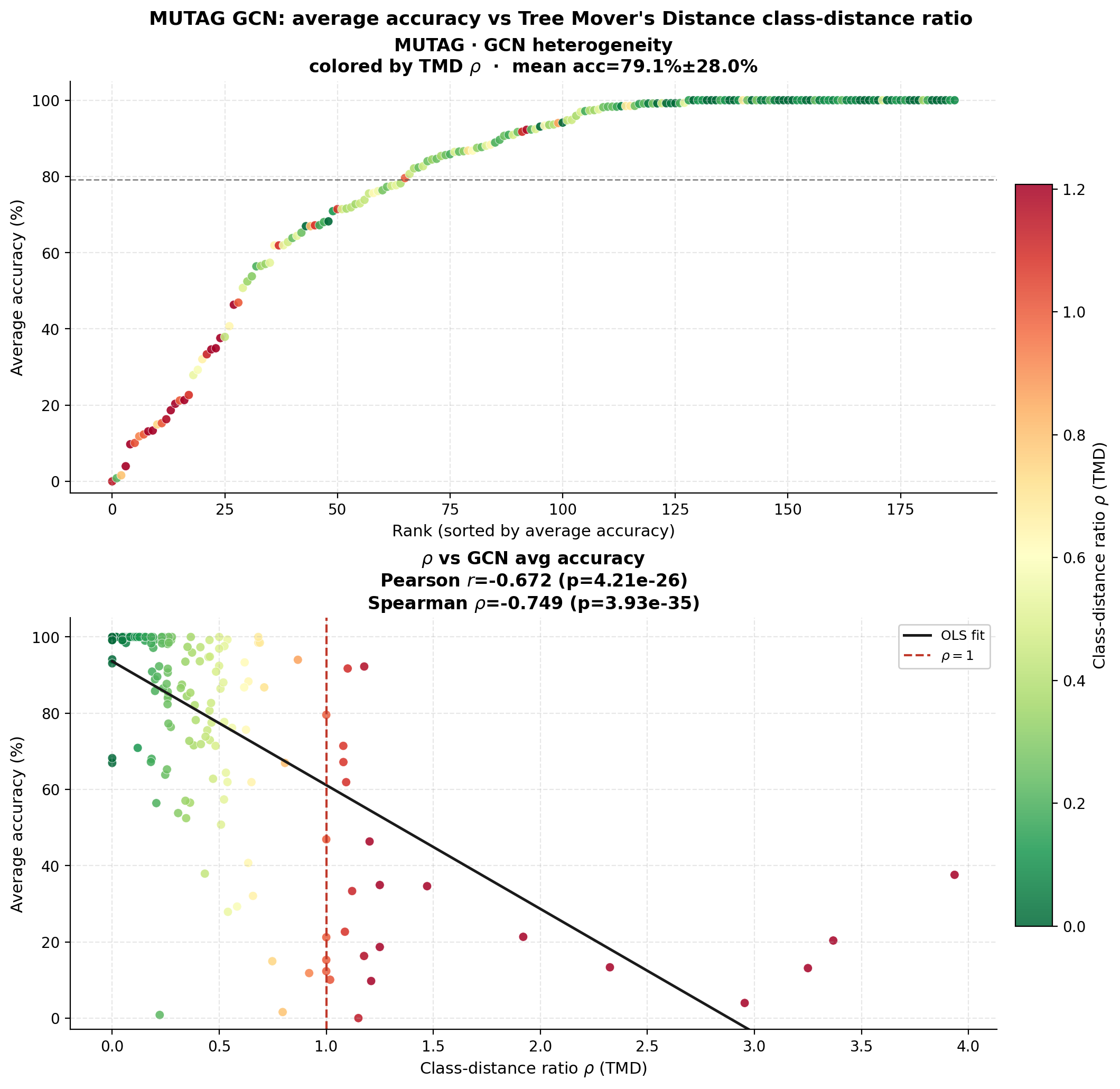}
    \caption{Heterogeneity profile of GCN on MUTAG and average accuracy against Class-distance ratio. Best seen in colour. The colorbar is clipped to the 95th percentile of $\rho$.
    } 
    \label{fig:gcn-mutag-by-class-augmented}
\end{figure}

\begin{figure}[h]
    \centering
    \includegraphics[width=1.0\textwidth]{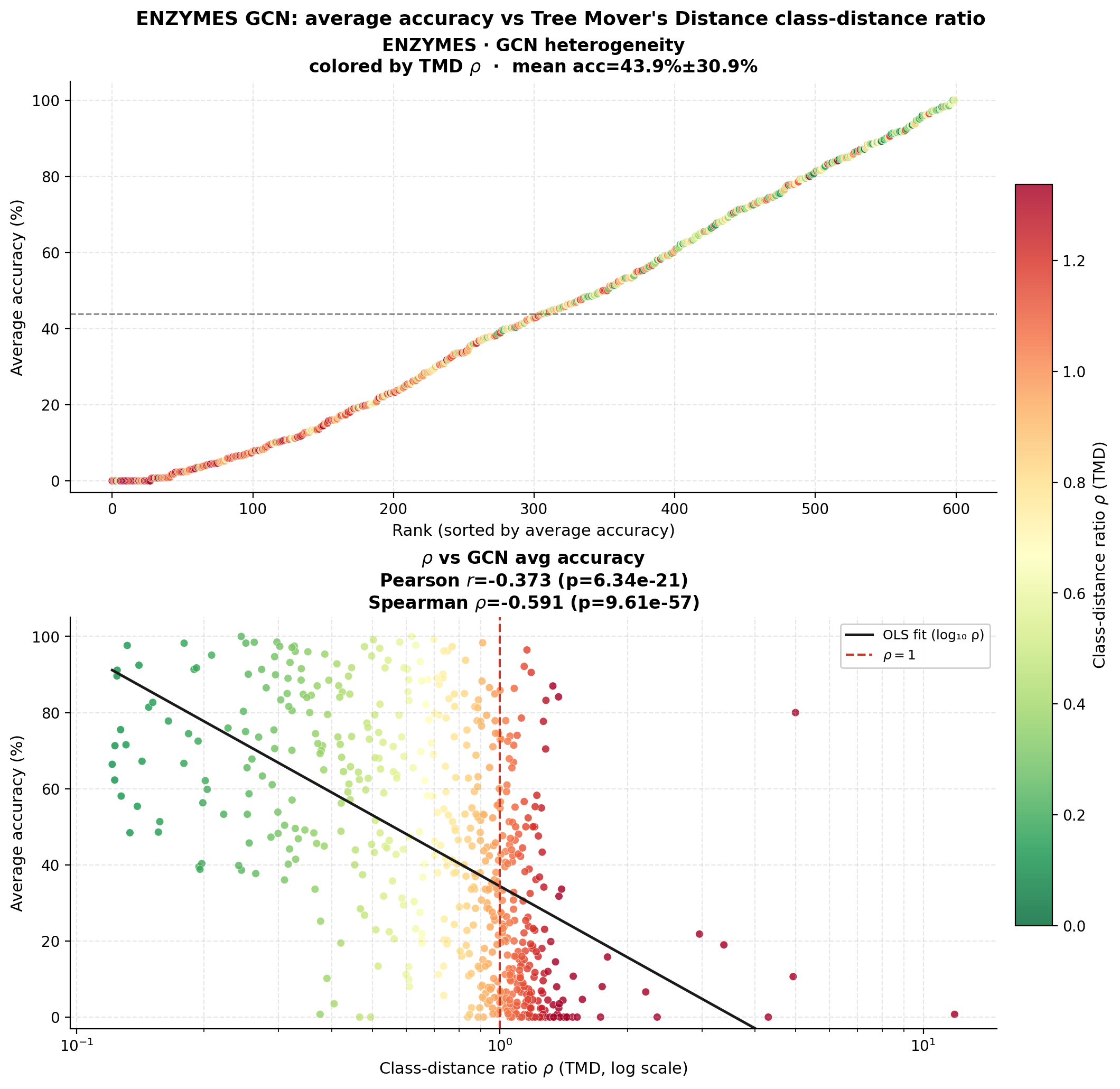}
    \caption{Heterogeneity profile of GCN on ENZYMES and average accuracy against Class-distance ratio. Best seen in colour. The colorbar is clipped to the 95th percentile of $\rho$. Note that on the bottom subplot, the x-axis is in log-scale. 
    }
    \label{fig:gcn-enzymes-by-class-dr}
\end{figure}

Fig.~\ref{fig:class-distance-ratios-mutag-by-class} and Fig.~\ref{fig:class-distance-ratios-enzymes-by-class} presents the class distance of graphs in MUTAG and ENZYMES by class. In MUTAG, graphs in Class $1$ tend to have (on average) lower class distance ratios. Furthermore, only $7\%$ of graphs in class $1$ have a  class distance ratio above $1$, compared to $19\%$ of graphs in class $0$. We thus expect graphs in class $0$ to be harder to classify: this intuition is verified by computing the mean accuracy by class over the runs we use to build the heterogeneity profile for MUTAG, illustrated in Fig.~\ref{fig:gcn-mutag-by-class} and Fig.~\ref{fig:gcn-mutag-by-class-augmented}. In ENZYMES, graphs in Class $5$ tend to have (on average) lower class distance ratios than graphs in Class $0$. $25\%$ of graphs in class $5$ have a  class distance ratio above $1$, compared to $47\%$ of graphs in class $0$. We thus expect graphs in class $0$ to be harder to classify than graphs in Class $5$. For ENZYMES, Fig.~\ref{fig:gcn-enzymes-by-class-dr} mirror the analysis we perform in Fig.~\ref{fig:gcn-mutag-by-class-augmented} for MUTAG. We note that even for SiGMA, graphs in class $0$ tend to be harder to classify than graphs in Class $5$ (see Fig.~\ref{fig:gcn-sigma-enzymes-by-class}).

\clearpage
\section{Model Details}
\label{app:model-details} 

\subsection{Elementwise Gating}
\label{app:element-wise-gating}

Here, we outline SiGMA's architecture when using elementwise gating. The differences between headwise gating and elementwise gating are summarized in App.~\ref{app:sigma_gate_modes}. The only difference is the dimension of the gates $\mathbf{g}_i$.

\paragraph{Elementwise-gated attention heads.} For an attention head $a\in\mathcal{H}_{\mathrm{att}}^\ell$, the only departure of SiGMA from standard attention is the gate, which is computed jointly
with the query:
\begin{equation}
\mathbf{W}_{qg}^a\widehat{ \mathbf{h}}_i^\ell
=
\left[
\mathbf{q}_i^a \;\Vert\; \mathbf{g}_i^{\mathrm{att},a}
\right] \in \mathbb{R}^{2\cdot d_h},
\qquad
\mathbf{k}_j^a=\mathbf{W}_k^a\widehat{ \mathbf{h}}_j^\ell \in \mathbb{R}^{d_h},
\qquad
\mathbf{v}_j^a=\mathbf{W}_v^a\widehat{ \mathbf{h}}_j^\ell \in \mathbb{R}^{d_h}.
\end{equation}
The head output is
\begin{equation}
\widetilde{\mathbf{a}}_i^{\mathrm{att},a}
=
\sigma(\mathbf{g}_i^{\mathrm{att},a})
\odot
\sum_{j\in V}
\alpha_{ij}^a \mathbf{v}_j^a \in \mathbb{R}^{d_h},
\qquad
\alpha_{ij}^a
=
\operatorname{softmax}_{j\in V}
\left(
(\mathbf{q}_i^a)^\top \mathbf{k}_j^a/\sqrt{d_h}+\mathbf{B}_{ij}
\right),
\end{equation}
where $\mathbf{B}_{ij}$ masks disallowed attention entries, including attention between different graphs in a batch.

\paragraph{Elementwise-gated message-passing heads.} For a message-passing head $p\in\mathcal{H}_{\mathrm{mp}}^\ell$, SiGMA computes a head-specific representation and gate,
\begin{equation}
\mathbf{W}_{hg}^p\widehat{ \mathbf{h}}_i^\ell
=
\left[
\mathbf{u}_i^p \;\Vert\; \mathbf{g}_i^{\mathrm{mp},p}
\right] \in \mathbb{R}^{2 \cdot d_h},
\end{equation}
then applies the chosen graph operator $\Phi_p$ and gates its
output:
\begin{equation}
\widetilde{\mathbf{a}}_i^{\mathrm{mp},p}
=
\sigma(\mathbf{g}_i^{\mathrm{mp},p})
\odot
\Phi_p\!\left(\{\mathbf{u}_j^p\}_{j\in V},E\right)_i  \in \mathbb{R}^{d_h}.
\end{equation}

The gated attention and message-passing heads are concatenated, projected,
and added through a residual update:
\begin{equation}
\mathbf{\Delta}_i^\ell
=
\mathbf{W}_{\mathrm{out}}^\ell
\left[
\big\Vert_{a\in\mathcal{H}_{\mathrm{att}}^\ell}
\widetilde{\mathbf{a}}_i^{\mathrm{att},a}
\;\Big\Vert\;
\big\Vert_{p\in\mathcal{H}_{\mathrm{mp}}^\ell}
\widetilde{\mathbf{a}}_i^{\mathrm{mp},p}
\right],
\qquad
\mathbf{h}_i^{\ell+1}=\mathbf{h}_i^\ell+\mathbf{\Delta}_i^\ell .
\end{equation}
The gated MPNN and attention block may be followed by the usual feed-forward, normalization, and dropout components.

Note that $\widetilde{\mathbf{a}}_i^{\mathrm{att},a}$ and $\widetilde{\mathbf{a}}_i^{\mathrm{mp},p}$ are both
in $\mathbb{R}^{d_h}$. One could also use separate head widths $d_h^{\mathrm{att}}$ and $d_h^{\mathrm{mp}}$ for attention and message-passing heads; in practice, we use a single shared width $d_h$ for both.

\subsection{Comparison of Gate Modes}
\label{app:sigma_gate_modes}
Let $\gamma_i$ denote the multiplicative modulator applied to a head output at
node $i$.
\begin{itemize}
  \item \textbf{Headwise}:
  $g_i\in\mathbb{R}$ and $\gamma_i=\sigma(g_i)\in(0,1)$ scales all $d_h$
  channels equally.
  \item \textbf{Elementwise}:
  $\mathbf{g}_i\in\mathbb{R}^{d_h}$ and $\boldsymbol{\gamma}_i=\sigma(\mathbf{g}_i)\in(0,1)^{d_h}$ scales
  channels independently via the Hadamard product with the head output.
\end{itemize}

In addition, in Tab.~\ref{tab:architecture-ablations} we consider the following case:
\begin{itemize}
  \item \textbf{Ungated}:
  no learned gate is predicted, and $\gamma_i\equiv 1$ (heads contribute at full
  scale). Attention and MP may also be gated asymmetrically
  (e.g.\ gated attention with ungated MPGNN heads, or the reverse). See App.~\ref{app:additional-models} for more details.
\end{itemize}

All three regimes share the same head geometry and fusion rule; only the
definition of $\gamma_i$ changes.

\subsection{MPGNN Head Types}
\label{app:mpgnn-heads}

Each message-passing head $p\in\mathcal{H}_{\mathrm{mp}}^\ell$ applies a
graph operator $\Phi_p$ to the projected head features
$\{\mathbf{u}_j^p\}_{j\in V}$ introduced above, producing an output in
$\mathbb{R}^{d_h}$ that is then modulated by the gate.
We instantiate $\Phi_p$ with standard MPGNN layers
(GCN, GIN, GraphSAGE, GAT, UniConv). Below, we recall their update rules; unless noted otherwise, they are applied
at head width $d_h$ (after the $\mathbf{W}_{hg}^p$ projection).

\subsubsection{GCN} 

GCN \cite{Kipf:2017tc} extends convolutional neural networks to graph-structured data. It derives a shared representation by integrating node features and graph connectivity through message-passing. Mathematically, a GCN layer is expressed (in matrix form) as  
\[
\mathbf{H}^{l+1} = \sigma \left( \tilde{\mathbf{D}}^{-1/2} \tilde{\mathbf{{A}}} \tilde{\mathbf{D}}^{-1/2} \mathbf{H}^{l} \mathbf{W}^{l+1} \right) \; ,
\]
where \(\mathbf{W}^{l+1}\) is the learnable weight matrix at layer \(l+1\), and  
$\tilde{\mathbf{D}}^{-1/2} \tilde{\mathbf{{A}}} \tilde{\mathbf{D}}^{-1/2}$ is the normalized adjacency matrix of the original graph with added self-loops. This graph has adjacency matrix $\tilde{\mathbf{A}}=\mathbf{A}+\mathbf{I}_n$, where $n$ is the number of nodes, and node degree matrix $\mathbf{\tilde{D}}$. The activation function \(\sigma\) is typically chosen as ReLU or a sigmoid function. 






\subsubsection{GCNE}

Graph Convolution with Edge Features (GCNE)~\cite{luo2025can} augments
message passing by adding edge attributes into neighbor messages.
Given node features $\mathbf{H}^{\ell}$ and edge features $\mathbf{e}_{uv}$
(projected to the node/head dimension when needed), a linear map
$\mathbf{H}'=\mathbf{H}^{\ell}\mathbf{W}^{\ell}$ is applied first.
Messages are then formed as
\begin{equation}
\mathbf{m}_{u\to v}
=
\mathrm{ReLU}\!\big(
\mathbf{h}_u'
+
\mathbf{e}_{uv}
\big),
\label{eq:gcne_msg}
\end{equation}
and aggregated over neighbors of $v$.
The resulting node representations may be followed by batch normalization,
a nonlinearity, residual connections, and a feed-forward block, as in the
GNN$^{+}$ backbone.

\subsubsection{GIN} 

GIN~\cite{xu2018powerful} is a MPGNN designed for maximum expressiveness, meaning it can learn a broader range of structural patterns compared to other MPGNNs like GCN. GIN is inspired by the Weisfeiler-Leman (WL) graph isomorphism test. Formally, the GIN layer is given (in node notation) by  
\begin{equation}
\mathbf{h}_i^{l+1} = \text{MLP}^{l} \left((1+\epsilon^{l+1})\cdot \mathbf{h}_i^{l}+ 
\sum_{j \in \mathcal{N}_i} \mathbf{h}_j^{l} \right)\; ,
\end{equation}
where \(\mathbf{x}_i^{l}\) denotes the feature of node \(i\) at layer \(l\), \(\mathcal{N}_i\) represents the neighbors of node \(i\), and \(\epsilon^{l+1}\) is a learnable parameter. The update step is carried out using a multi-layer perceptron \(\text{MLP}(\cdot)\), which is a fully connected neural network.

\subsubsection{GINE} 

Graph Isomorphism Networks with Edge features (GINE)~\cite{feng2022powerful,kiani2024unitary, luo2025can} extends the injective neighborhood aggregation of GIN to edges.
Denoting by $\mathrm{MLP}^{l}$ a multilayer perceptron and by $\epsilon^{l}$ a (learnable) scalar, node features are updated as
\begin{equation}
\mathbf{h}_v^{l+1}
=
\mathrm{MLP}^{l}\!
\left(
(1+\epsilon^{l+1})\cdot\,\mathbf{h}_v^{l}
+
\sum_{u\in\mathcal{N}_v}
\mathrm{ReLU}\!\big(
\mathbf{h}_u^{l}+\mathbf{e}_{uv}
\big)
\right),
\label{eq:gine}
\end{equation}
where $\mathbf{e}_{uv}$ is the feature of edge $(u,v)$ (projected to the node-feature dimension when needed).

\subsubsection{GAT} 

GAT~\cite{velivckovic2017graph} replaces fixed neighborhood weights by learned attention coefficients.
The update for a single head is given by:
\begin{equation}
\alpha_{vu}
=
\frac{
\exp\!\big(
\mathrm{LeakyReLU}\big(
\mathbf{a}^{\top}[\mathbf{W}\mathbf{h}_v^{(l)}\,\|\,\mathbf{W}\mathbf{h}_u^{(l)}]
\big)
\big)
}{
\sum_{u'\in\mathcal{N}_v\cup\{v\}}
\exp\!\big(
\mathrm{LeakyReLU}\big(
\mathbf{a}^{\top}[\mathbf{W}\mathbf{h}_v^{(l)}\,\|\,\mathbf{W}\mathbf{h}_{u'}^{(l)}]
\big)
\big)
},
\label{eq:gat_attn}
\end{equation}
and
\begin{equation}
\mathbf{h}_v^{(l+1)}
=
\sigma\!\left(
\sum_{u\in\mathcal{N}(v)\cup\{v\}}
\alpha_{vu}\,\mathbf{W}\mathbf{h}_u^{(l)}
\right),
\label{eq:gat}
\end{equation}
where $\mathbf{a}$ and $\mathbf{W}$ are trainable, and $\sigma$ is a nonlinearity.

\subsubsection{GraphSAGE}

GraphSAGE~\cite{hamilton2017inductive} updates each node by aggregating neighbor features and combining them with the node's own representation.
With the mean aggregator, the update is
\begin{equation}
\mathbf{h}_v^{(l+1)}
=
\sigma\!\left(
\mathbf{W}_1\mathbf{h}_v^{(l)}
+
\mathbf{W}_2\cdot
\mathrm{MEAN}\!\big\{\mathbf{h}_u^{(l)}:u\in\mathcal{N}(v)\big\}
\right),
\label{eq:sage}
\end{equation}
where $\mathbf{W}_1,\mathbf{W}_2$ are trainable and $\sigma$ is a nonlinearity.

\subsubsection{GatedGCN}

GatedGCN~\cite{bresson2017residual, luo2025can} augments message passing with
dense, edge-conditioned gates.
Given node features $\mathbf{H}^{\ell}$ and edge features $\mathbf{e}_{vu}$
(projected to the node and edge dimension when needed), for each edge $(v,u)$, the gate is
\begin{equation}
\boldsymbol{\eta}_{vu}
=
\sigma\!\left(
\mathbf{D}\mathbf{h}_v^{\ell}
+
\mathbf{E}\mathbf{h}_u^{\ell}
+
\mathbf{C}\mathbf{e}_{vu}
\right),
\label{eq:gatedgcn_gate}
\end{equation}
with $\sigma$ the sigmoid.
Neighbor messages are soft-normalized and aggregated as
\begin{equation}
\widetilde{\mathbf{h}}_v
=
\frac{
\sum_{u\in\mathcal{N}(v)}
\boldsymbol{\eta}_{vu}\odot
\bigl(\mathbf{B}\mathbf{h}_u^{\ell}\bigr)
}{
\sum_{u\in\mathcal{N}(v)}
\boldsymbol{\eta}_{vu}
+
\varepsilon
},
\qquad
\mathbf{h}_v^{\ell+1}
=
\mathbf{A}\mathbf{h}_v^{\ell}
+
\widetilde{\mathbf{h}}_v,
\label{eq:gatedgcn}
\end{equation}
where $\varepsilon>0$ is a small stabilizer.
Edge representations are updated in parallel from the pre-sigmoid logits.








\subsection{Attention Head Types}

\subsubsection{GRIT}
When using GRIT~\cite{ma2023graph}, each attention head uses
\emph{edge-aware} attention with Relative Random Walk Probabilities (RRWP),
rather than plain scaled-dot-product QK attention.
Given node states $\mathbf{h}_i,\mathbf{h}_j$ and a pairwise edge
representation $\mathbf{e}_{ij}$ (initialized from RRWP and projected to the
model/head width when needed), GRIT forms
\begin{equation}
\hat{\mathbf{e}}_{ij}
=
\sigma\!\Bigl(
\rho\!\bigl(
(\mathbf{W}_{Q}\mathbf{h}_i+\mathbf{W}_{K}\mathbf{h}_j)
\odot
\mathbf{W}_{EW}\mathbf{e}_{ij}
\bigr)
+
\mathbf{W}_{Eb}\mathbf{e}_{ij}
\Bigr),
\qquad
\alpha_{ij}
=
\mathrm{softmax}_{j\in\mathcal{N}(i)}\!\bigl(
\mathbf{W}_{A}\hat{\mathbf{e}}_{ij}
\bigr),
\end{equation}
and aggregates values over the sparse neighborhood
\begin{equation}
\mathbf{h}_i'
=
\sum_{j\in\mathcal{N}(i)}
\alpha_{ij}\,
\mathbf{W}_{V}\mathbf{h}_j
\end{equation}
(optionally with an edge-enhanced residual term).
Here $\rho$ is a signed square-root
nonlinearity, and $\sigma$ is a pointwise activation (ReLU).
Unlike vanilla Transformer attention, the scores depend explicitly on
learned relative structure, which injects graph inductive bias without a
separate message-passing module inside the attention head~\citep{ma2023graph}.

\subsection{Complexity and Runtime}
\label{app:complexity}

Let $n$ and $m$ denote the numbers of nodes and edges, $d$ the hidden dimension, $d_h$ the head dimension, and $A$ and $P$ the numbers of attention and message-passing heads. The input and output projections of a SiGMA layer require $\mathcal{O}\!\left(n(A+P)dd_h\right)$ operations. Full attention contributes $\mathcal{O}\!\left(An^2d_h\right)$ time and $\mathcal{O}\!\left(An^2\right)$ memory, while message-passing head $p$ contributes the cost of its underlying operator $\Phi_p$; for common sparse operators, neighborhood aggregation requires $\mathcal{O}\!\left(md_h\right)$ operations per head. For batches, the quadratic attention term depends on the sum of squared graph sizes rather than the square of the total number of nodes. Computing and applying the sigmoid gates adds only node-wise projections, sigmoid evaluations, and multiplications, and therefore does not change these asymptotic bounds. SiGMA consequently has the same asymptotic complexity as an ungated hybrid architecture with the same heads. 

We also compare SiGMA's runtime with that of an ungated hybrid model. Concretely, we time one training epoch of GraphGPS and SiGMA on the same stack and the same GPU. Both models use four layers of width 64, dropout 0, and no added positional encoding. Each layer runs one GatedGCN and one Transformer head over all nodes in the batch, masked to the graph each node belongs to, followed by the same feed-forward network. We use AdamW with learning rate $10^{-3}$. The clock covers the forward pass, the backward pass, and the optimizer step, and does not include validation. The first epoch is discarded. Table~\ref{tab:gps-sigma-time} reports the mean and sample standard deviation of the next three epochs on one H100. Batch size is chosen so the dense attention maps fit in memory. With headwise gating, SiGMA is about $15\%$ larger than GraphGPS and $6$--$18\%$ slower per training epoch.

\begin{table}[t]
  \centering
  \caption{Training-epoch wall time on one H100. Entries are the mean $\pm$ sample standard deviation, in seconds, of three epochs after one warmup epoch.}
  \label{tab:gps-sigma-time}
  \begin{tabular}{lcccc}
    \toprule
    Dataset & Batch & GraphGPS & SiGMA & SiGMA\,/\,GraphGPS \\
    \midrule
    MNIST           & 16 & $32.90 \pm 0.16$ & $35.12 \pm 0.07$ & 1.068 \\
    CIFAR10         & 16 & $41.91 \pm 0.08$ & $44.69 \pm 0.02$ & 1.066 \\
    PATTERN         & 16 & $6.79 \pm 0.22$  & $7.34 \pm 0.02$  & 1.081 \\
    CLUSTER         & 16 & $6.41 \pm 0.12$  & $7.28 \pm 0.03$  & 1.136 \\
    Peptides-func   &  8 & $14.31 \pm 0.05$ & $16.87 \pm 0.09$ & 1.179 \\
    Peptides-struct &  8 & $14.16 \pm 0.02$ & $16.73 \pm 0.06$ & 1.181 \\
    PascalVOC-SP    &  4 & $18.94 \pm 0.02$ & $20.14 \pm 0.05$ & 1.064 \\
    MalNet-Tiny     &  1 & $34.84 \pm 0.15$ & $40.42 \pm 0.90$ & 1.160 \\
    \bottomrule
  \end{tabular}
\end{table}

\clearpage
\section{Additional Models}
\label{app:additional-models} 

In addition to SiGMA, we introduce two variants, \underline{M}ixture of \underline{M}essage-Passing and \underline{A}ttention (MMA) and SiGMA-lite. Like SiGMA, these models can have an arbitrary number of MP heads and attention heads: the difference with SiGMA is that in MMA, we do not apply any gating, whereas SiGMA-lite replaces the per-head gates by a single gate on the mixed update of all heads, as in Gated-GPS~\cite{gao2025gated}. The differences between the architectures are outlined in Tab.~\ref{tab:sigma-variants}. We show in App.~\ref{app:routing analysis} that SiGMA can outperform MMA on synthetic routing tasks, while on real benchmarks, the differences in models' performances are often not statistically significant. We do not tune hyperparameters for MMA or SiGMA-lite: both are run as ablations that reuse the configuration selected for SiGMA, either removing the gates (MMA) or moving a single gate after the output projection (SiGMA-lite). Characterizing when gating is necessary is an interesting direction for future work.

\begin{table}[h]
\centering
\caption{Architectural differences between SiGMA and its two variants. All three use the same attention and message-passing heads, pre-normalization, residual
connection, and post-block FFN; they differ only in whether, and where, the sigmoid gate is applied.
$\mathbf a_i^r$ is the output of head $r\in\mathcal H^\ell=\mathcal H^\ell_{\mathrm{att}}\cup\mathcal H^\ell_{\mathrm{mp}}$,
$\mathbf m_i^\ell$ the mixed update of Eq.~\eqref{eq:mma}, $d$ the model width, and $d_h$ the head width. ''a$N$g$M$'' denotes $N$ attention heads and $M$ MP heads per hybrid block.}
\label{tab:sigma-variants}
\small
\renewcommand{\arraystretch}{1.25}
\begin{tabularx}{\linewidth}{@{}>{\raggedright\arraybackslash}p{2.5cm}
  >{\raggedright\arraybackslash}X >{\raggedright\arraybackslash}X >{\raggedright\arraybackslash}X@{}}
\toprule
 & \textbf{SiGMA} & \textbf{MMA} & \textbf{SiGMA-lite} \\
\midrule
Layer update
 & $\mathbf h_i^\ell + \mathbf W_{\mathrm{out}}^\ell\big[\Vert_{r}\,\gamma_i^r\,\mathbf a_i^r\big]$
 & $\mathbf h_i^\ell + \mathbf W_{\mathrm{out}}^\ell\big[\Vert_{r}\,\mathbf a_i^r\big]$
 & $\mathbf h_i^\ell + \mathbf W_1^\ell\big(\mathbf m_i^\ell \odot \boldsymbol\gamma_i^l\big)$ \\
Gate
 & one per head, $\gamma_i^r=\sigma(g_i^r)$
 & none 
 & one per layer\\
Gate position
 & on each head output, before concatenation 
 & --
 & on the mixed update, after $\mathbf W_{\mathrm{out}}^\ell$ \\
Granularity
 & headwise (scalar) or elementwise ($d_h$ per head)
 & --
 & elementwise ($d$), shared by all heads \\
Can down-weight one head at one node
 & yes
 & no
 & no, each gate entry scales a sum over all heads \\
Extra parameters vs.\ MMA
 & $|\mathcal H^\ell|(d{+}1)$ headwise; $|\mathcal H^\ell|\,d_h(d{+}1)$ otherwise
 & --
 & $2d(d{+}1)$ ($\mathbf W_g^\ell$, $\mathbf W_1^\ell$) \\
Closest prior architecture
 & --
 & MPGNN (a0g1), GTs (a$N$g0), GraphGPS (a$N$g1)
 & Gated-GPS~\cite{gao2025gated} \\
\bottomrule
\end{tabularx}
\end{table}

\subsection{MMA}
\label{app:MMA}

MMA removes the gates, i.e.\ fixes
$\gamma_i^r \equiv 1$ in~\eqref{eq:sigma_update} for all heads $r$. The MMA layer contains a set of attention heads $\mathcal{H}_{\mathrm{att}}^\ell$ and a set of message-passing heads $\mathcal{H}_{\mathrm{mp}}^\ell$ (both potentially empty). Each head produces a $d_h$-dimensional output. For an attention head $a \in \mathcal{H}_{\mathrm{att}}^\ell$:
\begin{equation}
\mathbf{q}_i^a  = \mathbf{W}_{q}^a\widehat{ \mathbf{h}}_i^\ell \in \mathbb{R}^{d_h},
\qquad
\mathbf{k}_j^a=\mathbf{W}_k^a\widehat{ \mathbf{h}}_j^\ell \in \mathbb{R}^{d_h},
\qquad
\mathbf{v}_j^a=\mathbf{W}_v^a\widehat{ \mathbf{h}}_j^\ell \in \mathbb{R}^{d_h}.
\end{equation}
The head output is
\begin{equation}
\mathbf{a}_i^{\mathrm{att},a}
=
\sum_{j\in V}
\alpha_{ij}^a \mathbf{v}_j^a \in \mathbb{R}^{d_h},
\qquad
\alpha_{ij}^a
=
\operatorname{softmax}_{j\in V}
\left(
(\mathbf{q}_i^a)^\top \mathbf{k}_j^a/\sqrt{d_h}+\mathbf{B}_{ij}
\right),
\end{equation}
where $\mathbf{B}_{ij}$ masks disallowed attention entries, including attention between different graphs in a batch. For a message-passing head $p\in\mathcal{H}_{\mathrm{mp}}^\ell$, MMA computes a head-specific representation,
\begin{equation}
\mathbf{W}_{h}^p\widehat{ \mathbf{h}}_i^\ell
=
\mathbf{u}_i^p   \in \mathbb{R}^{d_h},
\end{equation}
then applies the chosen graph operator $\Phi_p$:
\begin{equation}
\mathbf{a}_i^{\mathrm{mp},p}
=
\Phi_p\!\left(\{\mathbf{u}_j^p\}_{j\in V},E\right)_i  \in \mathbb{R}^{d_h}.
\end{equation}

The attention and message-passing heads are concatenated, projected,
and added through a residual update:
\begin{equation}
\label{eq:mma}
\mathbf{h}_i^{\ell+1}=\mathbf{h}_i^\ell+\mathbf{m}_i^\ell,
\qquad
\mathbf{m}_i^\ell
=
\mathbf{W}_{\mathrm{out}}^\ell
\left[
\big\Vert_{a\in\mathcal{H}_{\mathrm{att}}^\ell}
\mathbf{a}_i^{\mathrm{att},a}
\;\Big\Vert\;
\big\Vert_{p\in\mathcal{H}_{\mathrm{mp}}^\ell}
\mathbf{a}_i^{\mathrm{mp},p}
\right].
\end{equation}

\begin{figure}[h]
    \centering
    \includegraphics[width=\textwidth]{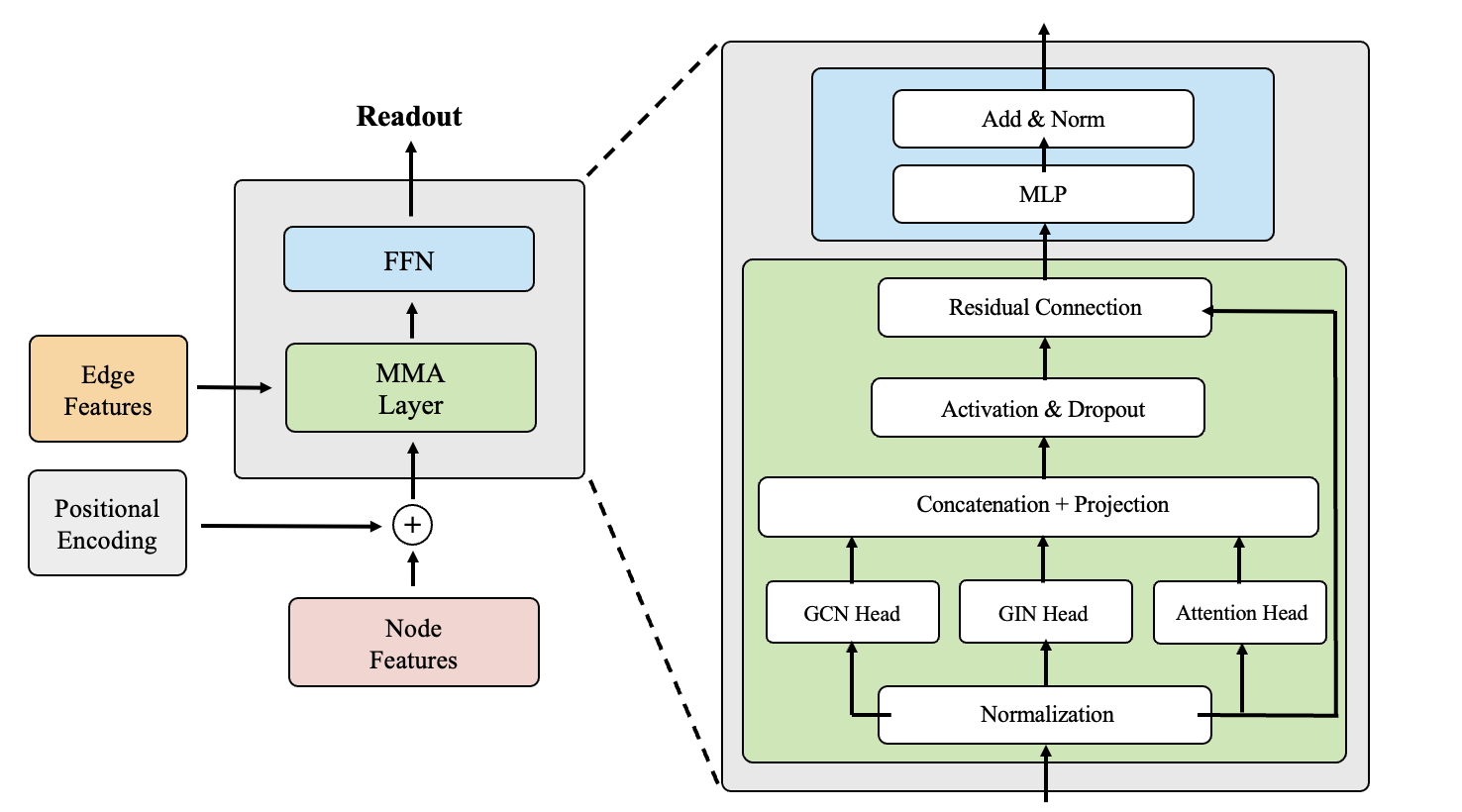}
    \caption{\textbf{The MMA architecture.} Node features, positional encodings, and edge features are processed by a stack of MMA layers followed by feed-forward blocks. Each MMA layer normalizes node representations, applies one or more parallel message-passing (here GCN and GIN) and attention heads (here only one), concatenates and projects the heads, and adds the projected update through a residual connection.}
    \label{fig:mma_architecture}
    \vspace{-5mm}
\end{figure}

With the surrounding layer structure (normalization, residual connection and FFN) held fixed,
MMA contains several standard architectures as special cases: a single MP head (MMA (a$0$g$1$)) gives an
MPGNN\footnote{Up to the projections $\mathbf{W}_h^p$ and $\mathbf{W}_{\text{out}}$.}, and one MP head plus one attention head (MMA (a$1$g$1$)) gives a GraphGPS-style
layer. MMA also covers layers with several, possibly different, MP heads, which neither MPGNNs nor GraphGPS
include. Fig.~\ref{fig:mma_architecture} shows the entire MMA architecture.

\subsection{SiGMA-lite}
\label{app:sigma-lite}

SiGMA-lite keeps the heads ungated, as in MMA, and instead applies a single elementwise gate to the
mixed update $\mathbf{m}_i^{\ell}$ of~\eqref{eq:mma}, in the spirit of Gated-GPS~\cite{gao2025gated}
but with any number of attention and MP heads:

\begin{equation}
\label{eq:sigma-lite}
\mathbf{h}_i^{\ell+1} = \mathbf{h}_i^{\ell} + \mathbf{W}^{\ell}_{1}\big(\mathbf{m}_i^{\ell}\odot\boldsymbol{\gamma}_i^{\ell}\big),
\qquad
\boldsymbol{\gamma}_i^{\ell} = \sigma\big(\mathbf{W}^{\ell}_{g}\,\mathrm{Norm}(\mathbf{h}_i^{\ell})\big) \in \mathbb{R}^{d}.
\end{equation}


\clearpage
\section{Theoretical Properties}
\label{app:proofs}

We analyze the stylized update \eqref{eq:sigma_update}, with the pre-head normalization $\mathrm{Norm}$ optional; the activation after the output projection, feed-forward blocks, other normalization layers, and dropout are omitted, while the residual connection in \eqref{eq:sigma_update} is required. Gates are written elementwise, $\mathbf{s}_i \odot \mathbf{a}_i$ with $\mathbf{s}_i \in (0,1)^{d_h}$; headwise gating is the special case of equal coordinates.

\subsection{Expressivity}

\begin{proposition}[Expressivity inherited from message passing]
Consider an $L$-layer SiGMA network whose message-passing heads include,
in each layer $\ell$, a head capable, together with the output projection,
of realizing the residual increment of the $\ell$-th update of some
message-passing GNN $\mathcal{M}$. Assume this realization is possible with
the normalization and feature projection applied before the head, that the
same input embedding and graph-level readout can be used, and that any
additional sublayers can be omitted or set to identity. Then every graph
function representable by $\mathcal{M}$ is also representable by SiGMA.
Consequently, SiGMA is at least as expressive as the most expressive
message-passing architecture satisfying these conditions.
\end{proposition}

\begin{proof}
We make explicit the realization condition used in the statement. Let
$\mathcal{M}$ be an $L$-layer message-passing GNN with node representations
$\mathbf{z}_i^\ell\in\mathbb{R}^d$, the same input embedding as SiGMA, and the same
graph-level readout. Write the $\ell$-th layer of $\mathcal{M}$ in residual
form as
\[
\mathbf{z}_i^{\ell+1}
=
\mathbf{z}_i^\ell + F_\ell(\mathbf{Z}^\ell,E)_i,
\]
where $\mathbf{Z}^\ell=(\mathbf{z}_i^\ell)_{i\in V}$. For a non-residual message-passing layer
$\mathbf{z}_i^{\ell+1}=T_\ell(\mathbf{Z}^\ell,E)_i$, this corresponds to taking
$F_\ell(\mathbf{Z}^\ell,E)_i=T_\ell(\mathbf{Z}^\ell,E)_i-\mathbf{z}_i^\ell$.
Thus the realization requirement is that the selected SiGMA message-passing
head, together with the output projection, can realize the residual increment
of the corresponding layer of $\mathcal{M}$, including the normalization and
feature projection before the head. Any additional sublayers are omitted or
set to identity.

By assumption, for each layer $\ell$ there is a message-passing head
$p_\ell\in\mathcal{H}_{\mathrm{mp}}^\ell$ and a linear map
$\mathbf{R}_\ell\in\mathbb{R}^{d\times d_h}$ such that its ungated output
$\mathbf{a}_i^{\mathrm{mp},p_\ell}$ satisfies
\[
\mathbf{R}_\ell \mathbf{a}_i^{\mathrm{mp},p_\ell}
=
F_\ell(\mathbf{H}^\ell,E)_i
\]
whenever $\mathbf{H}^\ell=\mathbf{Z}^\ell$. We choose the gate parameters of this selected
head so that
\[
\mathbf{g}_i^{\mathrm{mp},p_\ell}=\mathbf{0}
\qquad\text{for all } i\in V,
\]
and hence
\[
\sigma(\mathbf{g}_i^{\mathrm{mp},p_\ell})
=
\frac{1}{2}\mathbf{1}.
\]
We then set the block of $\mathbf{W}_{\mathrm{out}}^\ell$ corresponding to the
selected head equal to $2\mathbf{R}_\ell$, and set all blocks corresponding to every
other attention or message-passing head equal to zero. The other heads may
therefore produce arbitrary outputs, but they do not contribute to the
residual update. For the selected head, the gated contribution is
\[
2\mathbf{R}_\ell
\left(
\sigma(\mathbf{g}_i^{\mathrm{mp},p_\ell})
\odot
\mathbf{a}_i^{\mathrm{mp},p_\ell}
\right)
=
2\mathbf{R}_\ell
\left(
\frac{1}{2}\mathbf{a}_i^{\mathrm{mp},p_\ell}
\right)
=
F_\ell(\mathbf{H}^\ell,E)_i .
\]
Thus the SiGMA residual update satisfies
\[
\mathbf{h}_i^{\ell+1}
=
\mathbf{h}_i^\ell+\mathbf{\Delta}_i^\ell
=
\mathbf{h}_i^\ell+F_\ell(\mathbf{H}^\ell,E)_i .
\]

We now prove equality of the SiGMA and $\mathcal{M}$ representations by
induction over layers. The input embeddings are chosen to be the same, so
$\mathbf{H}^0=\mathbf{Z}^0$. If $\mathbf{H}^\ell=\mathbf{Z}^\ell$, the construction above gives
\[
\mathbf{h}_i^{\ell+1}
=
\mathbf{h}_i^\ell+F_\ell(\mathbf{H}^\ell,E)_i
=
\mathbf{z}_i^\ell+F_\ell(\mathbf{Z}^\ell,E)_i
=
\mathbf{z}_i^{\ell+1}
\]
for every node $i$. Hence $\mathbf{H}^{\ell+1}=\mathbf{Z}^{\ell+1}$. By induction,
$\mathbf{H}^L=\mathbf{Z}^L$. Since the graph-level readout is also chosen to be the same,
SiGMA represents exactly the same graph function as $\mathcal{M}$.

Therefore every graph function representable by any message-passing GNN
whose residual increments can be instantiated by selectable SiGMA heads
under these conditions is also representable by SiGMA. In particular,
SiGMA is at least as expressive as the most expressive message-passing
architecture satisfying these conditions.
\end{proof}

\subsection{Over-smoothing}

\begin{proposition}[No forced oversmoothing]
Consider an $L$-layer SiGMA network. Assume each gate logit can be set to an
arbitrarily negative constant independently of the ungated head computations
(e.g., through a gate bias). Under uniform boundedness assumptions on the
ungated head outputs and output projections, for every graph $G$, depth $L$,
and $\varepsilon>0$, there exist gate values such that
$\max_{i\in V}\|\mathbf{h}_i^L-\mathbf{h}_i^0\|_2\leq\varepsilon$.
\end{proposition}

\begin{proof}
We prove the claim for the hybrid SiGMA update. Any optional
feed-forward block with a residual connection can be omitted, or its residual
branch can be set to zero, and therefore does not affect the near-identity
construction.

Let
\[
\mathcal{H}^\ell
=
\mathcal{H}_{\mathrm{att}}^\ell
\cup
\mathcal{H}_{\mathrm{mp}}^\ell
\]
denote the set of heads in layer $\ell$, and write $\mathbf{a}_i^{\ell,r}$ for the
ungated output of head $r\in\mathcal{H}^\ell$ at node $i$. Its gated output
has the form
\[
\widetilde{\mathbf{a}}_i^{\ell,r}
=
\mathbf{s}_i^{\ell,r}\odot \mathbf{a}_i^{\ell,r},
\qquad
\mathbf{s}_i^{\ell,r}\in(0,1)^{d_h},
\]
where $\mathbf{s}_i^{\ell,r}$ is the corresponding sigmoid gate. The layer update is
\[
\mathbf{h}_i^{\ell+1}
=
\mathbf{h}_i^\ell+\mathbf{\Delta}_i^\ell,
\qquad
\mathbf{\Delta}_i^\ell
=
\mathbf{W}_{\mathrm{out}}^\ell
\left[
\big\Vert_{r\in\mathcal{H}^\ell}
\widetilde{\mathbf{a}}_i^{\ell,r}
\right].
\]

Assume, as in the proposition, that the ungated head outputs and output
projections are uniformly bounded. That is, for each layer $\ell$ there are
finite constants $A_\ell$ and $W_\ell$ such that
\[
\|\mathbf{a}_i^{\ell,r}\|_2\leq A_\ell
\quad
\text{and}
\quad
\|\mathbf{W}_{\mathrm{out}}^\ell\|_{\mathrm{op}}\leq W_\ell
\]
for all nodes $i$ and heads $r\in\mathcal{H}^\ell$, with constants independent
of the gate values used below. Let
$m_\ell=|\mathcal{H}^\ell|$ and define
\[
R
=
\sum_{\ell=0}^{L-1}
W_\ell A_\ell \sqrt{m_\ell}.
\]
If $R=0$, then every residual update is zero and the result is immediate.
Otherwise, fix $\varepsilon>0$ and choose the gate values so that every
coordinate of every gate satisfies
\[
0\leq \mathbf{s}_i^{\ell,r}\leq \eta,
\qquad
\eta
=
\min\left\{1,\frac{\varepsilon}{R}\right\}.
\]
By the assumed gate parameterization, such values can be obtained by taking
the corresponding constant sigmoid gate logits sufficiently negative.

For each node $i$ and layer $\ell$, we then have
\[
\|\widetilde{\mathbf{a}}_i^{\ell,r}\|_2
=
\|\mathbf{s}_i^{\ell,r}\odot \mathbf{a}_i^{\ell,r}\|_2
\leq
\eta A_\ell .
\]
Therefore the concatenated gated head vector satisfies
\[
\left\|
\big\Vert_{r\in\mathcal{H}^\ell}
\widetilde{\mathbf{a}}_i^{\ell,r}
\right\|_2
\leq
\eta A_\ell \sqrt{m_\ell},
\]
and hence
\[
\|\mathbf{\Delta}_i^\ell\|_2
\leq
W_\ell A_\ell \sqrt{m_\ell}\,\eta .
\]
Summing the residual updates over layers gives
\[
\|\mathbf{h}_i^L-\mathbf{h}_i^0\|_2
\leq
\sum_{\ell=0}^{L-1}
\|\mathbf{\Delta}_i^\ell\|_2
\leq
\eta
\sum_{\ell=0}^{L-1}
W_\ell A_\ell \sqrt{m_\ell}
\leq
\varepsilon .
\]
Thus
\[
\max_{i\in V}\|\mathbf{h}_i^L-\mathbf{h}_i^0\|_2\leq \varepsilon .
\]

It remains to show that the Dirichlet energy can be kept arbitrarily close
to its initial value. Let
\[
\mathbf{\delta}_i=\mathbf{h}_i^L-\mathbf{h}_i^0 .
\]
From the bound above, $\|\mathbf{\delta}_i\|_2\leq\varepsilon$ for all $i\in V$.
For each edge $(i,j)\in E$, write
\[
\mathbf{d}_{ij}^0=\mathbf{h}_i^0-\mathbf{h}_j^0,
\qquad
\mathbf{d}_{ij}^L=\mathbf{h}_i^L-\mathbf{h}_j^L
=
\mathbf{d}_{ij}^0+(\delta_i-\delta_j).
\]
Since $\|\delta_i-\delta_j\|_2\leq 2\varepsilon$, we have
\[
\left|
\|\mathbf{d}_{ij}^L\|_2^2-\|\mathbf{d}_{ij}^0\|_2^2
\right|
\leq
2\|\mathbf{d}_{ij}^0\|_2\|\delta_i-\delta_j\|_2
+
\|\delta_i-\delta_j\|_2^2
\leq
4\varepsilon \|\mathbf{h}_i^0-\mathbf{h}_j^0\|_2
+
4\varepsilon^2 .
\]
Using the definition
\[
\mathcal{E}_G(\mathbf{H})
=
\frac{1}{2}
\sum_{(i,j)\in E}
\|\mathbf{h}_i-\mathbf{h}_j\|_2^2,
\]
we obtain
\[
\left|
\mathcal{E}_G(\mathbf{H}^L)-\mathcal{E}_G(\mathbf{H}^0)
\right|
\leq
2\varepsilon
\sum_{(i,j)\in E}
\|\mathbf{h}_i^0-\mathbf{h}_j^0\|_2
+
2|E|\varepsilon^2 .
\]
For a fixed graph and fixed initial representation, the right-hand side
goes to zero as $\varepsilon\to 0$. Equivalently, for $\varepsilon\leq 1$,
\[
\left|
\mathcal{E}_G(\mathbf{H}^L)-\mathcal{E}_G(\mathbf{H}^0)
\right|
\leq
C_{G,H^0}\varepsilon
\]
with
\[
C_{G,H^0}
=
2
\sum_{(i,j)\in E}
\|\mathbf{h}_i^0-\mathbf{h}_j^0\|_2
+
2|E|.
\]
Thus SiGMA contains arbitrarily deep near-identity parameter regimes whose
Dirichlet energy remains arbitrarily close to that of the input
representation. In particular, increasing the nominal depth does not by
itself force Dirichlet-energy collapse or oversmoothing.
\end{proof}

\subsection{Over-squashing}

\begin{proposition}[No forced over-squashing]
Consider a SiGMA layer containing at least one attention head whose mask
allows the entry $(i,j)$, with $i\neq j$. Under suitable nondegeneracy
assumptions on the weight matrices, normalization, and input embedding,
$\mathbf{h}_i^{\ell+1}$ can depend directly on $\mathbf{h}_j^\ell$, and hence
the influence $S_{ij}^{\ell+1}$ need not vanish even when
$\operatorname{dist}_G(i,j)>\ell+1$.
\end{proposition}

\begin{proof}
We prove the statement in two steps. First, we recall the hard locality
constraint for purely local message passing. Let $N_r(i)$ denote the
$r$-hop neighborhood of node $i$. In any one-hop message-passing network,
the representation of node $i$ after $r$ layers is a function only of the
initial features of nodes in $N_r(i)$. This follows by induction. For
$r=0$, $\mathbf{h}_i^0$ depends only on $\mathbf{x}_i$. If $\mathbf{h}_k^r$ depends only on
$\{\mathbf{x}_u : u\in N_r(k)\}$, then a one-hop update at node $i$ depends only on
the representations of nodes $k\in N(i)\cup\{i\}$, and hence only on
\[
\bigcup_{k\in N(i)\cup\{i\}} N_r(k)
\subseteq
N_{r+1}(i).
\]
Therefore, if $\operatorname{dist}_G(i,j)>r$, then $\mathbf{h}_i^r$ is independent
of $\mathbf{x}_j$, and
\[
\frac{\partial \mathbf{h}_i^r}{\partial \mathbf{x}_j}=\mathbf{0}.
\]
Equivalently, the pairwise influence $S_{ij}^r$ vanishes outside the
$r$-hop neighborhood.

Second, consider an attention head $a$ and distinct nodes $i,j$ for which
the fixed attention mask allows the entry $(i,j)$. Let $\mathcal{A}_i$
denote the set of nodes to which node $i$ is allowed to attend. Choose
$\mathbf{W}_k^a=\mathbf{0}$, so that the attention weights are independent
of the node representations and $\alpha_{ij}^a>0$. Set the gate-projection
rows and any gate biases to zero, so that $\mathbf{s}_i^a=\frac12\mathbf{1}$.
Write $\widehat{\mathbf{h}}_j^\ell=\operatorname{Norm}(\mathbf{h}_j^\ell)$
for the node-wise normalization and let
\[
\mathbf{D}_j^\ell
=
\frac{\partial\widehat{\mathbf{h}}_j^\ell}{\partial\mathbf{h}_j^\ell}
\]
at a point where this derivative exists. The contribution of this head is
\[
\boldsymbol{\Delta}_{i,a}^\ell
=
\mathbf{W}_{\mathrm{out},a}^\ell
\left(
\mathbf{s}_i^a
\odot
\sum_{r\in\mathcal{A}_i}
\alpha_{ir}^a \mathbf{W}_v^a \widehat{\mathbf{h}}_r^\ell
\right).
\]
Since the attention weights and gate are constant with respect to the node
representations, we have
\[
\frac{\partial \boldsymbol{\Delta}_{i,a}^\ell}
{\partial \mathbf{h}_j^\ell}
=
\alpha_{ij}^a
\mathbf{W}_{\mathrm{out},a}^\ell
\operatorname{diag}(\mathbf{s}_i^a)
\mathbf{W}_v^a
\mathbf{D}_j^\ell .
\]
For finite gate logits, every coordinate of $\mathbf{s}_i^a$ is strictly positive.
Thus, under the mild nondegeneracy condition
\[
\mathbf{W}_{\mathrm{out},a}^\ell
\operatorname{diag}(\mathbf{s}_i^a)
\mathbf{W}_v^a
\mathbf{D}_j^\ell
\neq \mathbf{0},
\]
the Jacobian block
$\partial \boldsymbol{\Delta}_{i,a}^\ell/\partial \mathbf{h}_j^\ell$ is nonzero.
We may set the output blocks corresponding to all other heads to zero, so
no other head is needed for this direct dependence. Hence the layer map satisfies
\[
\frac{\partial \mathbf{h}_i^{\ell+1}}{\partial \mathbf{h}_j^\ell}
\neq \mathbf{0}.
\]
Therefore $\mathbf{h}_i^{\ell+1}$ can depend directly on $\mathbf{h}_j^\ell$ in a single
layer, regardless of the graph distance between $i$ and $j$, provided that
the attention mask allows the entry $(i,j)$.

To connect this dependence to the original input, choose all preceding
layers to be identity maps by setting their output projections to zero and
omitting or setting to identity any additional sublayers. Then
$\mathbf{h}_j^\ell=\mathbf{h}_j^0=e(\mathbf{x}_j)$, where $e$ is the
node-wise input embedding. Let
$\mathbf{J}_j=\partial e(\mathbf{x}_j)/\partial\mathbf{x}_j$ at a point
where this derivative exists. The chain-rule terms involving other nodes
vanish, giving
\[
\frac{\partial \mathbf{h}_i^{\ell+1}}{\partial \mathbf{x}_j}
=
\frac{\partial \mathbf{h}_i^{\ell+1}}{\partial \mathbf{h}_j^\ell}
\frac{\partial \mathbf{h}_j^\ell}{\partial \mathbf{x}_j}
=
\alpha_{ij}^a
\mathbf{W}_{\mathrm{out},a}^\ell
\operatorname{diag}(\mathbf{s}_i^a)
\mathbf{W}_v^a
\mathbf{D}_j^\ell\mathbf{J}_j.
\]
Under the nondegeneracy condition
\[
\mathbf{W}_{\mathrm{out},a}^\ell
\operatorname{diag}(\mathbf{s}_i^a)
\mathbf{W}_v^a
\mathbf{D}_j^\ell\mathbf{J}_j
\neq\mathbf{0},
\]
this derivative is nonzero. For an identity input embedding,
$\mathbf{J}_j=\mathbf{I}$ and this reduces to the earlier nondegeneracy
condition. Consequently, $S_{ij}^{\ell+1}$ can be nonzero even when
$\operatorname{dist}_G(i,j)>\ell+1$. Thus, unlike purely local one-hop message
passing, SiGMA is not forced to transmit long-range information only along
graph paths, and attention heads can provide direct paths that bypass local
graph bottlenecks associated with over-squashing.
\end{proof}

\clearpage
\section{Datasets}
\label{app:datasets}

We use a suite of graph-learning benchmarks spanning small bioinformatics graphs, large-scale molecular and protein datasets, and image-derived superpixel graphs. 

The GNN Benchmark suite of Dwivedi et al. \cite{JMLR:Dwivedi} provides ZINC with $12,000$ molecular graphs from the ZINC database, MNIST with $70,000$ superpixel graphs from handwritten digits, CIFAR10 with $60,000$ superpixel graphs from natural images, PATTERN with $14,000$ synthetic SBM graphs, CLUSTER with $12,000$ synthetic SBM graphs. 

The Long Range Graph Benchmark \cite{dwivedi2022LRGB} adds Peptides-func with $15,535$ peptide molecular graphs, Peptides-struct with the same $15,535$ graphs, and two node segmentation tasks: PascalVOC-SP with $11,355$ superpixel graphs from PASCAL VOC 2011, and COCO-SP with $123,286$ superpixel graphs.

We additionally use MalNet-Tiny which consists of $5,000$ function-call graphs. The task is malware classification.

From the TU Dortmund collection \cite{morris2020tudataset}, we use MUTAG, ENZYMES, PROTEINS, IMDB-BINARY, COLLAB and REDDIT-BINARY.

\begin{table}[h!]
\centering
\caption{Dataset statistics used in our experiments.
Avg.\ nodes/edges are per graph.
Task level indicates whether labels are assigned per graph or per node.
Splits are train/val/test graph counts (or percentages when fixed ratios).
For regression tasks we report the number of targets instead of classes.}
\label{tab:dataset-stats}
\small
\setlength{\tabcolsep}{3.5pt}
\begin{tabular}{lrrrccc}
\toprule
\textbf{Dataset} & \textbf{\# Graphs} & \textbf{Avg.\ Nodes} & \textbf{Avg.\ Edges} & \textbf{Level} & \textbf{\# Classes / targets} & \textbf{Split} \\
\midrule
\multicolumn{7}{l}{\textit{GNN Benchmark (Dwivedi et al.)}} \\
ZINC (12K subset)
  & 12{,}000 & 23.16 & 49.83 & Graph & 1 (regr.) & 10k/1k/1k \\
MNIST
  & 70{,}000 & 70.57 & 564.53 & Graph & 10 & 55k/5k/10k \\
CIFAR10
  & 60{,}000 & 117.63 & 941.07 & Graph & 10 & 45k/5k/10k \\
PATTERN
  & 14{,}000 & 117.47 & 4{,}749.15 & Node & 2 & 10k/2k/2k \\
CLUSTER
  & 12{,}000 & 117.20 & 4{,}301.72 & Node & 6 & 10k/1k/1k \\
\midrule
\multicolumn{7}{l}{\textit{Long Range Graph Benchmark (LRGB)}} \\
Peptides-func
  & 15{,}535 & 150.94 & 307.30 & Graph & 10 (multi-label) & 70/15/15 \\
Peptides-struct
  & 15{,}535 & 150.94 & 307.30 & Graph & 11 (regr.) & 70/15/15 \\
PascalVOC-SP
  & 11{,}355 & 479.40 & 2{,}710.48 & Node & 21 & 8{,}498/1{,}428/1{,}429 \\
COCO-SP
  & 123{,}286 & 476.88 & 2{,}693.67 & Node & 81 & 113{,}286/5k/5k \\
\midrule
\multicolumn{7}{l}{\textit{Malware}} \\
MalNet-Tiny
  & 5{,}000 & 1{,}410.31 & 2{,}859.94 & Graph & 5 & 70/10/20 \\
\midrule
\multicolumn{7}{l}{\textit{TU Dortmund (TUDataset)}} \\
MUTAG
  & 188 & 17.93 & 19.79 & Graph & 2 & 50/25/25 \\
ENZYMES
  & 600 & 32.63 & 62.14 & Graph & 6 & 50/25/25 \\
PROTEINS
  & 1{,}113 & 39.06 & 72.82 & Graph & 2 & 50/25/25 \\
IMDB-BINARY
  & 1{,}000 & 19.77 & 96.53 & Graph & 2 & 50/25/25 \\
COLLAB
  & 5{,}000 & 74.49 & 2{,}457.78 & Graph & 3 & 50/25/25 \\
REDDIT-BINARY
  & 2{,}000 & 429.63 & 497.75 & Graph & 2 & 50/25/25 \\
\bottomrule
\end{tabular}
\end{table}

\clearpage
\section{Experimental Details}
\label{app:experimental-details} 

\subsection{Training and Hyperparameters}

We report the  hyperparameters for SiGMA for Tab~\ref{tab:benchmark-results-1} and Tab~\ref{tab:lrgb-results} in Tab~\ref{tab:sigma_hparams_table3} and Tab~\ref{tab:sigma_hparams_table4} as well as Tab~\ref{tab:sigma_hparams_extra}, Tab~\ref{tab:sigma_hparams_extra-tab3} and Tab~\ref{tab:sigma_hparams_extra_tab4}.
All reported SiGMA models use the gated hybrid architecture presented in Fig.~\ref{fig:sigma_architecture}: parallel attention and message-passing heads,
each modulated by a learned sigmoid gate, followed by concatenation, projection,
and a residual update.
Unless noted, we use AdamW, cosine learning-rate decay with warmup, and select
the checkpoint with best validation performance. $d_h$ is the per-head width and $d$ the model hidden width.

\begin{table*}[h]
\centering
\caption{SiGMA hyperparameters for  Tab.~\ref{tab:benchmark-results-1}. ``a$N$g$M$'' denotes $N$ attention heads and $M$ MP heads per hybrid block.}
\label{tab:sigma_hparams_table3}
\resizebox{\textwidth}{!}{%
\begin{tabular}{lccccccccccccc}
\toprule
Dataset & Arch & MP types & Attention types & Gate & Norm & Attn mask & $L$ & $d$ & $d_h$ & LR & Warmup / epochs & Batch \\
\midrule
ZINC
  & a2g1 & GINE & GRIT$\times$2
  & headwise & RMS & graph
  & 12 & 80 & 80
  & $1\!\times\!10^{-3}$ & 50 / 1000 & 64 \\
MNIST
  & a2g2 & GATEDGCN$\times$2 & Vanilla
  & elementwise & LN & full
  & 6 & 60 & 64
  & $5\!\times\!10^{-4}$ & 5 / 200 & 16 \\
CIFAR10
  & a8g4 & GATEDGCN$\times$4 & Vanilla
  & headwise & LN & full
  & 10 & 35 & 256
  & $5\!\times\!10^{-4}$ & 10 / 400 & 16 \\
PATTERN
  & a2g2 & GCNE$\times$2 &  GRIT
  & elementwise & RMS & graph
  & 12 & 90 & 90
  & $2\!\times\!10^{-3}$ & 5 / 200 & 32 \\
CLUSTER
  & a0g2 & GATEDGCN$\times$2 & --
  & headwise & LN & full
  & 16 & 56 & 64
  & $1\!\times\!10^{-3}$ & 5 / 100 & 16 \\
\bottomrule
\end{tabular}%
}
\end{table*}

\begin{table*}[h]
\centering
\caption{SiGMA hyperparameters for Tab.~\ref{tab:lrgb-results}.}
\label{tab:sigma_hparams_table4}
\resizebox{\textwidth}{!}{%
\begin{tabular}{lccccccccccccc}
\toprule
Dataset & Arch & MP types & Attention types & Gate & Norm & Attn mask & $L$ & $d$ & $d_h$ & LR & Warmup / epochs & Batch \\
\midrule
Peptides-func
  & a1g2 & GCN$\times$2 & Vanilla
  & elementwise & RMS & graph
  & 3 & 275 & 128
  & $2\!\times\!10^{-4}$ & 10 / 900 & 16 \\
Peptides-struct
  & a1g1 & GINE & Vanilla
  & elementwise & LN & full
  & 5 & 200 & 200
  & $7\!\times\!10^{-4}$ & 5 / 250 & 32 \\
PascalVOC-SP
  & a2g2 & GATEDGCN$\times$2 & Vanilla
  & headwise & RMS & full
  & 16 & 95 & 64
  & $3\!\times\!10^{-4}$ & 10 / 200 & 8 \\
COCO-SP
  & a1g1 & GATEDGCN & Vanilla
  & headwise & LN & full
  & 20 & 52 & 52
  & $1\!\times\!10^{-3}$ & 10 / 300 & 16 \\
MalNet-Tiny
  & a1g2 & GATEDGCN$\times$2 & exphormer
  & headwise & RMS & graph
  & 5 & 80 & 80
  & $5\!\times\!10^{-4}$ & 10 / 150 & 32 \\
\bottomrule
\end{tabular}%
}
\end{table*}


\begin{table}[t]
\centering
\caption{Shared / dataset-specific training details for SiGMA (for Tab.~\ref{tab:benchmark-results-1} and Tab.~\ref{tab:lrgb-results}).}
\label{tab:sigma_hparams_extra}
\small
\begin{tabular}{ll}
\toprule
Setting & Value \\
\midrule
Optimizer & AdamW \\
LR scheduler & cosine with linear warmup (all rows above) \\
Gradient clipping & enabled \\
\bottomrule
\end{tabular}
\end{table}

\begin{table*}[t]
\centering
\caption{Dataset-specific SiGMA training settings for Tab.~\ref{tab:benchmark-results-1}.
Dropout is written as node~/~attn~/~MP.
FFN denotes the post-layer feed-forward around the hybrid block. Res denotes the post-layer residual around the hybrid block. Pool is the graph readout; Activation is the pointwise nonlinearity in the hybrid layers.}
\label{tab:sigma_hparams_extra-tab3}
\small
\setlength{\tabcolsep}{3.5pt}
\begin{tabular}{l*{5}{c}}
\toprule
 & \textbf{ZINC} & \textbf{MNIST} & \textbf{CIFAR10} & \textbf{PATTERN} & \textbf{CLUSTER} \\
\midrule
WD
  & $10^{-4}$ & $10^{-5}$ & $10^{-5}$ & $10^{-5}$ & $10^{-5}$ \\
FFN
  & \checkmark & \checkmark & -- & -- & -- \\
Res
  & \checkmark & \checkmark & \checkmark & \checkmark & \checkmark \\
Dropout
  & $-/0.1/0.05$ & $0.15/0.5/0.15$ & $0.15/0.2/0.1$ & $0.05/0.2/0.15$ & $0.2/0.1/0.2$ \\
Pool
  & add & mean & mean & -- & -- \\
Activation
  & ReLU & ReLU & ReLU & ReLU & ReLU \\
Node encodings
  & TypeDict+RWSE & Linear & Linear & Linear+RWSE & Linear+RWSE \\
Edge encodings
  & TypeDict+RWPE & Linear & Linear & Dummy & Dummy \\
\bottomrule
\end{tabular}
\end{table*}

\begin{table*}[t]
\centering
\caption{Dataset-specific SiGMA training settings for Tab.~\ref{tab:lrgb-results}.
Dropout is written as node~/~attn~/~MP.
FFN denotes the post-layer feed-forward around the hybrid block. Res denotes the post-layer residual around the hybrid block. Pool is the graph readout; Activation is the pointwise nonlinearity in the hybrid layers.}
\label{tab:sigma_hparams_extra_tab4}
\small
\setlength{\tabcolsep}{3.5pt}
\begin{tabular}{l*{5}{c}}
\toprule
 & \textbf{Pep-func} & \textbf{Pep-struct} & \textbf{VOC} & \textbf{COCO} & \textbf{MalNet} \\
\midrule
WD
  & $0$ & $0$ & $0$ & $0$ & $10^{-5}$ \\
FFN
  & -- & -- & \checkmark & -- & \checkmark \\
Res
  & -- & -- & \checkmark & \checkmark & \checkmark \\
Dropout
  & $0.2/0.1/0.2$ & $0.2/0.1/0.2$ & $0.1/0.1/0.15$ & $0.05/0.1/0.05$ & $0/0/0$ \\
Pool
  & mean & mean & -- & -- & max \\
Activation
  & GELU & GELU & GELU & GELU & ReLU \\
Node encodings
  & Atom+RWSE & Atom+RWSE & VOC+RWSE & COCO & Atom+LDP \\
Edge encodings
  & Bond & Bond & VOC & COCO & RWPE \\
\bottomrule
\end{tabular}
\end{table*}

\subsection{Compute and Parameter Counts}

We report the best SiGMA runs under a fixed budget (100K, 500K, 1M) in Tab.~\ref{tab:sigma_budget}.

\begin{table*}[h]
\centering
\caption{Best SiGMA performance under parameter budgets.
Higher is better except for Peptides-struct (MAE~$\downarrow$).
Main Tab.~\ref{tab:benchmark-results-1} and Tab~\ref{tab:lrgb-results} SiGMA used when the number of parameters already fits the budget; otherwise the best documented multi-seed alternative under the budget - no grid search is run at all for these models. To meet a budget we start from the main SiGMA recipe and shrink capacity: fewer attention and MP heads (typically to $a{=}1$, $g{\le}2$), then smaller model width and head dimension $d_h$ (and depth $L$ if needed) without a hyperparameter grid search.}
\label{tab:sigma_budget}
\small
\setlength{\tabcolsep}{2.5pt}
\resizebox{\textwidth}{!}{%
\begin{tabular}{@{}lcccccc@{}}
\toprule
\textbf{Budget}
  & \textbf{MNIST}$\uparrow$
  & \textbf{PATTERN}$\uparrow$
  & \textbf{CLUSTER}$\uparrow$
  & \textbf{Pep-func}$\uparrow$
  & \textbf{Pep-struct}$\downarrow$
  & \textbf{VOC}$\uparrow$ \\
\midrule
${\sim}100\mathrm{k}$
  & $98.54{\pm}0.15$
  & $86.90{\pm}0.03$
  & $78.10{\pm}0.11$
  & $0.6079{\pm}0.0130$
  & $0.2534{\pm}0.0009$
  & $0.3712{\pm}0.0084$ 
  \\
${\sim}500\mathrm{K}$
  & $98.63{\pm}0.11$ 
  & $87.01{\pm}0.09$ 
  & $78.79{\pm}0.09$
  & $0.7016{\pm}0.0086$
  & $0.2491{\pm}0.0012$
  & $0.4334{\pm}0.0130$ \\
${\sim}1\mathrm{M}$
  & $98.63{\pm}0.11$
  & $87.40{\pm}0.02$ 
  & $79.09{\pm}0.16$
  & $0.7080{\pm}0.0063$
  & $0.2491{\pm}0.0012$
  & $0.4398{\pm}0.0060$ \\
\bottomrule
\end{tabular}%
}
\end{table*}

\clearpage

\section{Additional Results on the TU Datasets}
\label{app:additional-results-TU}


We also evaluate SiGMA on standard TU graph-classification benchmarks with a random 50/25/25 train/val/test split, five random seeds, and a shared training recipe (the depth $L$ is set at 12, the model width $d$ at $64$, and we use AdamW). Dataset statistics are given in Tab~\ref{tab:dataset-stats}. We compare (i) a plain GCN; (ii) a plain GIN; (iii) a plain GraphSAGE; (iv) a plain GAT; (v) SiGMA with 2 attention heads and 4 identical GCN MP heads (GCN$\times$4), which we denote as SiGMA (homo); and (vi) SiGMA with the same 2 attention heads but heterogeneous MP heads (GCN, GIN, GraphSAGE, GAT), which we denote as SiGMA (hetero). Although SiGMA uses multiple heads, each operates at reduced width $d_h{=}16$ (with $d{=}64$), so the parameter count is only about $1.7\times$ that of a single GCN under the same depth and model width recipe, and is not proportional to the number of heads. We do not perform an extensive hyperparameter search: for SiGMA we try two learning rates ($10^{-3}$ and $10^{-2}$) and report the better learning rate per family.  Across most datasets, SiGMA improves over the plain MPGNNs, and the heterogeneous MPGNN mixture further improves over the homogeneous SiGMA variant (most clearly on MUTAG, COLLAB and REDDIT). The results are outlined in Tab.~\ref{tab:tu_dataset_GCN_vs_sigma}. We also run SiGMA with the same heads, but each operating at reduced width $d_h{=}4$ (with $d{=}64$), so the parameter count is only about $1\times$ that of a single GCN. The results are outlined in Tab.~\ref{tab:tu_dataset_GCN_vs_sigma_1x}. 


\begin{table}[h]
\centering
\caption{TU graph classification test accuracy (\%).
Mean~$\pm$~std over 5 seeds.
SiGMA uses 2 attention heads and 4 MP heads ($d_h=16$). The reported SiGMA model is  the better of two sets of runs with learning rate $\{10^{-3},10^{-2}\}$ (selected per family). We do not perform any hyperparameter search for MMA and SiGMA-lite: we just report the values after removing gating or placing it after the concatenation.}
\label{tab:tu_dataset_GCN_vs_sigma}
\resizebox{\textwidth}{!}{%
\begin{tabular}{lcccccccc}
\toprule
Dataset & GCN & GIN & GraphSAGE & GAT & SiGMA (homo) & SiGMA (hetero) & MMA & SiGMA-lite\\
\midrule
MUTAG     & $75.74{\pm}8.85$ & $84.26{\pm}6.99$ & $74.89{\pm}9.20$ & $72.34{\pm}3.98$ & $73.19{\pm}4.15$ & $84.68{\pm}3.50$ & $\mathbf{85.53\pm4.09}$ & $81.28\pm7.58$ \\
ENZYMES   & $41.87{\pm}5.24$ & $41.87{\pm}5.55$ & $41.07{\pm}7.24$ & $41.33{\pm}3.59$ & $47.07{\pm}1.98$ & $47.60{\pm}6.99$ & $47.87\pm7.26$ & $\mathbf{48.53\pm5.66}$  \\
PROTEINS  & $72.97{\pm}3.39$ & $72.90{\pm}1.69$ & $70.75{\pm}1.69$ & $72.19{\pm}2.04$ & $73.41{\pm}1.99$ & $\mathbf{74.12{\pm}1.06}$ & $73.84\pm1.95$ & $73.69\pm2.89$ \\
COLLAB    & $76.14{\pm}0.92$ & $\mathbf{78.40{\pm}0.90}$ & $52.35{\pm}1.18$ & $45.06{\pm}10.83$ & $73.97{\pm}0.88$ & $77.25{\pm}0.95$ & $76.88\pm1.30$ & $77.63\pm 1.39$\\
IMDB-BINARY & $65.44{\pm}1.85$ & $69.44{\pm}2.97$ & $53.04{\pm}3.97$ & $49.28{\pm}0.91$ & $66.64{\pm}1.19$ & $69.92{\pm}2.75$ & $\mathbf{70.56\pm2.93}$  & $\mathbf{70.56\pm3.69}$\\
REDDIT-BINARY & $92.60{\pm}1.62$ & $90.56{\pm}1.63$ & $49.40{\pm}1.74$ & $49.24{\pm}1.73$ & $87.92{\pm}7.51$ & $92.72{\pm}1.01$ & $93.04\pm1.26$ & $\mathbf{93.40\pm0.65}$  \\
\bottomrule
\end{tabular}%
}
\vspace{0.25em}
\end{table}

\begin{table}[h]
\centering
\caption{TU graph classification test accuracy (\%).
Mean~$\pm$~std over 5 seeds.
SiGMA uses 2 attention heads and 4 MP heads with head width matched to $\sim$1$\times$ GCN parameters ($d_h{=}4$).
The reported SiGMA result is the better of two sets of runs with learning rate $\{10^{-3},10^{-2}\}$ (selected per family). The GCN, GIN, GraphSAGE, and GAT values are taken from  Tab.~\ref{tab:tu_dataset_GCN_vs_sigma}. We do not perform any hyperparameter search for MMA and SiGMA-lite: we just report the values after removing gating or placing it after the concatenation.}
\label{tab:tu_dataset_GCN_vs_sigma_1x}
\resizebox{\textwidth}{!}{%
\begin{tabular}{lcccccccc}
\toprule
Dataset & GCN & GIN & GraphSAGE & GAT & SiGMA (homo) & SiGMA (hetero) & MMA & SiGMA-lite  \\
\midrule
MUTAG     & $75.74{\pm}8.85$ & $\mathbf{84.26{\pm}6.99}$ & $74.89{\pm}9.20$ & $72.34{\pm}3.98$ & $73.19{\pm}3.23$ & $80.43{\pm}5.90$ & $83.40\pm 6.28$ & $80.43\pm 9.44$ \\
ENZYMES   & $41.87{\pm}5.24$ & $41.87{\pm}5.55$ & $41.07{\pm}7.24$ & $41.33{\pm}3.59$ & $46.67{\pm}6.00$ & $47.60{\pm}4.46$ & $47.20\pm3.25$ & $\mathbf{48.27\pm5.71}$\\
PROTEINS  & $72.97{\pm}3.39$ & $72.90{\pm}1.69$ & $70.75{\pm}1.69$ & $72.19{\pm}2.04$ & $73.05{\pm}1.87$ & $74.34{\pm}1.20$ &  $\mathbf{74.98\pm0.93}$ & $73.12\pm 2.91$ \\
COLLAB    & $76.14{\pm}0.92$ & $\mathbf{78.40{\pm}0.90}$ & $52.35{\pm}1.18$ & $45.06{\pm}10.83$ & $73.04{\pm}1.94$ & $76.74{\pm}1.32$ & $77.25\pm1.66$ & $77.23\pm 0.90$ \\
IMDB-BINARY & $65.44{\pm}1.85$ & $69.44{\pm}2.97$ & $53.04{\pm}3.97$ & $49.28{\pm}0.91$ & $65.36{\pm}1.78$ & $71.04{\pm}1.97$ & $\mathbf{71.44\pm1.08}$ & $71.12\pm2.70$\\
REDDIT-BINARY & $92.60{\pm}1.62$ & $90.56{\pm}1.63$ & $49.40{\pm}1.74$ & $49.24{\pm}1.73$ & $89.96{\pm}5.08$ & $90.36{\pm}5.42$ & $92.40\pm2.08$ & $\mathbf{92.76{\pm}0.88}$ \\
\bottomrule
\end{tabular}%
}
\end{table}

\begin{figure}[h]
    \centering
    \includegraphics[width=1.0\textwidth]{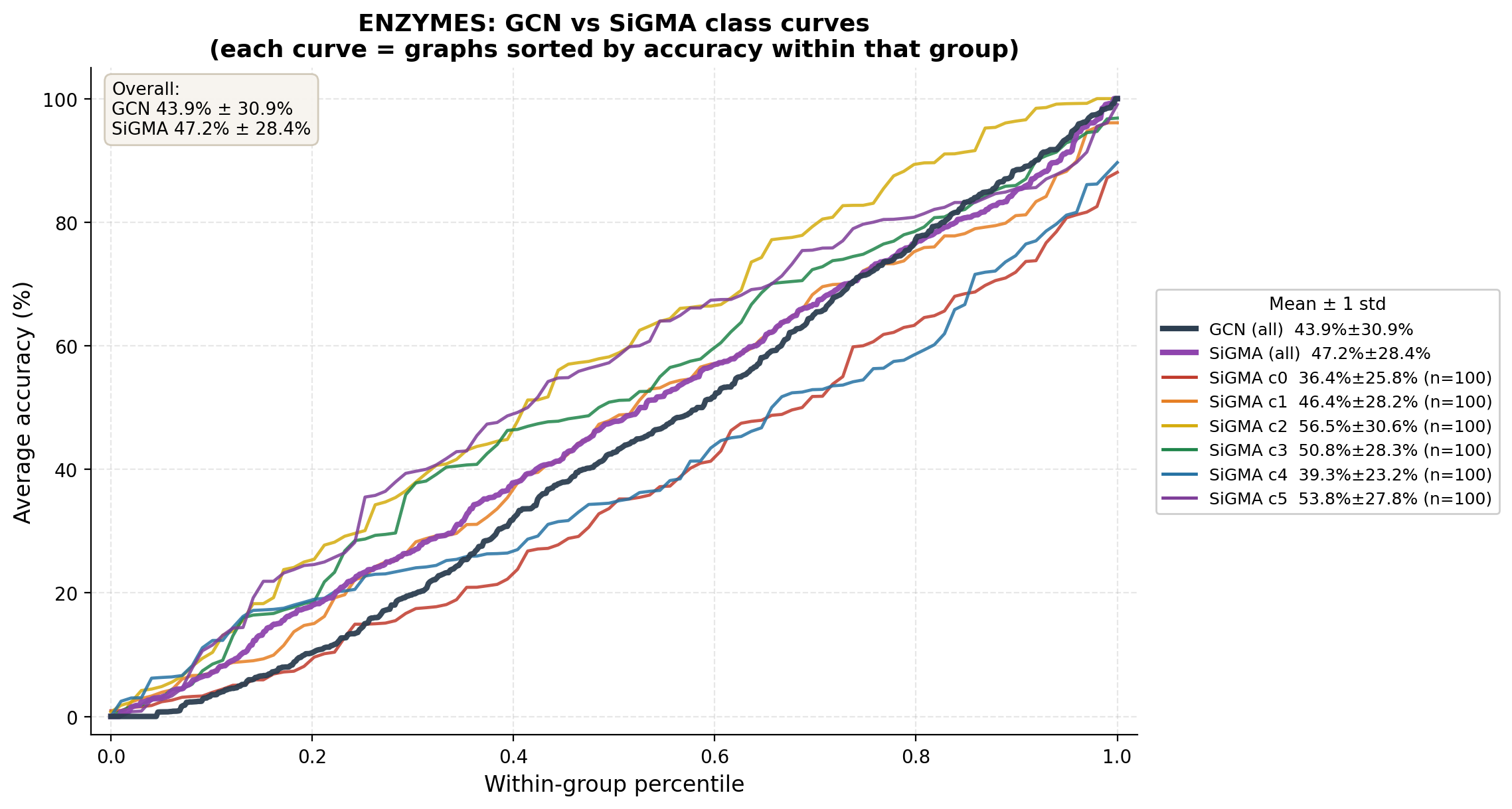}
    \caption{SiGMA and GCN performance on ENZYMES. We note that class 0 is harder to classify, even for SiGMA. Here SiGMA consists of 2 attention heads and 2 MPGNN heads (GCN,GIN), with 4 stacked blocks.}
    \label{fig:gcn-sigma-enzymes-by-class}
\end{figure}

Additionally, we present the heterogeneity profile of a smaller SiGMA architecture, with only two MPGNN heads, in Fig.~\ref{fig:gcn-sigma-enzymes-by-class}.

\subsection{TU Datasets under the protocol of Errica et al.}


Additionally, we follow the experimental protocol of Errica et al.~\cite{errica2019fair}: we use 10-fold stratified cross-validation splits, an inner 90/10
train/validation split per fold, early stopping on validation accuracy,
and per-fold hyperparameter selection before reporting test accuracy
over three random seeds.
For classical message-passing baselines, we use the published search
spaces whenever available: the full GIN grid (64 configurations) and the full
GraphSAGE grid (72 configurations) \cite{errica2019fair}.
When the public repository does not have configuration files, such as for GCN or GAT,
we construct grids in the same style
(32 configurations each), covering batch size, learning rate, width, pooling,
dropout, and the early-stopping criterion.




After per-fold selection, all models (including SiGMA) are re-evaluated with
the selected hyperparameters over three seeds; we report mean~$\pm$~std of
test accuracy over the ten folds (seed-averaged within each fold) in Tab.~\ref{tab:tu_errica_grid_eval}.

\begin{table}[t]
\centering
\caption{TU graph classification test accuracy (\%) under the protocol from Errica et al.~\cite{errica2019fair}: 10-fold cross-validation, per-fold hyperparameter selection,
3 seeds per fold; mean$\pm$std over folds.}
\label{tab:tu_errica_grid_eval}
\resizebox{\textwidth}{!}{%
\begin{tabular}{lccccc}
\toprule
Dataset & GCN & GIN & GraphSAGE & GAT & SiGMA (hetero) \\
\midrule
ENZYMES & $50.39{\pm}5.21$ & $45.39{\pm}5.24$ & $51.00{\pm}4.70$ & $42.11{\pm}6.99$ & $\mathbf{52.44{\pm}3.60}$ \\
PROTEINS & $\mathbf{73.88{\pm}3.97}$ & $73.43{\pm}4.38$ & $72.98{\pm}3.19$ & $72.69{\pm}3.09$ & $73.68{\pm}3.08$ \\
NCI1 & $80.74{\pm}1.53$ & $80.38{\pm}1.46$ & $\mathbf{81.58{\pm}2.32}$ & $75.41{\pm}2.38$ & $80.68{\pm}1.57$ \\
DD & $71.92{\pm}4.21$ & $73.73{\pm}5.22$ & $72.83{\pm}3.01$ & $72.85{\pm}9.15$ & $\mathbf{74.70{\pm}3.42}$ \\
IMDB-BINARY & $65.73{\pm}3.48$ & $71.13{\pm}4.48$ & $50.50{\pm}1.09$ & $50.40{\pm}2.00$ & $\mathbf{72.27{\pm}4.27}$ \\
REDDIT-BINARY & $\mathbf{92.62{\pm}1.02}$ & $92.45{\pm}1.10$ & $73.38{\pm}4.05$ & $74.65{\pm}2.34$ & $91.43\pm 1.43$ \\
COLLAB & $76.98{\pm}2.09$ & $76.49{\pm}2.54$ & $52.47{\pm}3.17$ & $47.55{\pm}7.87$ & $\mathbf{78.17{\pm}1.23}$ \\
\bottomrule
\end{tabular}%
}
\end{table}

\clearpage



\section{Gate Activity Analysis}
\label{app:gate-analysis} 

\begin{figure}[h!]
    \centering
    \includegraphics[width=0.6\textwidth]{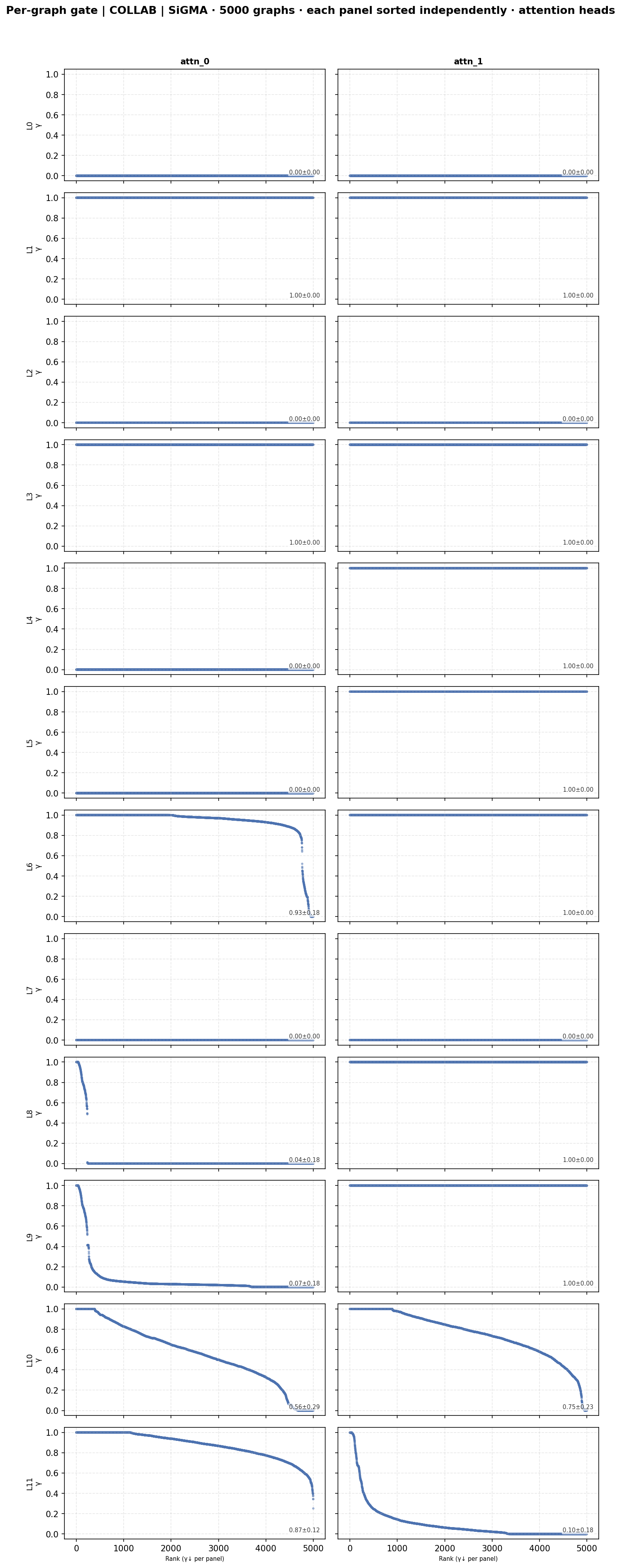}
    \caption{SiGMA gate values by head by layer (attention) at the end of training on COLLAB.}
    \label{fig:collab_SiGMA_gate_values_att}
\end{figure}

\begin{figure}[h!]
    \centering
    \includegraphics[width=0.7\textwidth]{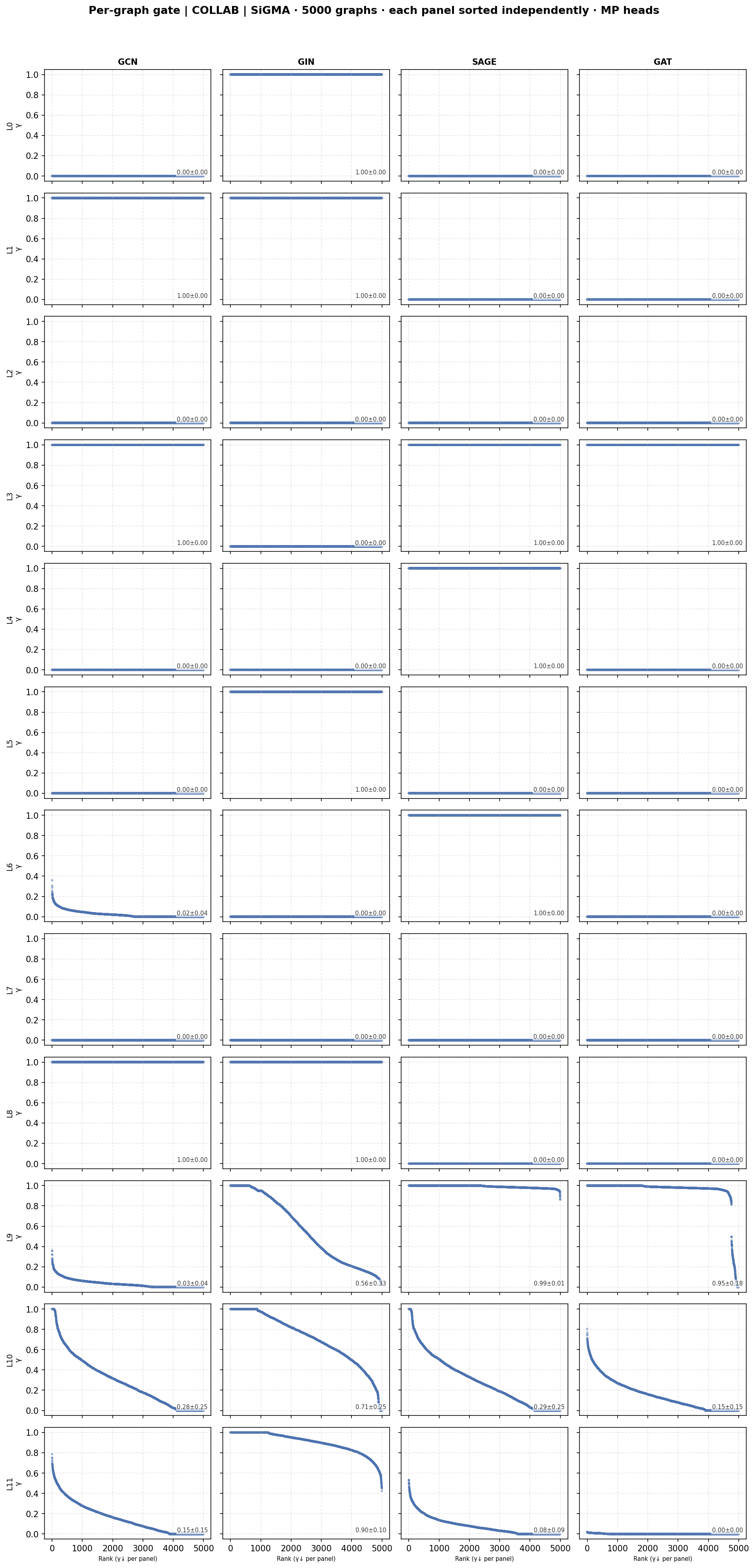}
    \caption{SiGMA gate values by head by layer (MPGNN) at the end of training on COLLAB.}
    \label{fig:collab_SiGMA_gate_values}
\end{figure}

\begin{figure}[h!]
    \centering
    \includegraphics[width=0.6\textwidth]{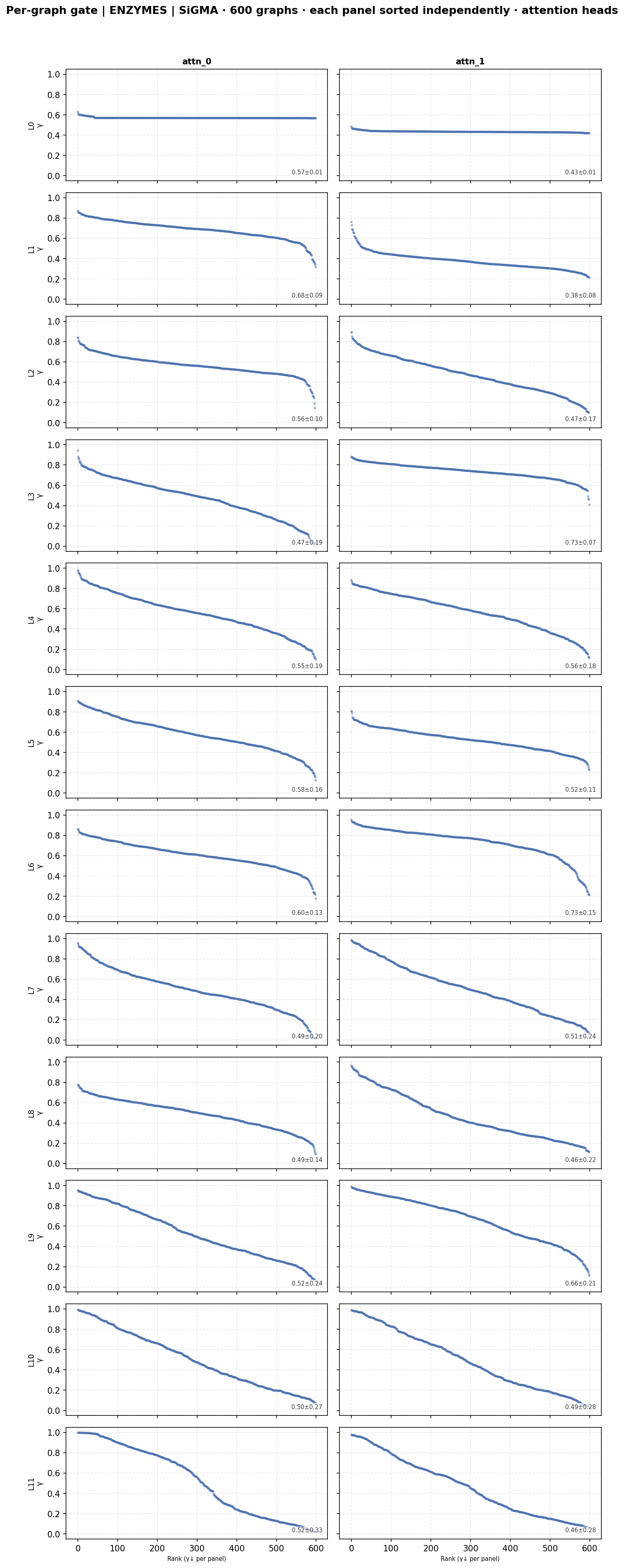}
    \caption{SiGMA gate values by head by layer (attention) at the end of training on ENZYMES.}
    \label{fig:enzymes_SiGMA_gate_values_att}
\end{figure}

\begin{figure}[h!]
    \centering
    \includegraphics[width=0.7\textwidth]{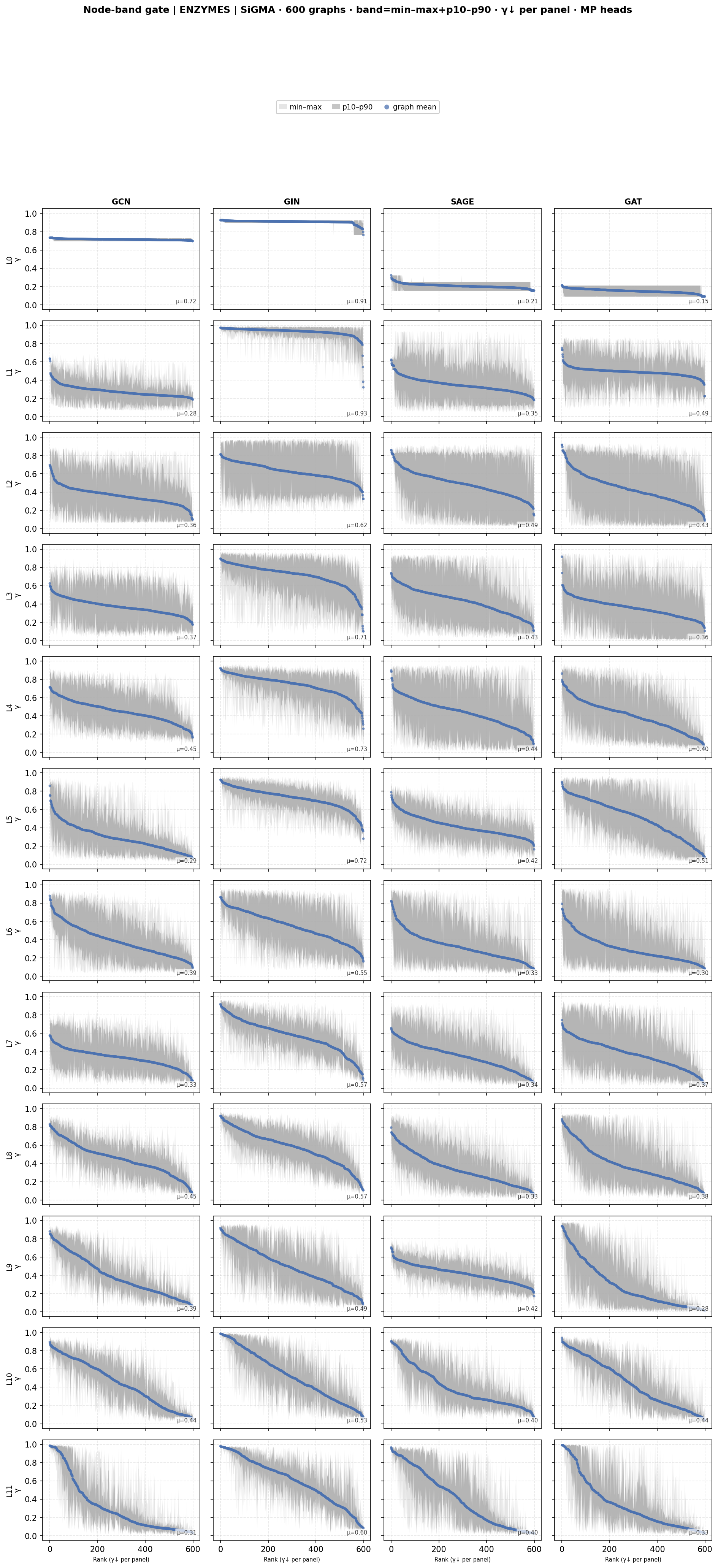}
    \caption{SiGMA gate values by head by layer (MPGNN) at the end of training on ENZYMES. Here, in addition to the graph-averaged gate value, we also show the underlying node value.}
\label{fig:enzymes_SiGMA_gate_values_nodeband}
\end{figure}

\begin{figure}[h!]
    \centering
    \includegraphics[width=0.6\textwidth]{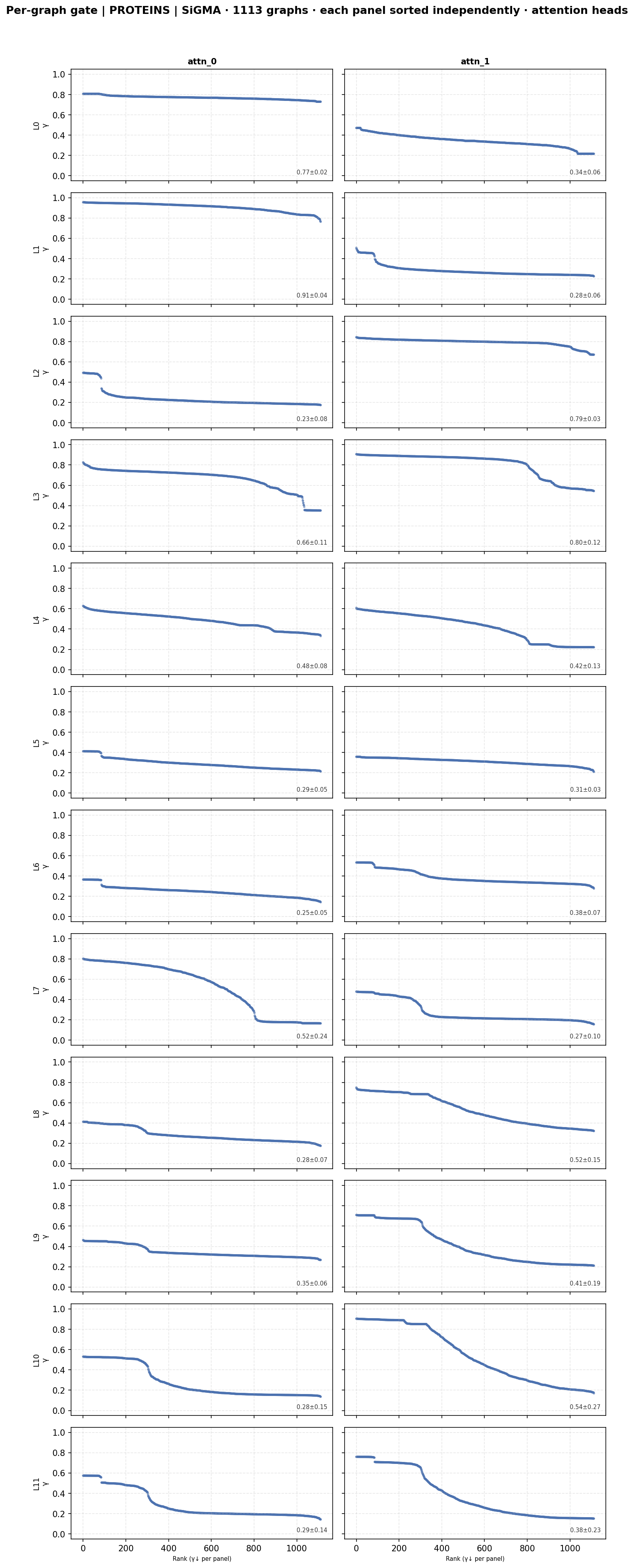}
    \caption{SiGMA gate values by head by layer (attention) at the end of training on PROTEINS.}
    \label{fig:proteins_SiGMA_gate_values_att}
\end{figure}

\begin{figure}[h!]
    \centering
    \includegraphics[width=0.7\textwidth]{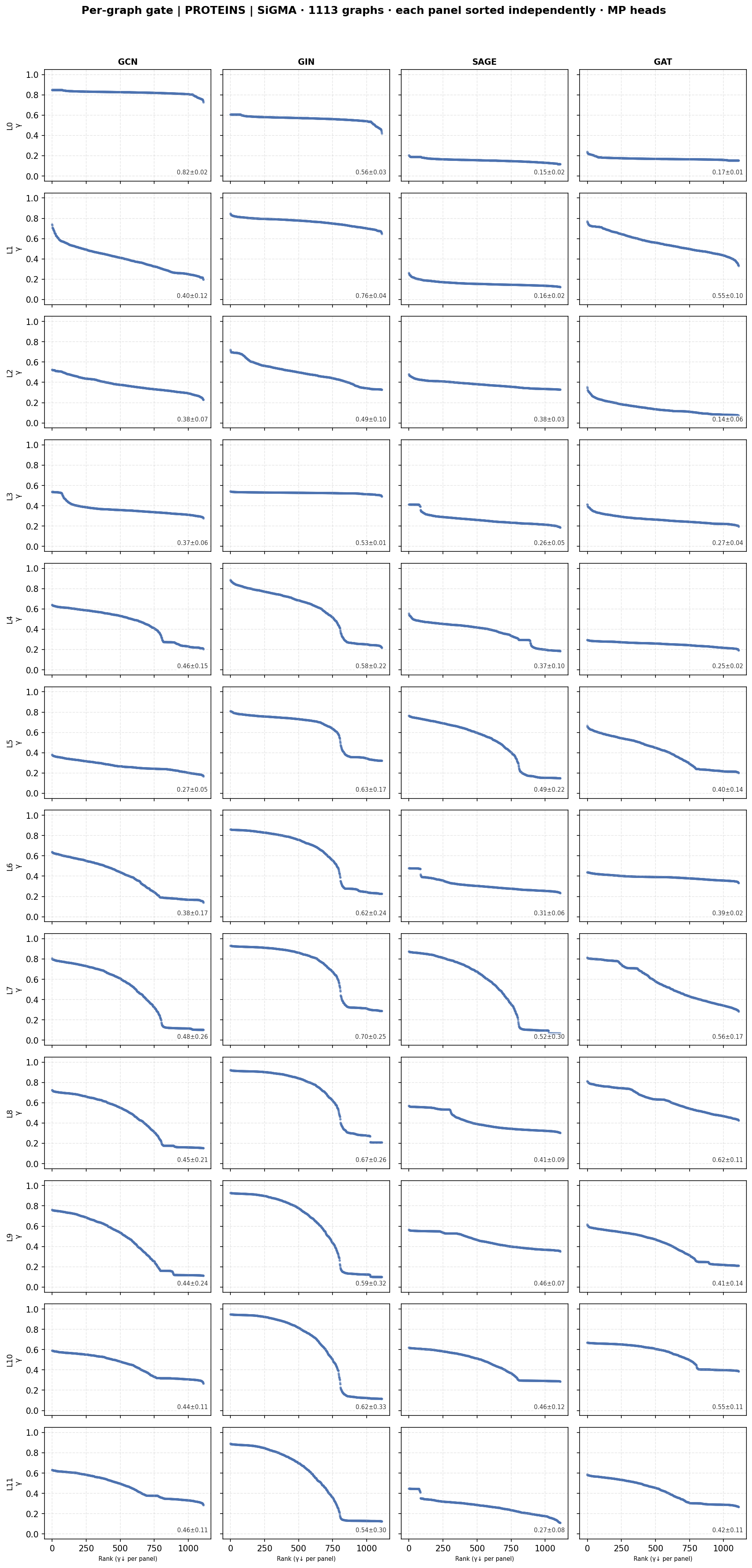}
    \caption{SiGMA gate values by head by layer (MPGNN) at the end of training on PROTEINS.}
    \label{fig:proteins_SiGMA_gate_values}
\end{figure}

\begin{figure}[h!]
    \centering
    \includegraphics[width=0.6\textwidth]{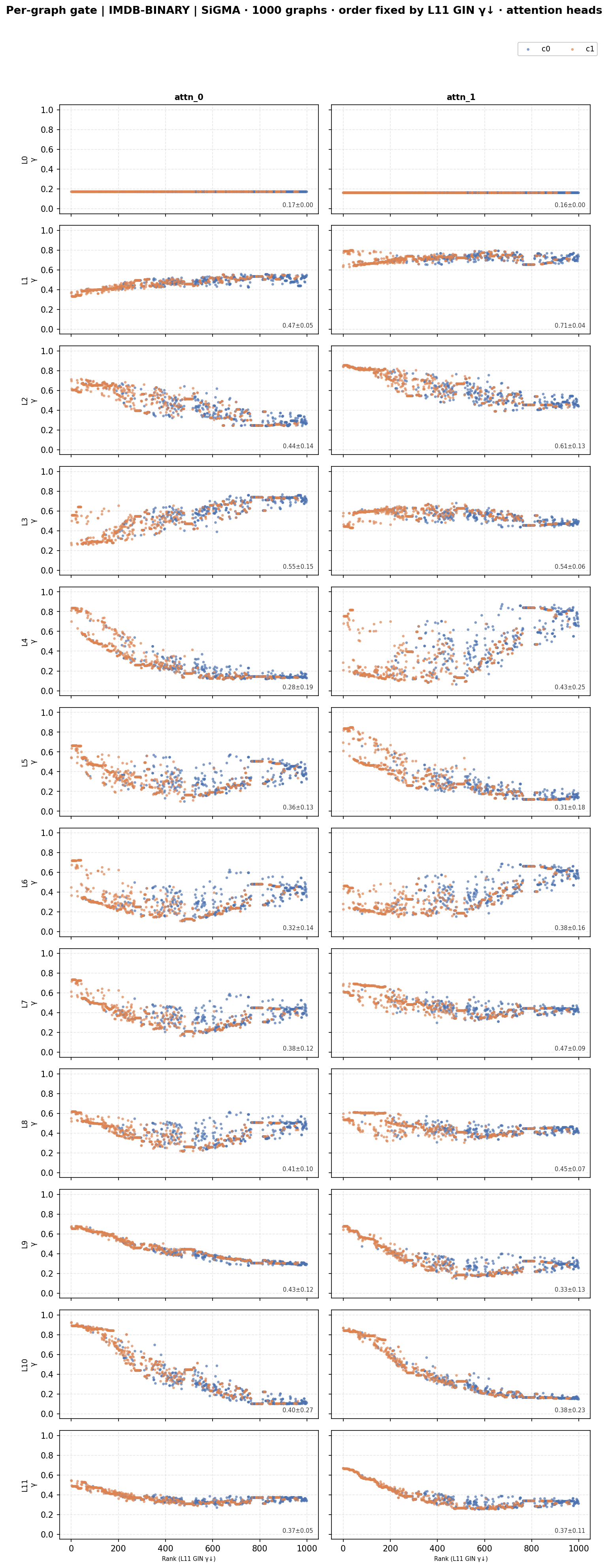}
    \caption{SiGMA gate values by head by layer (attention) at the end of training on IMDB.}
    \label{fig:imdb_SiGMA_gate_values_att}
\end{figure}

\begin{figure}[h!]
    \centering
    \includegraphics[width=0.7\textwidth]{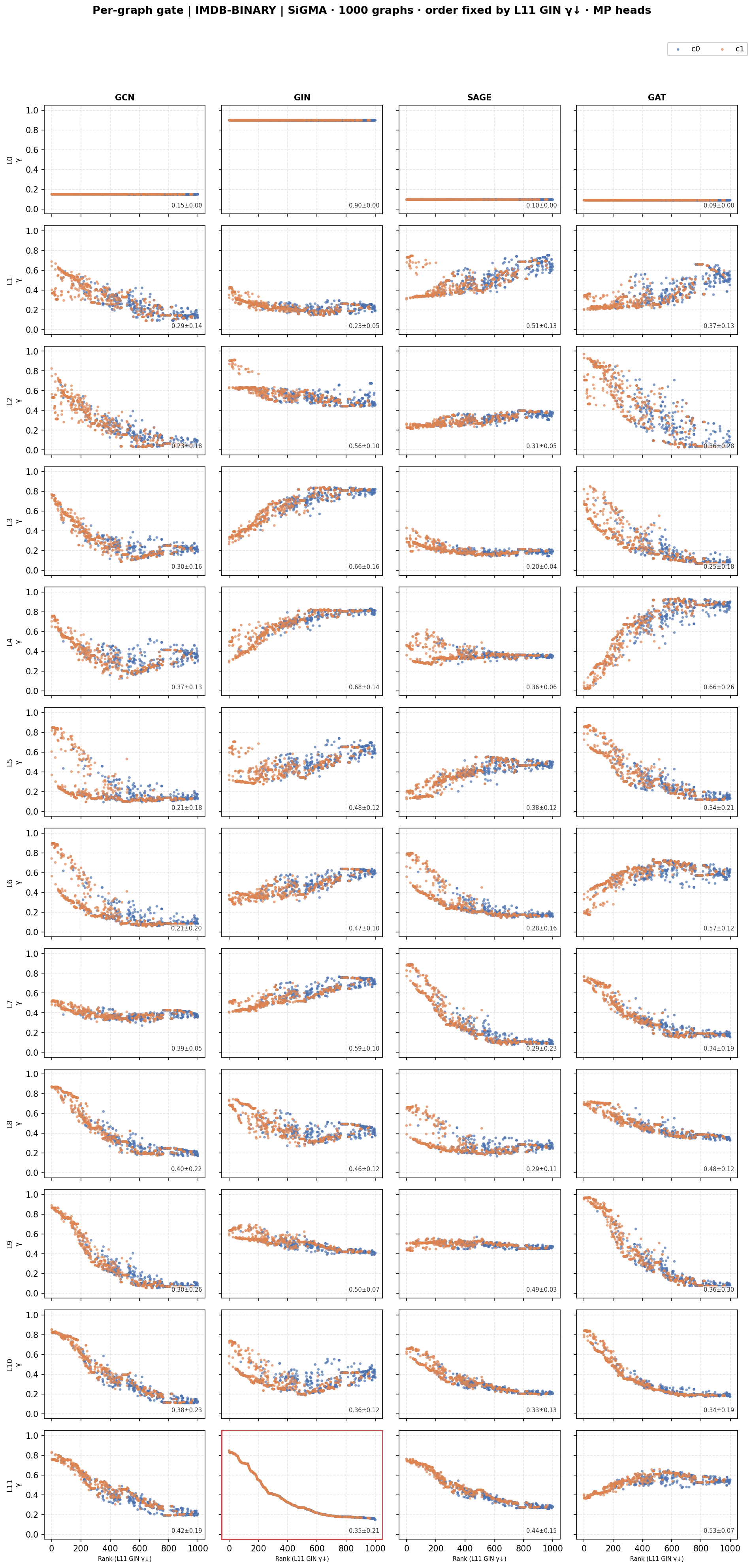}
    \caption{SiGMA gate values by head by layer (MPGNN) at the end of training on IMDB.}
    \label{fig:imdb_SiGMA_gate_values}
\end{figure}

\begin{figure}[h!]
    \centering
    \includegraphics[width=0.6\textwidth]{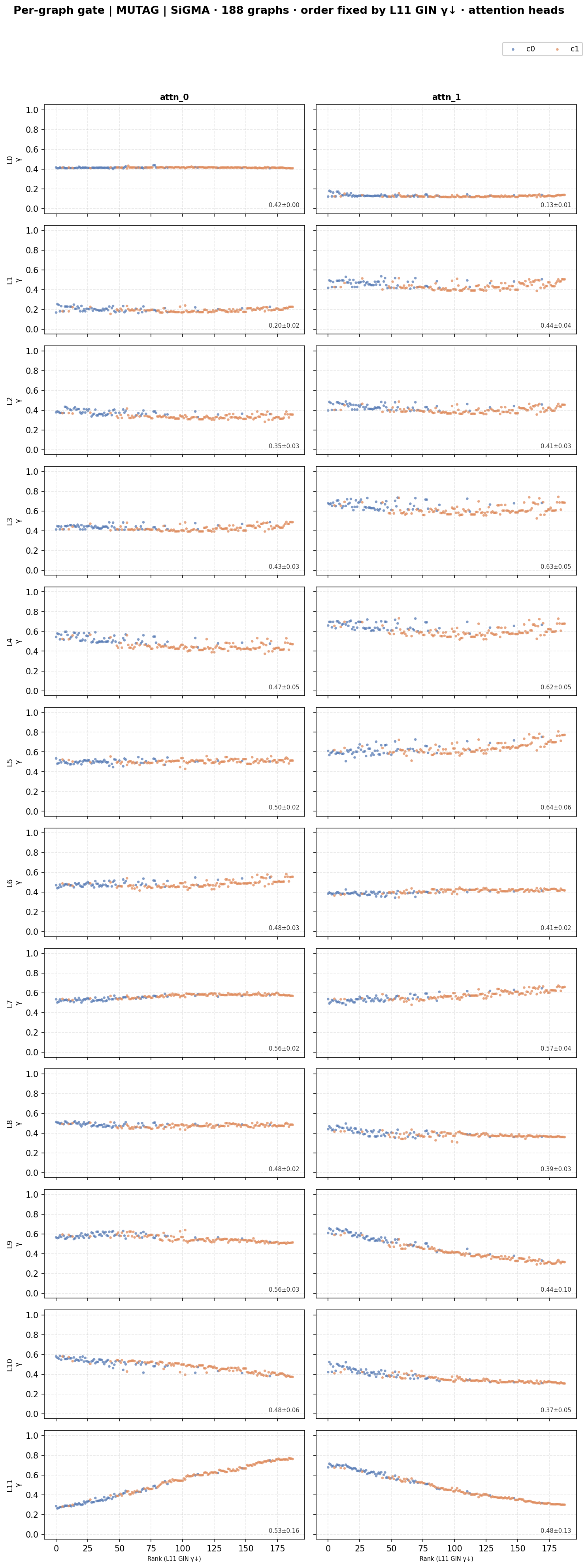}
    \caption{SiGMA gate values by head by layer (attention) at the end of training on MUTAG.}
    \label{fig:mutag_SiGMA_gate_values_att}
\end{figure}

\begin{figure}[h!]
    \centering
    \includegraphics[width=0.7\textwidth]{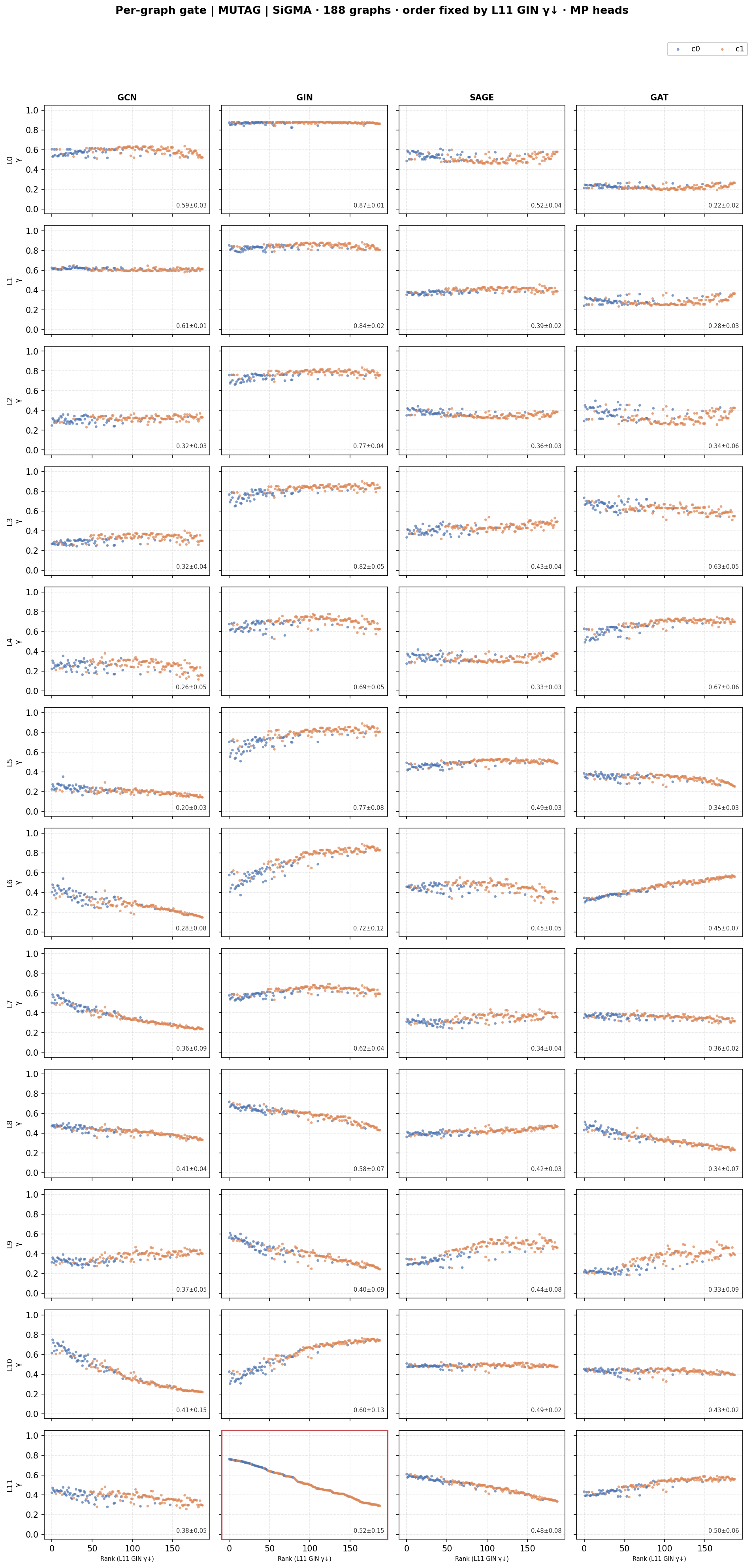}
    \caption{SiGMA gate values by head by layer (MPGNN) at the end of training on MUTAG.}
    \label{fig:mutag_SiGMA_gate_values}
\end{figure}

\begin{figure}[h!]
    \centering
    \includegraphics[width=1\textwidth]{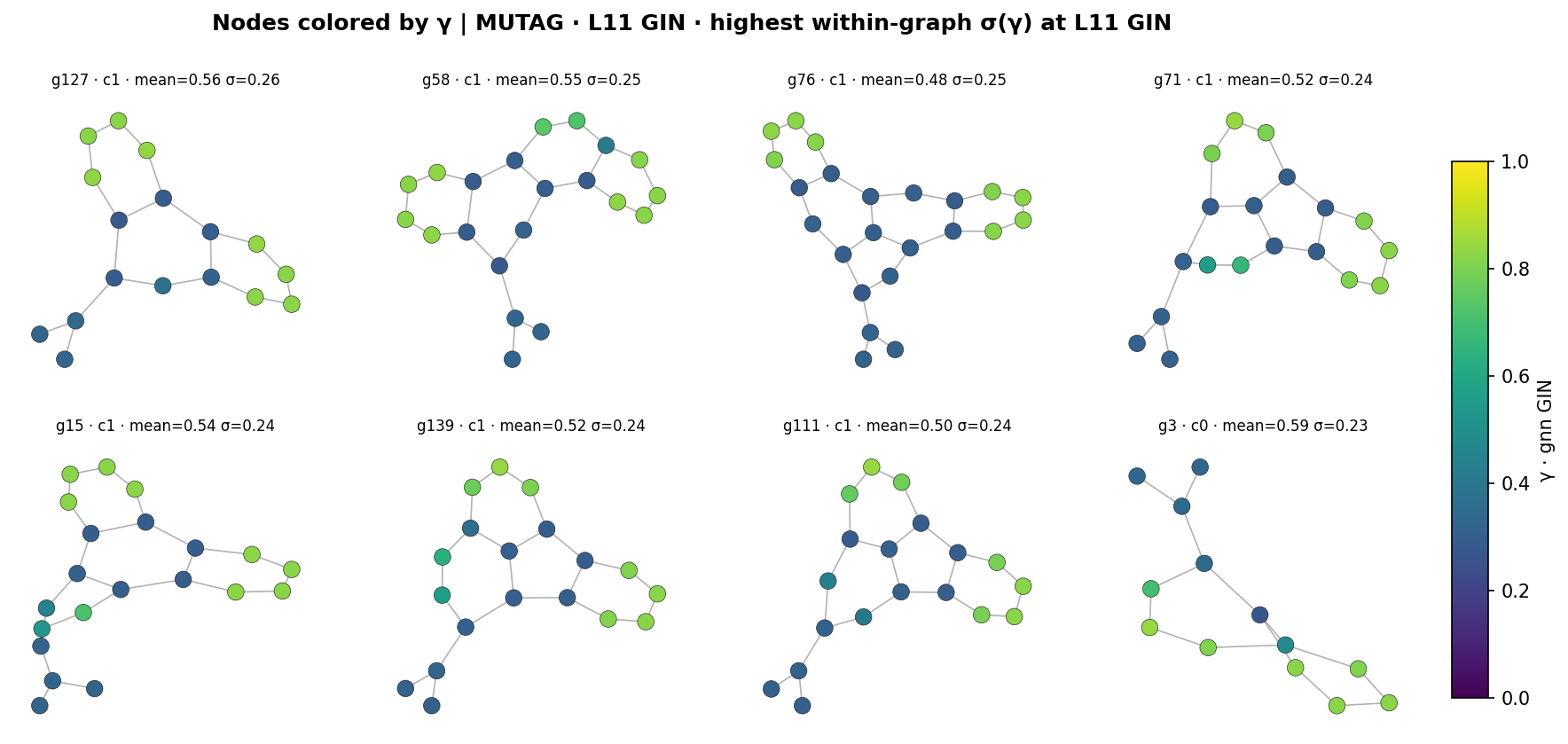}
    \includegraphics[width=1\textwidth]{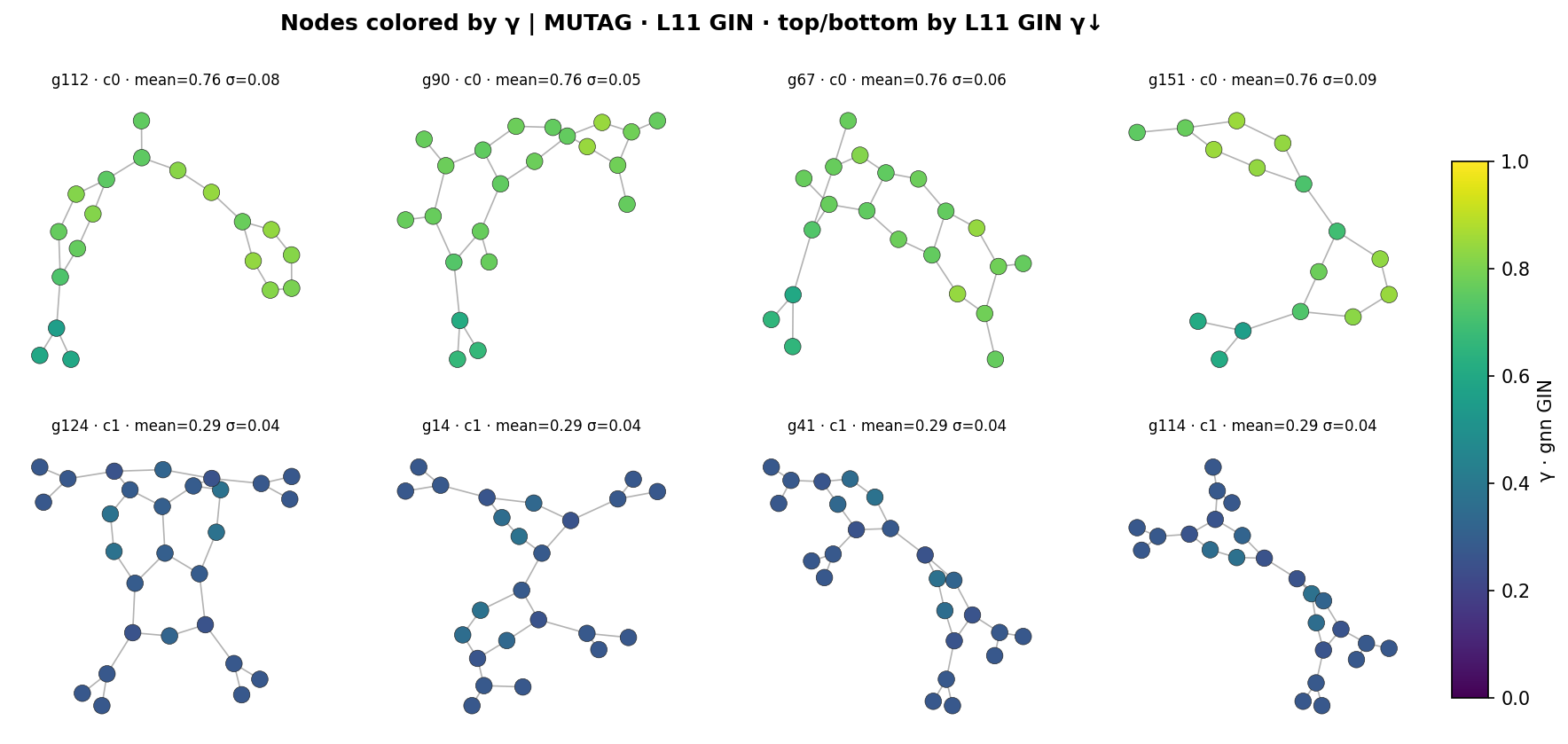}
    \caption{SiGMA gate values per node for the GIN last layer at the end of training on some graphs in MUTAG. Top: graphs with the highest per-graph gate standard deviation. Bottom: graph with the highest and lowest graph-average gate values.}
    \label{fig:mutag_SiGMA_gate_values_per_graph}
\end{figure}

We report the graph-average gate value (for the best validation checkpoint) for each graph (for each head and each layer) at the end of training a SiGMA model with 2 attention heads and 4 MPGNN heads (GCN, GIN, GraphSAGE, GAT) in  Fig.~\ref{fig:collab_SiGMA_gate_values_att} and Fig.~\ref{fig:collab_SiGMA_gate_values} for COLLAB,
Fig.~\ref{fig:enzymes_SiGMA_gate_values_att} and Fig.~\ref{fig:enzymes_SiGMA_gate_values_nodeband} for ENZYMES, 
Fig.~\ref{fig:proteins_SiGMA_gate_values_att} and  Fig.~\ref{fig:proteins_SiGMA_gate_values} for PROTEINS, Fig.~\ref{fig:imdb_SiGMA_gate_values_att} and  Fig.~\ref{fig:imdb_SiGMA_gate_values} for IMDB and Fig.~\ref{fig:mutag_SiGMA_gate_values_att} and  Fig.~\ref{fig:mutag_SiGMA_gate_values} for MUTAG. In Fig.~\ref{fig:collab_SiGMA_gate_values}, Fig.~\ref{fig:collab_SiGMA_gate_values_att},
Fig.~\ref{fig:enzymes_SiGMA_gate_values_att},  
Fig.~\ref{fig:enzymes_SiGMA_gate_values_nodeband}, 
Fig.~\ref{fig:proteins_SiGMA_gate_values_att}, and  Fig.~\ref{fig:proteins_SiGMA_gate_values} each panel is sorted independently by gate value. In Fig.~\ref{fig:imdb_SiGMA_gate_values_att},  Fig.~\ref{fig:imdb_SiGMA_gate_values}, Fig.~\ref{fig:mutag_SiGMA_gate_values_att}, and Fig.~\ref{fig:mutag_SiGMA_gate_values}, we sort the graphs once (according to their last layer's GIN's gate value) and then plot all panels with the same graph order. We also color points by their class. 
The SiGMA models used are (one seed from the 5 runs of the models) from Tab.~\ref{tab:tu_dataset_GCN_vs_sigma}. Interestingly, the gates of some heads are sometimes fully closed or fully opened at some layers for all graphs in COLLAB. At layer 0, only the GIN MPGNN head is active, while the GCN, GraphSAGE and GAT  heads are shut off. At layer 0, both attention heads are also shut off. Only at layer 6 do graphs start to get different gate values (on GCN and one of the attention heads). Overall, the standard deviation of the graph-average gate values within each head tends to increase as we go to later layers. In Fig.~\ref{fig:mutag_SiGMA_gate_values_per_graph}, we plot some graphs in MUTAG with their gate values at each node for the GIN MPGNN head at the last layer.


\clearpage



\newpage
\section{Synthetic GCN versus GIN operator-routing benchmark}
\label{app:routing analysis} 

We create a controlled synthetic benchmark used to test whether headwise gating routes to the correct message-passing operator
per graph. We create $14,000$ graphs ($10,000$ used for training, $2,000$ used for validation, and $2,000$ used as the test set). Each graph is assigned a type $\tau\in\{0,1\}$ that specifies
which aggregation rule defines the ground-truth graph label $y$. The benchmark complements the real-data
heterogeneity analyzes by isolating operator routing from confounding dataset factors.

\begin{figure}[h]
    \centering
    \includegraphics[width=0.5\textwidth]{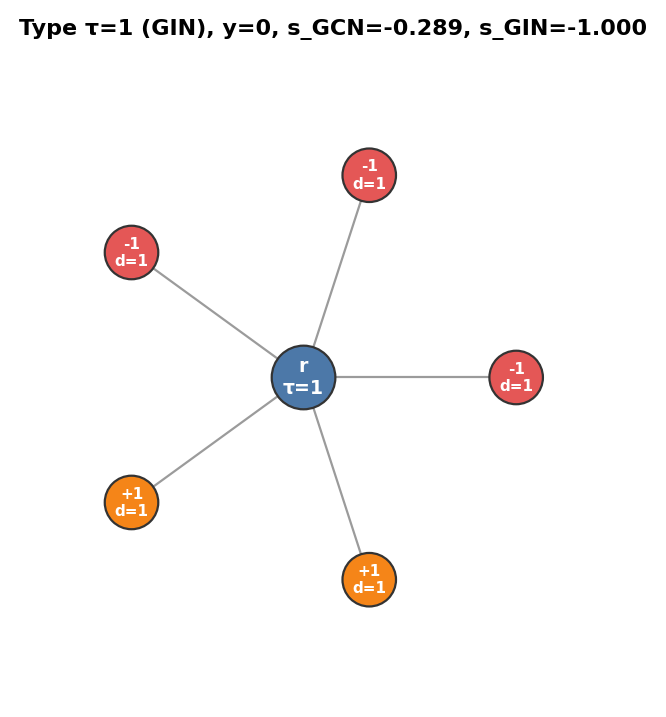}
    \caption{Example rooted star. The root is in blue, while the signal neighbors are in red and orange and are colored by their signal value. Here all signal neighbors have degree $d=1$. The type is $
    \tau=1$ so we first compute: $s_{\text{GIN}} = \sum_{u\in\mathcal{N}(r)} x_u = -1$ and then find that $y = \mathbbm{1}\!\left[ s_{\text{GIN}}  > 0 \right] =0$ using  Eq.~\ref{eq:app-gin-rule}. Best seen in color.}
    \label{fig:rooted-star}
\end{figure}

\subsection{Dataset construction}
\label{app:gcn_gin_routing:data}

\begin{figure}[h]
    \centering
    \includegraphics[width=0.6\textwidth]{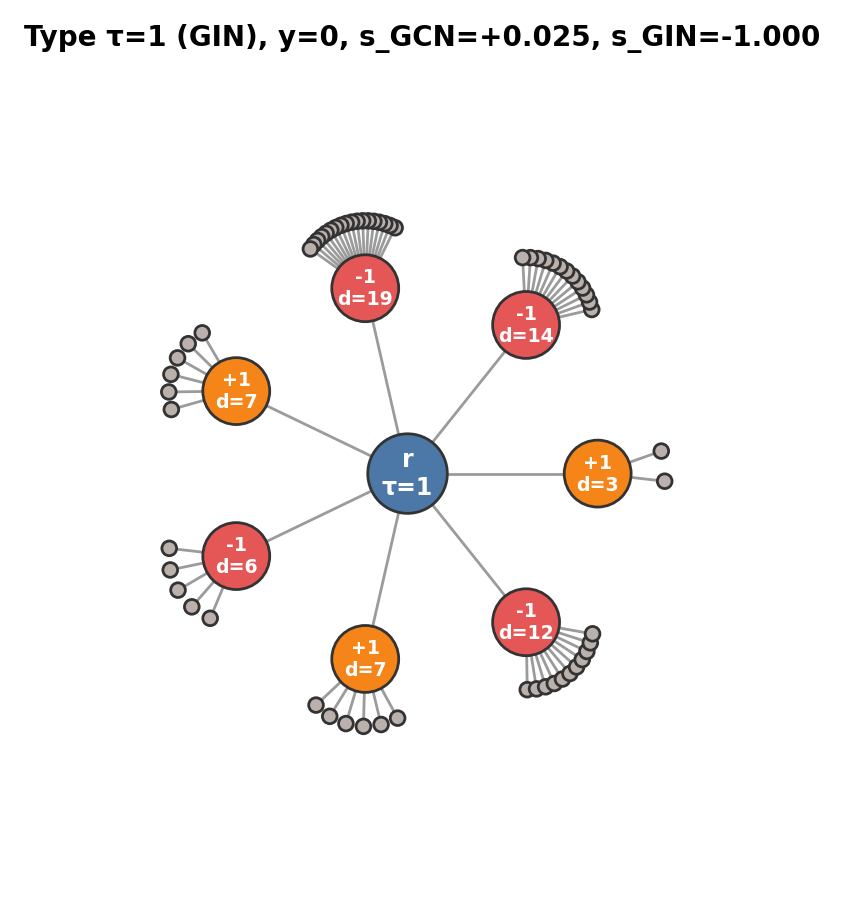}
    \caption{Example rooted star with different score signs $\mathrm{sign}(s_{\mathrm{GCN}}){\neq}\mathrm{sign}(s_{\mathrm{GIN}})$. The root is in blue, while the signal neighbors are in red and orange and are colored by their signal value. The dummy leaves attached to the signal nodes are in grey. Here, the type is $
    \tau=1$, $s_{\text{GIN}} = \sum_{u\in\mathcal{N}(r)} x_u = -1$ so Eq.~\ref{eq:app-gin-rule} gives the label $y=0$. Best seen in color.}
    \label{fig:rooted-star-diff-sign}
\end{figure}

Every instance is a rooted star. A central root node $r$ connects to
$k\in\{4,\ldots,8\}$ signal neighbors $u_1,\ldots,u_k$. Fig.~\ref{fig:rooted-star} and Fig.~\ref{fig:rooted-star-diff-sign} show representative stars. Each signal neighbor $u$ carries a binary feature $x_u\in\{-1,+1\}$ and may
attach up to $24$ dummy leaf nodes (that carry zero-vectors features) so that neighbors'
degrees' $d_u$ vary. Each node has a 2-dimensional feature vector $[\text{signal},\,\text{type}]^T$. The root $r$ has signal $=0$, and type channel $=\tau$. Signal neighbors have signal $=x_u$, and type $=0$. Dummy leaves have both signal and type channel equal to $0$. The type $\tau$ is visible at the root (for gating and classification) but
is not part of the neighbor aggregation sums used to define labels. Each graph is assigned a  type $\tau\in\{0,1\}$ and the binary labels are constructed as follows:

\begin{align}
  \text{($\tau=0$):}\quad
  y &= \mathbbm{1}\!\left[
    \sum_{u\in\mathcal{N}(r)}
    \frac{x_u}{\sqrt{(d_r+1)(d_u+1)}} > 0
  \right], \label{eq:app-gcn-rule}\\
  \text{($\tau=1$):}\quad
  y &= \mathbbm{1}\!\left[ \sum_{u\in\mathcal{N}(r)} x_u > 0 \right].
  \label{eq:app-gin-rule}
\end{align}

where $\mathbbm{1}$ denotes the indicator function. Labels are  read out from the root node. Degree normalization therefore matters for the GCN-type rule but not for the GIN-type rule. Because the label constructions mimic the GCN and GIN aggregation rules, we refer to $\tau=0$-graphs as GCN-type and to $\tau=1$-graphs as GIN-type. We define the normalized GCN sum and the unnormalized GIN sums as follows:

\begin{align}
    s_{\text{GCN}} & := \sum_{u\in\mathcal{N}(r)}
    \frac{x_u}{\sqrt{(d_r+1)(d_u+1)}}, \label{eq:s-gcn-rule}\\
  s_{\text{GIN}} &:= \sum_{u\in\mathcal{N}(r)} x_u  \label{eq:s-gin-rule}
\end{align}

During generation, we first reserve up to $\lfloor 0.2N_{\mathrm{split}}\rfloor$ twin pairs per split (corresponding to $\lfloor 0.4N_{\mathrm{split}}\rfloor$ graphs), and aim to fill the remainder slots with singleton graphs drawn from the same random neighbor-feature and dummy-leaf sampler.

We repeatedly sample graphs. For each candidate draw we compute both the unnormalized GIN sum $s_{\mathrm{GIN}}$ and the degree-normalized GCN sum $s_{\mathrm{GCN}}$. Whenever $s_{\mathrm{GIN}}$ and  $s_{\mathrm{GCN}}$ disagree in sign (and the split budget allows),
we emit a matched twin pair: two graphs with identical topology and
neighbor features, but $\tau\in\{0,1\}$. Because $\tau$ selects which aggregation rule defines the label, the two twins
always receive opposite binary labels $y$. For example, the graph in Fig.~\ref{fig:rooted-star-diff-sign} would receive a different label, namely $y=1$, if $\tau=0$ instead of $1$. See Tab.~\ref{tab:app-routing-dataset-breakdown} for a breakdown of the graphs. In practice, about one quarter of training graphs belong to
such twins ($1{,}205$ pairs out of $10{,}000$ graphs); the
generator targets $\lfloor 0.2\,N_{\mathrm{split}}\rfloor$ twin pairs per
split, but the realized count is limited by the overall graph budget.
By construction, one member of a twin pair should be hard to distinguish for each MP head; we expect that a model that uses graph-level type information
(via gating) will route to the correct operator. If $s_{\mathrm{GIN}}$ and  $s_{\mathrm{GCN}}$ agree in sign (or the twin budget is full), we instead emit a single graph and sample $\tau\sim\mathrm{Unif}\{0,1\}$: both aggregation rules would assign the same label on that instance. Including these aligned graphs where $s_{\mathrm{GIN}}$ and $s_{\mathrm{GCN}}$ have the same-sign prevents the benchmark from being exclusively adversarial.

\begin{table}[t]
\centering
\caption{Graph counts by split and type $\tau$.}
\label{tab:app-routing-sizes}
\begin{tabular}{lrrrr}
\toprule
Split & Total & $\tau{=}0$ (GCN-type) & $\tau{=}1$ (GIN-type) \\
\midrule
Train & 10{,}000 & 4{,}975 & 5{,}025 \\
Val   &  2{,}000 &   973 & 1{,}027 \\
Test  &  2{,}000 &   986 & 1{,}014 \\
\midrule
All   & 14{,}000 & 6{,}934 & 7{,}066 \\
\bottomrule
\end{tabular}
\end{table}

We generate $14,000$ graphs and use $N_{\mathrm{train}}=10{,}000$, $N_{\mathrm{val}}=2{,}000$,
$N_{\mathrm{test}}=2{,}000$ graphs.
The test set contains $986$ $\tau{=}0$ and $1{,}014$ $\tau{=}1$ graphs (see Tab.\ref{tab:app-routing-sizes}, Tab.~\ref{tab:app-routing-dataset-breakdown}).

\begin{table*}[t]
\centering
\small
\setlength{\tabcolsep}{5pt}
\caption{Synthetic dataset characteristics.
Each row counts graphs with the given split and type $\tau$.
The last two columns classify graphs by whether the GCN and GIN
labeling rules (Eq.~\ref{eq:s-gcn-rule} and \ref{eq:s-gin-rule}) would assign the same binary class, i.e.\
$\mathbbm{1}(s_{\mathrm{GCN}}{>}0)=\mathbbm{1}(s_{\mathrm{GIN}}{>}0)$, independently of~$\tau$. The graphs in the last column are the twin pairs.
Italic subtotal rows aggregate over both $\tau$ values in that split.}
\label{tab:app-routing-dataset-breakdown}
\begin{tabular}{llrrr}
\toprule
Split & $\tau$ & $N$ &
\makecell{Rules agree\\$\mathbbm{1}(s_{\mathrm{GCN}}{>}0)=\mathbbm{1}(s_{\mathrm{GIN}}{>}0)$} &
\makecell{Rules disagree\\$\mathbbm{1}(s_{\mathrm{GCN}}{>}0)\neq \mathbbm{1}(s_{\mathrm{GIN}}{>}0)$} \\
\midrule
\multirow{2}{*}{Train}
  & 0 & 4{,}975 & 3{,}770 & 1{,}205 \\
  & 1 & 5{,}025 & 3{,}820 & 1{,}205 \\
\cmidrule(lr){1-5}
\multicolumn{2}{r}{\textit{Train total}} & 10{,}000
  & 7{,}590 (75.9\%) & 2{,}410 (24.1\%) \\
\addlinespace
\multirow{2}{*}{Val}
  & 0 & 973 & 756 & 217 \\
  & 1 & 1{,}027 & 810 & 217 \\
\cmidrule(lr){1-5}
\multicolumn{2}{r}{\textit{Val total}} & 2{,}000
  & 1{,}566 (78.3\%) & 434 (21.7\%) \\
\addlinespace
\multirow{2}{*}{Test}
  & 0 & 986 & 758 & 228 \\
  & 1 & 1{,}014 & 786 & 228 \\
\cmidrule(lr){1-5}
\multicolumn{2}{r}{\textit{Test total}} & 2{,}000
  & 1{,}544 (77.2\%) & 456 (22.8\%) \\
\midrule
\multirow{2}{*}{All}
  & 0 & 6{,}934 & 5{,}284 & 1{,}650 \\
  & 1 & 7{,}066 & 5{,}416 & 1{,}650 \\
\cmidrule(lr){1-5}
\multicolumn{2}{r}{\textit{All total}} & 14{,}000
  & 10{,}700 (76.4\%) & 3{,}300 (23.6\%) \\
\bottomrule
\end{tabular}
\end{table*}


\subsection{Experimental protocol}
\label{app:gcn_gin_routing:protocol}

To study the effect of SiGMA and gating, we train four model families on the same splits (we repeat the procedure five times for five splits in total), with two tracks that differ in convolution implementation:
\paragraph{Track~A}
In track A, we use one hybrid layer with no attention head and 2 MPGNN heads (GIN-type sum convolution and GCN-type
normalized-sum convolution), with head width $d_h{=}1$,
headwise MP gating, and root readout. The convolutions implemented match Eqs.~\eqref{eq:app-gcn-rule}--\eqref{eq:app-gin-rule} exactly, without self-loops on the root aggregation path.
\paragraph{Track~B}
In Track B, we use the same architecture template but with PyTorch Geometric's \texttt{GINConv} and \texttt{GCNConv}
heads and $d_h{=}4$, matching the SiGMA stack used in the rest of the paper.

For both tracks, we compare a GCN-only head, a GIN-only head, and SiGMA with two MPGNN heads (GCN and GIN) and no attention head. We additionally consider MMA (the two heads are simply concatenated). We perform training of the models with Adam, learning rates in $\{10^{-3},10^{-2}\}$, for up to $200$
epochs (with cosine warmup). We train each model family five times with five different seeds. This yields $80$ runs in total
($2$ tracks $\times$ $4$ models $\times$ $2$ LRs $\times$ $5$ seeds).

\subsection{Results}
\label{app:gcn_gin_routing:results}

We report four families of quantities on the held-out test split ($n{=}2{,}000$ graphs). Let $y_i\in\{0,1\}$ denote the label of graph $i$ and $\hat{y}_i$ the predicted
class.

\begin{figure}[t]
  \centering
  \includegraphics[width=\linewidth]{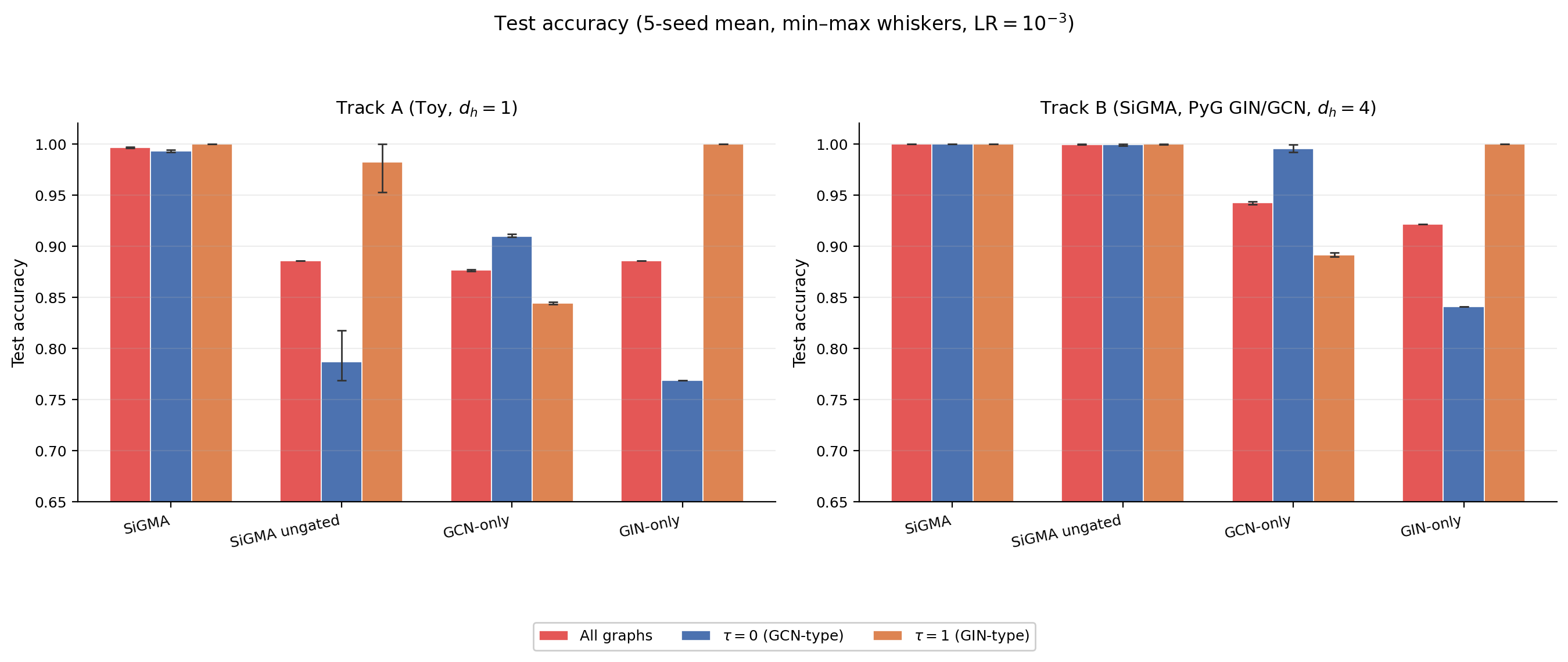}
  \caption{Per-type test accuracy of the four models on Track A and Track B. Single-operator baselines excel on their matching type and do not perform as well on the other; SiGMA succeeds on both. The advantage of gating, as opposed to simple concatenation (SiGMA vs MMA (SiGMA ungated)), is pronounced in Track A.}
  \label{fig:app-routing-baseline}
\end{figure}

\paragraph{Accuracy.} 
We report overall test accuracy and the accuracies on the $\tau{=}0$
(GCN-type) and $\tau{=}1$ (GIN-type) subsets separately. Tab.~\ref{tab:app-routing-acc} and Fig.~\ref{fig:app-routing-baseline} report $5$-seed mean test accuracy at
learning rate $10^{-3}$ (best LR for gated models). The GCN-only model performs better on $\tau=0$ graphs than on $\tau=1$ graphs in both tracks. Likewise, the GIN-only model performs better on $\tau=1$ graphs than on $\tau=0$ graphs in both tracks. SiGMA, on the other hand, is able to achieve $100\%$ test accuracy over the 5 seeds in track B, while performing close to perfectly in track A ($99.3\pm0.1$ accuracy on $\tau=0$ graphs and $100.0\pm0.0$ on $\tau=1$ graphs).

\begin{table}[t]
  \centering
  \caption{Test accuracy (\%) on the synthetic routing benchmark
  ($\mathrm{lr}=10^{-3}$, mean $\pm$ std over $5$ seeds).
  SiGMA is the only model near ceiling on \emph{both} graph types
  in both tracks.}
  \label{tab:app-routing-acc}
  \small
  \begin{tabular}{llccc}
    \toprule
    Track & Model & All graphs $\uparrow$ & $\tau{=}0$ graphs $\uparrow$ & $\tau{=}1$ graphs $\uparrow$ \\
    \midrule
    \multirow{4}{*}{A ($d_h{=}1$)}
      &  SiGMA      & $99.6 \pm 0.0$ & $99.3 \pm 0.1$ & $100.0 \pm 0.0$ \\
      &  MMA    & $88.6 \pm 0.0$ & $78.7 \pm 2.1$ & $98.2 \pm 2.1$ \\
      & GCN-only             & $87.6 \pm 0.1$ & $91.0 \pm 0.1$ & $84.4 \pm 0.1$ \\
      & GIN-only             & $88.6 \pm 0.0$ & $76.9 \pm 0.0$ & $100.0 \pm 0.0$ \\
    \midrule
    \multirow{4}{*}{B ($d_h{=}4$)}
      &  SiGMA      & $100.0 \pm 0.0$ & $100.0 \pm 0.0$ & $100.0 \pm 0.0$ \\
      &  MMA    & $99.9 \pm 0.0$ & $99.9 \pm 0.1$ & $100.0 \pm 0.0$ \\
      & GCN-only             & $94.3 \pm 0.1$ & $99.5 \pm 0.3$ & $89.2 \pm 0.2$ \\
      & GIN-only             & $92.2 \pm 0.0$ & $84.1 \pm 0.0$ & $100.0 \pm 0.0$ \\
    \bottomrule
  \end{tabular}
\end{table}

\begin{figure}[t]
  \centering
  \includegraphics[width=0.9\linewidth]{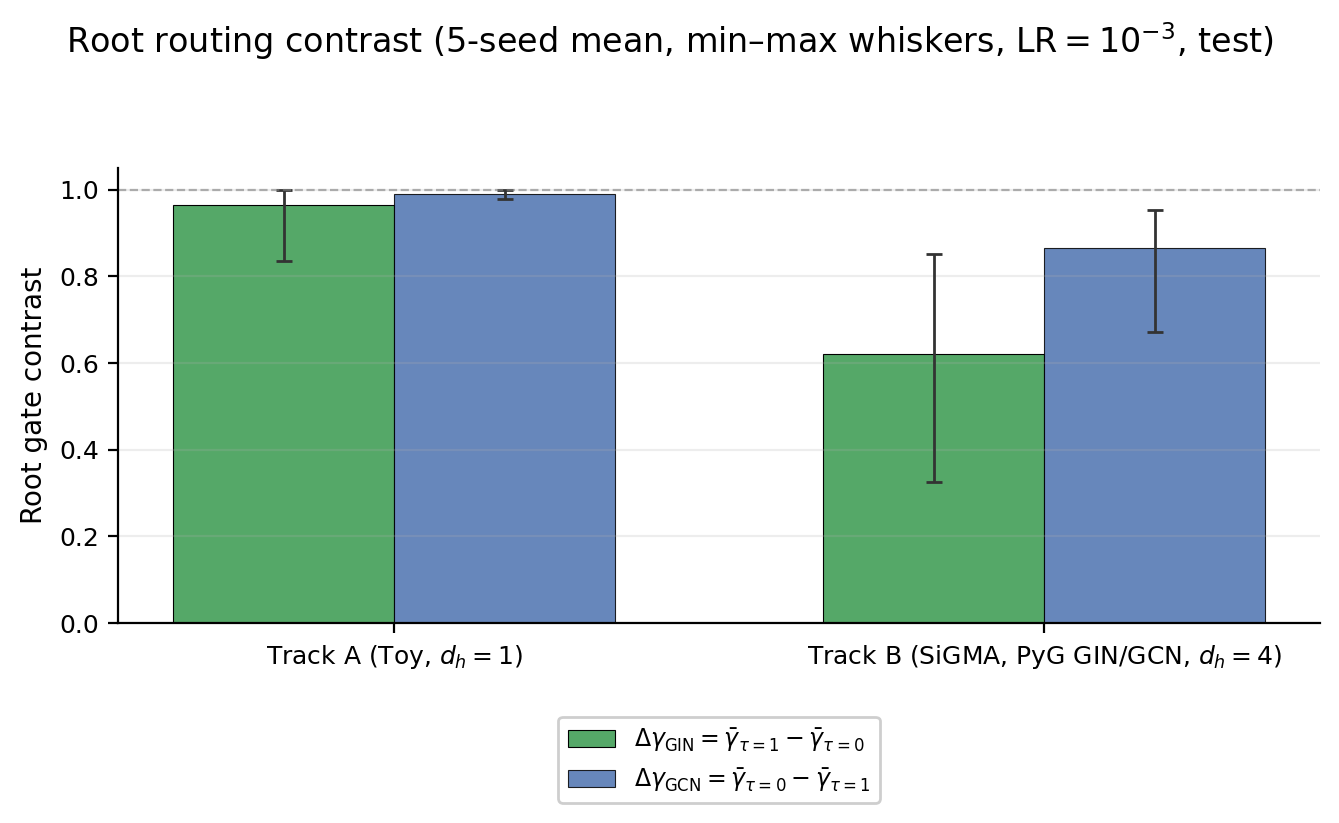}
  \caption{Per-seed gate contrasts
  $\Delta\gamma_{\mathrm{GCN}}$ and $\Delta\gamma_{\mathrm{GIN}}$ at the root on the test set, using SiGMA. Positive values indicate routing in the intended direction.}
  \label{fig:app-routing-delta}
\end{figure}

\paragraph{Routing contrasts and root gate values.} For each SiGMA checkpoint we record the headwise MP gates
$\gamma_{\mathrm{GIN}}$ and $\gamma_{\mathrm{GCN}}$ at the classification
root $r$ (for the unique layer).
We summarize them by graph type using empirical means, e.g.\
$\bar{\gamma}_{\mathrm{GCN}}(\tau{=}0)$ and
$\bar{\gamma}_{\mathrm{GCN}}(\tau{=}1)$, and report the routing contrasts:
\begin{equation}
\Delta\gamma_{\mathrm{GCN}}
  := \bar{\gamma}_{\mathrm{GCN}}(\tau{=}0) - \bar{\gamma}_{\mathrm{GCN}}(\tau{=}1)
\end{equation}
\begin{equation}
  \Delta\gamma_{\mathrm{GIN}}
  := \bar{\gamma}_{\mathrm{GIN}}(\tau{=}1) - \bar{\gamma}_{\mathrm{GIN}}(\tau{=}0) 
\end{equation}
Positive values indicate that the corresponding head receives higher gate mass
on graphs whose label is defined by that operator's aggregation rule. The routing contrast values are reported in Fig.~\ref{fig:app-routing-delta}.

\begin{figure}[t]
  \centering
  \includegraphics[width=0.85\linewidth]{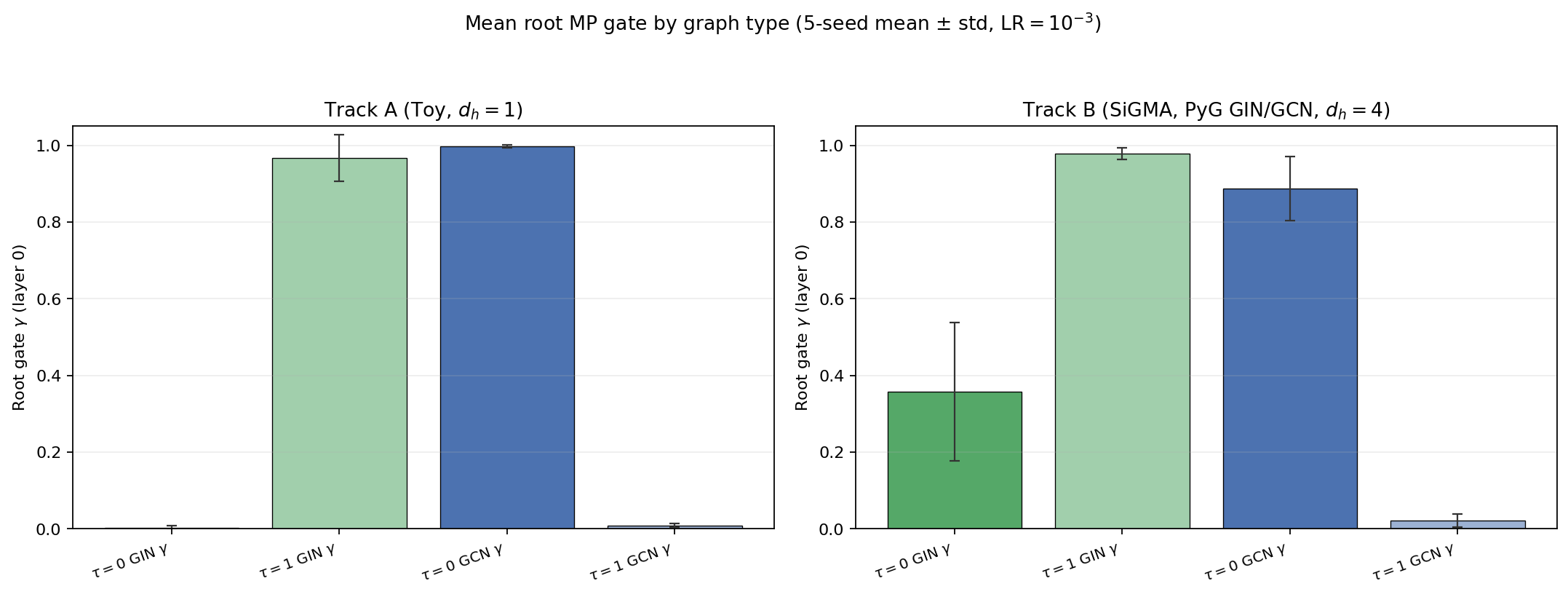}
  \caption{Root gates by graph type using the SiGMA model, on the test set. In track A, we observe a sharp type--gate correspondence. This correspondence is softer in Track B, but we still observe a clear separation between the gate values for GCN or GIN within each $\tau$ family.}
  \label{fig:app-routing-gates}
\end{figure}

\begin{figure}[t]
  \centering
  \includegraphics[width=0.8\linewidth]{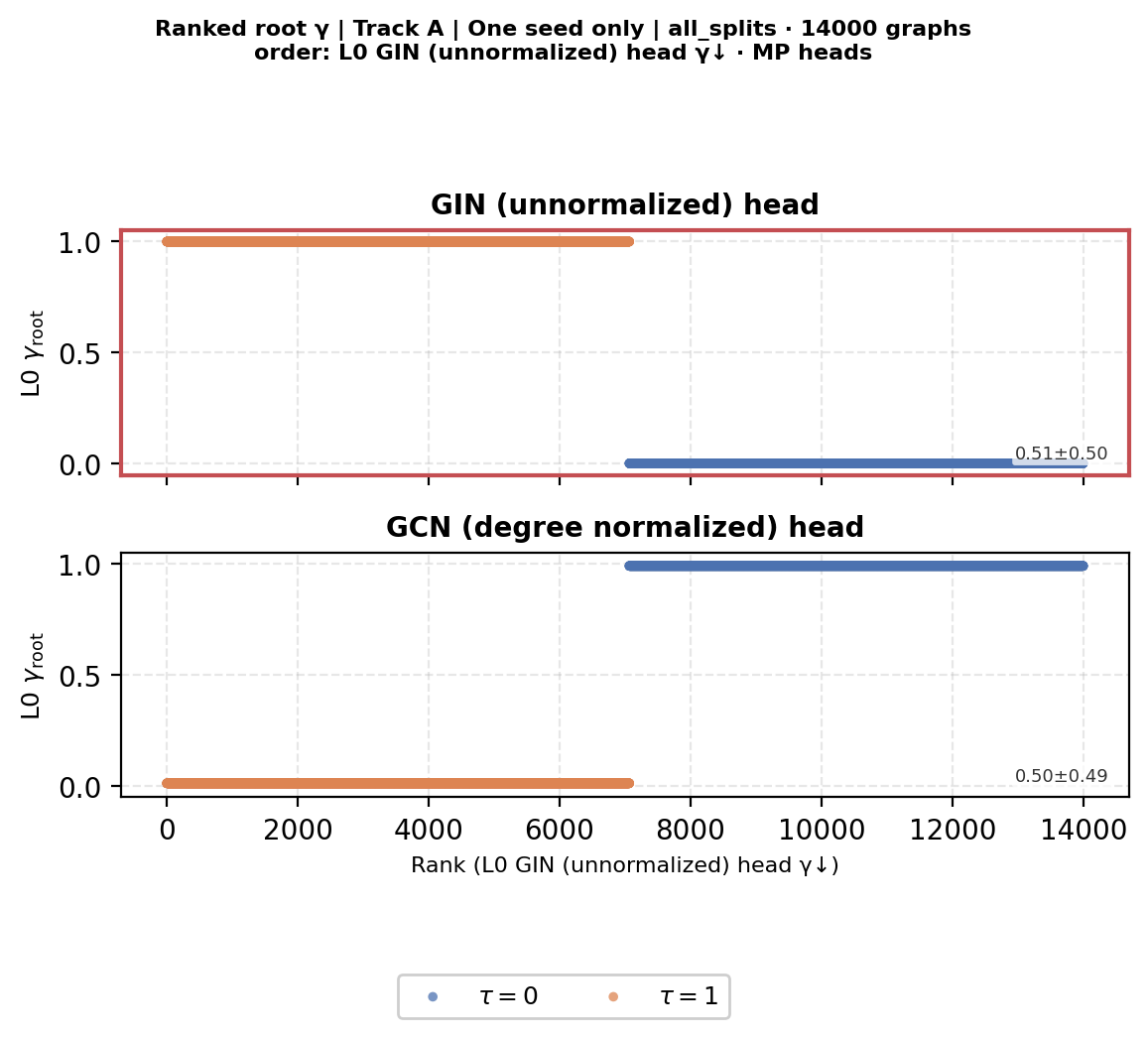}
  \caption{Gate values at the root on Track~A. The top panel is the GIN head ($\gamma_{\mathrm{GIN}}$) and the
    bottom panel is the GCN head ($\gamma_{\mathrm{GCN}}$).
    Graphs are sorted by descending root $\gamma_{\mathrm{GIN}}$.
    $\tau{=}0$ (GCN-type) graphs are in blue, \ $\tau{=}1$ (GIN-type) graphs are in orange.
    The bottom annotations report the mean$\,\pm\,$std over all graphs in the panel. We note that all $
    \tau=1$ graphs have $\gamma_{\mathrm{GIN}}\approx 1$ and $\gamma_{\mathrm{GCN}}\approx 0$, while $\tau=0$ graphs have $\gamma_{\mathrm{GIN}}\approx 0$ and $\gamma_{\mathrm{GCN}}\approx 1$.
  }
  \label{fig:ranked-root-gates-toy}
\end{figure}

\begin{figure}[t]
  \centering
  \includegraphics[width=0.8\linewidth]{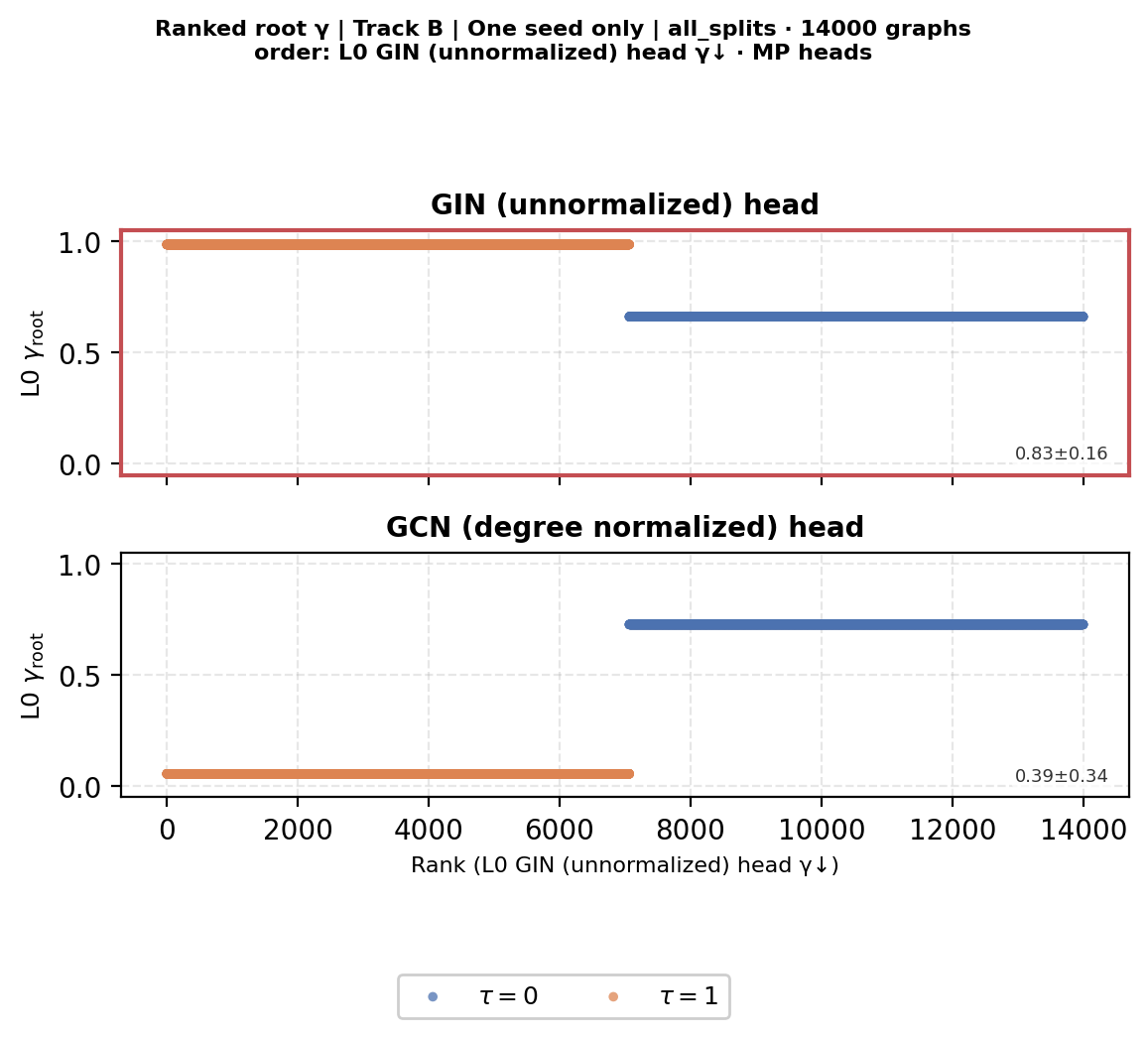}
  \caption{
Gate values at the root on Track~B. The top panel is the GIN head ($\gamma_{\mathrm{GIN}}$) and the
    bottom panel is the GCN head ($\gamma_{\mathrm{GCN}}$).
    Graphs are sorted by descending root $\gamma_{\mathrm{GIN}}$.
    $\tau{=}0$ (GCN-type) graphs are in blue, \ $\tau{=}1$ (GIN-type) graphs are in orange.
    The bottom annotations report the mean$\,\pm\,$std over all graphs in the panel. We note that all $
    \tau=1$ graphs have $\gamma_{\mathrm{GIN}}\approx 1$ and $\gamma_{\mathrm{GCN}}\approx 0$, while $\tau=0$ graphs have $\gamma_{\mathrm{GIN}} < \gamma_{\mathrm{GCN}}$.
  }
  \label{fig:ranked-root-gates-sigma}
\end{figure}

We also report the values of the gates at the root directly in   Fig.~\ref{fig:app-routing-gates} and the plots for a specific seed can be found in  Fig.~\ref{fig:ranked-root-gates-toy} and Fig.~\ref{fig:ranked-root-gates-sigma}.  Each panel ranks all graphs by descending $\gamma_{\mathrm{GIN}}$ on layer~0
(shared order across both heads) and the points are colored by graph type
($\tau{=}0$ GCN-type vs.\ $\tau{=}1$ GIN-type).
The gate value $\gamma$ at the root reflects the routing used at the readout node where the label is predicted.
On Track~A, routing at
the root is nearly disentangled:
$\gamma_{\mathrm{GIN}}$ close to $0$ on $\tau{=}0$ graphs and
$\gamma_{\mathrm{GIN}}$ close to $1$ on $\tau{=}1$ graphs, with the GCN head
showing the mirror pattern (see Fig.~\ref{fig:app-routing-gates}).
On track B, the same
directional separation holds but is softer on the GIN head for $\tau{=}0$ graphs. This indicates that SiGMA still routes preferentially toward GCN on
GCN-type instances while reserving stronger GIN mass for GIN-type instances.

\begin{figure}[h!]
  \centering
  \includegraphics[width=\linewidth]{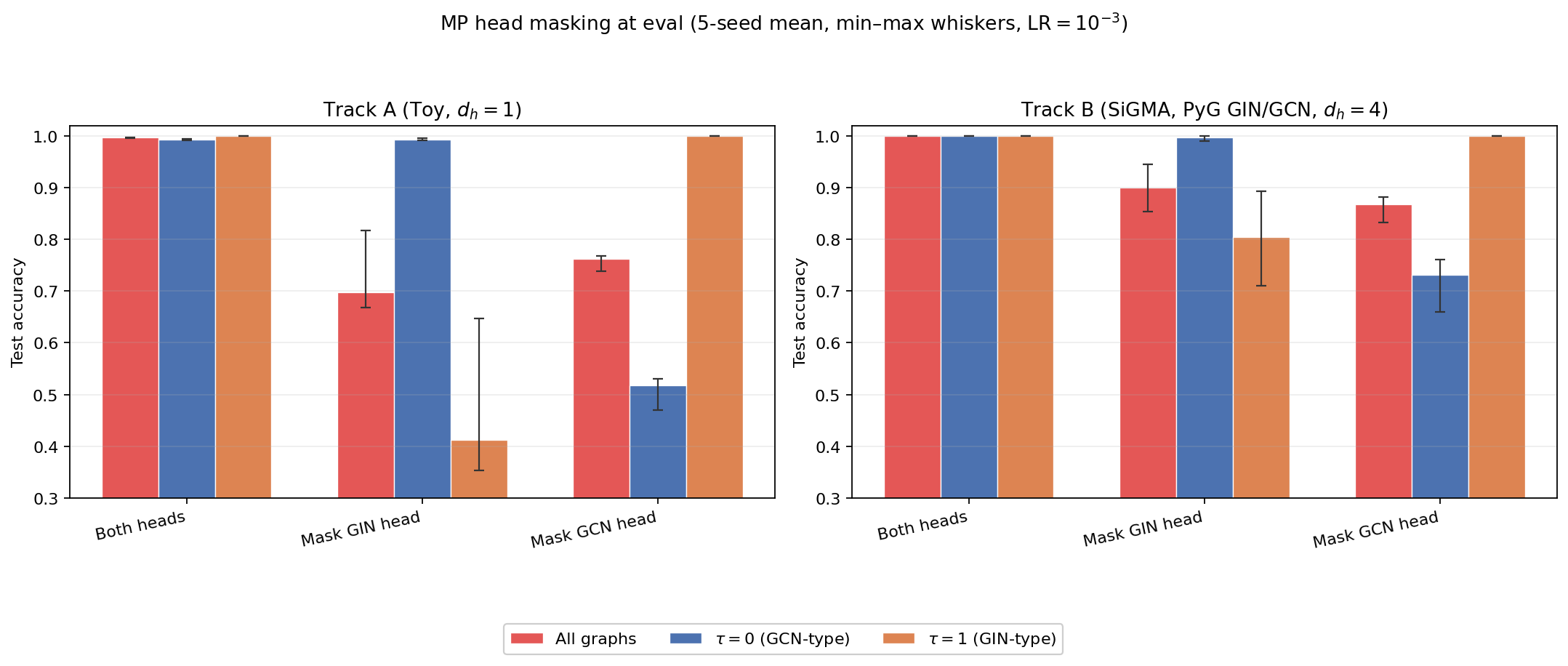}
  \caption{Head-masking ablation on both tracks on the test set using SiGMA. Removing the head that
  matches the graph type disproportionately hurts the corresponding per-type
  accuracy - this is more pronounced in track A.}
  \label{fig:app-routing-mask}
\end{figure}

\paragraph{Masking.}
With SiGMA, we zero out one MP head at a time and
re-run inference on the test set, measuring how much per-type accuracy drops relative to the
unmasked model. We present these results in Tab.~\ref{tab:app-routing-mask} and Fig.~\ref{fig:app-routing-mask}.
We note the presence of asymmetric drops in accuracy (masking GCN hurts $\tau{=}0$ more than $\tau{=}1$ and masking GIN hurts $\tau{=}1$ more than $\tau{=}0$). On Track A specifically,
masking GCN collapses $\tau{=}0$ accuracy from $\sim$99\% to $\sim$52\% while leaving
$\tau{=}1$ at $100\%$; masking GIN has the reverse effect
with $\tau{=}1$ dropping from $100\%$ to $\sim$41\%, while the accuracy on $\tau{=}0$ graphs stays at $99\%$).

\begin{table}[h!]
  \centering
  \caption{MP head masking ablation on gated SiGMA ($\mathrm{LR}=10^{-3}$, $5$-seed mean $\pm$ std, test set).}
  \label{tab:app-routing-mask}
  \begin{tabular}{llccc}
    \toprule
    Track & Eval mask & All graphs $\uparrow$& $\tau{=}0$ graphs $\uparrow$ & $\tau{=}1$ graphs $\uparrow$ \\
    \midrule
    \multirow{3}{*}{A ($d_h=1$)}
      & None (full)   & $99.6 \pm 0.0$ & $99.3 \pm 0.1$ & $100.0\pm 0.0$ \\
      & Mask GCN head & $76.3 \pm 1.2$ & $51.8 \pm 2.4$ & $100.0\pm 0.0$ \\
      & Mask GIN head & $69.8 \pm 6.0$ & $99.3 \pm 0.1$ & $41.2 \pm 11.8$ \\
    \midrule
    \multirow{3}{*}{B ($d_h=4$)}
      & None (full)   & $100.0\pm 0.0$ & $100.0\pm 0.0$ & $100.0\pm 0.0$ \\
      & Mask GCN head & $86.8 \pm 1.8$ & $73.1 \pm 3.7$ & $100.0\pm 0.0$ \\
      & Mask GIN head & $90.0 \pm 3.1$ & $99.7 \pm 0.4$ & $80.5 \pm 6.2$ \\
    \bottomrule
  \end{tabular}
\end{table}

\begin{figure}[t]
  \centering
  \includegraphics[width=\linewidth]{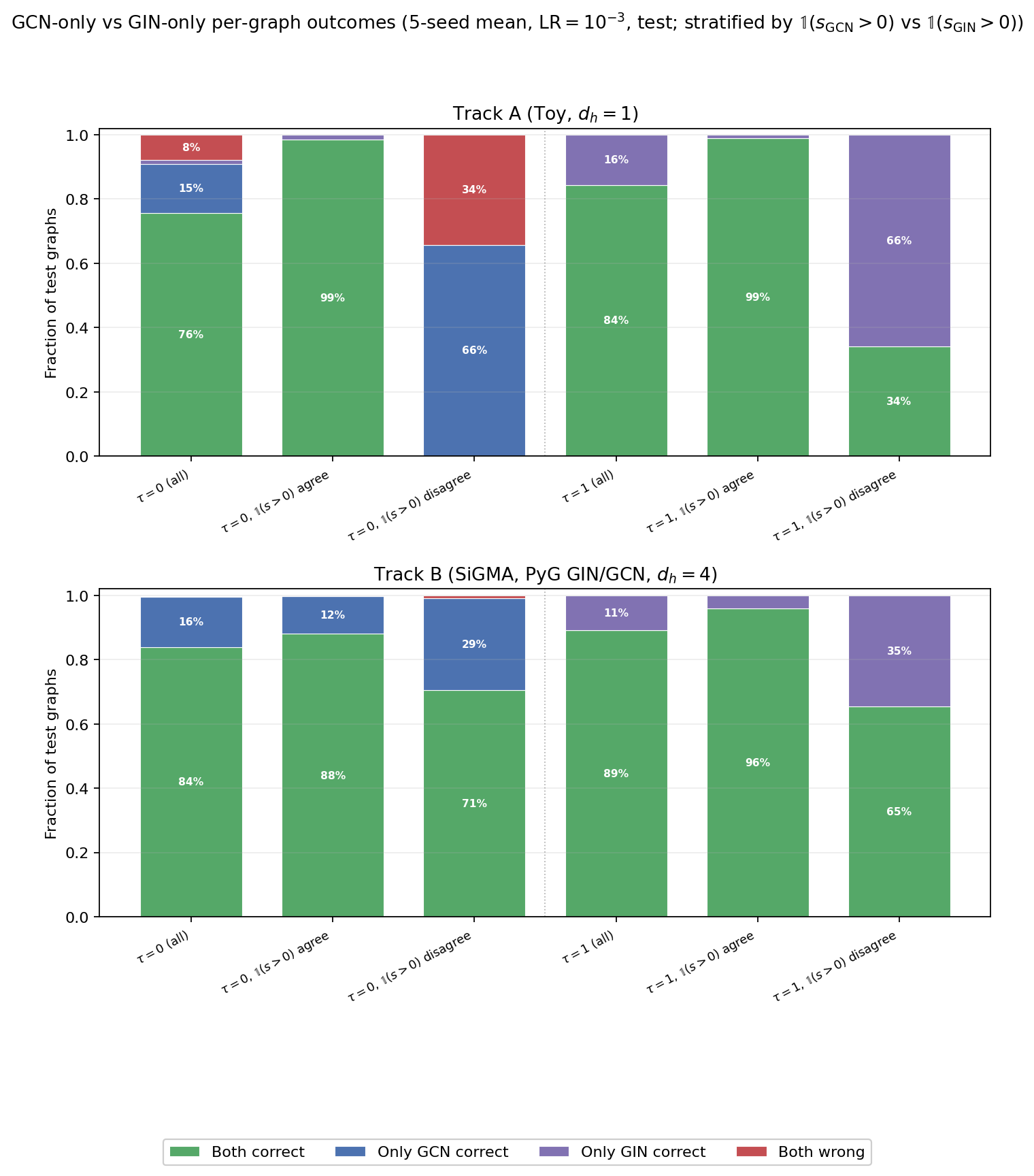}
\caption{%
  Per-graph specialist disagreement on the full test set (GCN-only vs.\ GIN-only).
  Bars partition graphs by routing label $\tau$ and by whether the analytic score signs agree, $\mathbbm{1}(s_{\mathrm{GCN}}{>}0)=\mathbbm{1}(s_{\mathrm{GIN}}{>}0)$.
  On both tracks, sign agreement is the easy regime: when the specialists agree on the classification sign ($\sim$76\% of $\tau{=}0$ and $\sim$77\% of $\tau{=}1$ graphs), both models are correct on $\gtrsim$88\% of graphs (Track~A: $\sim$99\%; Track~B: 88--96\%).
  Sign disagreement is rarer (the 228 opposite-sign twin graphs) but revealing: on Track~A, whenever signs disagree the $\tau$-matching specialist essentially always wins and the complementary specialist essentially always loses ($\tau{=}0$: 66\% only-GCN, 34\% both wrong; $\tau{=}1$: 66\% only-GIN, 34\% both correct).
  In Track~B, many disagreeing graphs are still solved by both specialists, but the residual split still favors the $\tau$-aligned rule (29\% only-GCN vs.\ 0\% only-GIN at $\tau{=}0$; 35\% only-GIN vs.\ 0\% only-GCN at $\tau{=}1$). We report 5-seed means, and use $\mathrm{LR}=10^{-3}$.
}
  \label{fig:app-routing-pairwise}
\end{figure}

\paragraph{Rule disagreement and specialist baselines.}
Figure~\ref{fig:app-routing-pairwise} summarizes, on the test set, how
independently trained GCN-only and GIN-only specialists partition
per-graph outcomes (both correct, only one correct, both wrong).
On $\tau{=}0$ graphs, only $\sim$15\% are correctly classified by the
GCN specialist alone; on
$\tau{=}1$ graphs, only $\sim$11\% (Track B) to $\sim$16\% (Track A) by the
GIN specialist alone.

\textbf{Pair-level outcomes on opposite-sign twins.}
Among test graphs that belong to matched opposite-sign twin pairs (identical
topology and neighbor features, but $\tau\in\{0,1\}$ and therefore opposite
labels), we report the fraction of pairs on which a model is correct on
both members, only the $\tau{=}0$ member, only the $\tau{=}1$ member, or
neither in Fig.~\ref{fig:app-routing-opposite-sign-pairs}. This subset tests whether the models route to the operator appropriate for each
graph type. GCN and GIN never route both twins in a pair correctly, while SiGMA is able to in $\gtrsim99$\% of instances.




\subsection{Synthesis}

\begin{figure}[t]
  \centering
  \includegraphics[width=0.9\linewidth]{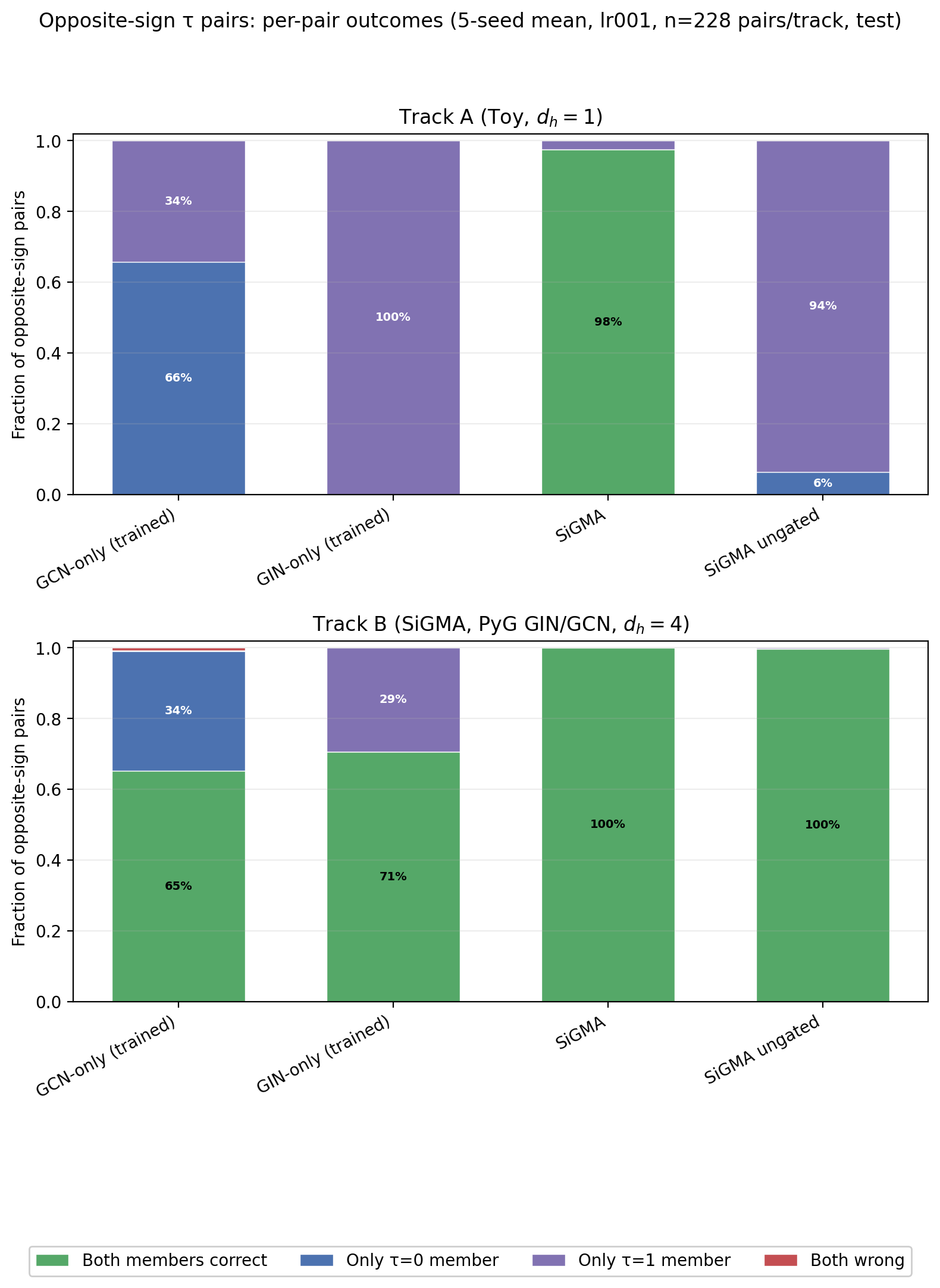}
\caption{We report the pair-level outcomes for twin pairs. We group the $n{=}228$ matched test pairs whose members share identical
  neighbor topology and features but differ in root type $\tau$ (and thus in the
  labeling rule).
  Bars show the fraction of pairs for which a method is correct on
  both members of each pair (in green), only the $\tau{=}0$ member (in blue), only the
  $\tau{=}1$ member (in purple), or neither (in red). GCN and GIN specialists never route both twins correctly on
  Track~A; gated SiGMA succeeds on ${\sim}98\%$ of pairs (Track~A) and
  $100\%$ (Track~B), while the ungated variant MMA (SiGMA ungated) collapses to GIN-like routing on
  Track~A but matches gated performance on Track~B. We report averages over 5 seeds with lr$=10^{-3}$.}
  \label{fig:app-routing-opposite-sign-pairs}
\end{figure}

Together, these analyses show that the synthetic benchmark isolates the specific form of
graph-level heterogeneity: within the same dataset, different graphs require different
message-passing operators, and no single fixed rule or specialist head is uniformly
correct.
Accuracy alone already separates the models: GCN-only and GIN-only baselines are
strong on their matching type but asymmetric on the other, whereas gated SiGMA is
near ceiling on both tracks (Tab.~\ref{tab:app-routing-acc} and  Fig.~\ref{fig:app-routing-baseline}).
The gate and masking analyses explain why: root $\gamma$ values track graph
type (Figs.~\ref{fig:app-routing-delta} and~\ref{fig:app-routing-gates} as well as   Figs.~\ref{fig:ranked-root-gates-toy} and~\ref{fig:ranked-root-gates-sigma}), and
post-hoc head masking selectively destroys the accuracy of the corresponding
$\tau$-family (Table~\ref{tab:app-routing-mask}; Fig.~\ref{fig:app-routing-mask}).
On the hardest instance, namely the matched opposite-sign twin pairs that share topology but differ
in $\tau$, only gated SiGMA reliably solves both members of a pair
(Fig.~\ref{fig:app-routing-opposite-sign-pairs}).
We summarize the main takeaways below.

\paragraph{Headline findings summary.}
\begin{enumerate}
  \item \textbf{SiGMA and gating enable mixed-type generalization.} SiGMA reaches $99.6$\% (track A) and $100\%$ (Track B) test accuracy overall,
    while GCN-only and GIN-only models have <$89$\% accuracies on Track A and <$95$\% accuracies on Track B, with large per-type asymmetry
    (Table~\ref{tab:app-routing-acc}).
  \item \textbf{Ungated hybrid is insufficient on the toy track.}
    With $d_h{=}1$, keeping both heads fully active and simply concatenating them (MMA) yields $\sim$88.6\% overall and
    only $\sim$79\% on $\tau{=}0$ graphs, vs.\ $\sim$99\% for the gated model.
    On track B ($d_h{=}4$), representation capacity partially compensates for the absence of gating
    ($\sim$99.9\% test accuracy for MMA), but gating still yields perfect per-type accuracy.
  \item \textbf{Root gates align with graph type.}
    In Track A, SiGMA's root gates are near-binary:
    $\gamma_{\mathrm{GCN}}\approx 1$ for $\tau{=}0$ and
    $\gamma_{\mathrm{GIN}}\approx 1$ for $\tau{=}1$
    (Fig.~\ref{fig:app-routing-gates}).
    In Track B, SiGMA's gates are softer but contrasts remain large
    ($\Delta\gamma_{\mathrm{GIN}}\approx 0.62$,
    $\Delta\gamma_{\mathrm{GCN}}\approx 0.87$ on average across seeds).
\end{enumerate}


\begin{table}[t]
\centering
\caption{Welch's $t$-tests of SiGMA against each variant on the synthetic routing task
(5 seeds per model; $\Delta$ in pp, positive $\Delta$ means higher accuracy for SiGMA).
$^{*}p{<}0.05$, $^{**}p{<}0.01$, $^{***}p{<}0.001$; n.s.\ = not significant.
$^{\dagger}$Both models give the same accuracy on every seed, so the difference is deterministic
and no test is needed; ``=''~denotes identical accuracy.}
\label{tab:routing-ttest}
\begin{tabular}{llccc}
\toprule
\textbf{Track} & \textbf{Variant} & $\Delta$ Acc all & $\Delta$ Acc $\tau{=}0$ & $\Delta$ Acc $\tau{=}1$ \\
\midrule
A (toy)   & MMA & $+11.0^{***}$ & $+20.6^{***}$ & $+1.8^{\mathrm{n.s.}}$ \\
A (toy)   & GCN-only      & $+12.0^{***}$ & $+8.3^{***}$  & $+15.6^{***}$ \\
A (toy)   & GIN-only      & $+11.0^{***}$ & $+22.4^{***}$ & $=$ \\
\midrule
B (SiGMA) & MMA & $+0.06^{*}$   & $+0.10^{\mathrm{n.s.}}$ & $+0.02^{\mathrm{n.s.}}$ \\
B (SiGMA) & GCN-only      & $+5.7^{***}$  & $+0.49^{*}$   & $+10.8^{***}$ \\
B (SiGMA) & GIN-only      & $+7.9^{\dagger}$ & $+15.9^{\dagger}$ & $=$ \\
\bottomrule
\end{tabular}
\end{table}


\clearpage

\newpage
\section{Hardware specifications and libraries}

\noindent All experiments in this paper were implemented in Python using PyTorch and PyTorch Geometric. 

\noindent Our experiments were conducted on the FASRC Cannon cluster supported by the FAS Division of Science Research Computing Group at Harvard University and on the Kempner Institute AI Cluster
at Harvard University, with the specifications presented in  Tab.~\ref{tab:hardware_specs}. 

\begin{table}[ht]
\footnotesize
\centering
\caption{Server specifications (one GPU per run).}
\label{tab:hardware_specs}
\begin{tabular}{lc}
\toprule
\textbf{Components} & \textbf{Specifications} \\
\midrule
Architecture & x86\_64 \\
OS & Rocky Linux 8 \\
CPU & AMD EPYC 9374F 32-Core \\ 
GPU & NVIDIA H200 / H100  \\
RAM & 128\,GB allocated per run \\
\bottomrule
\end{tabular}
\end{table}

\clearpage

\end{document}